\documentclass[10pt,dvipsnames]{article} 
\usepackage[preprint]{style/tmlr}

\usepackage{amsmath,amsfonts,bm}

\def\eqref#1{equation~\ref{#1}}

\def\1{\bm{1}}

\DeclareMathAlphabet{\mathsfit}{\encodingdefault}{\sfdefault}{m}{sl}
\SetMathAlphabet{\mathsfit}{bold}{\encodingdefault}{\sfdefault}{bx}{n}

\DeclareMathOperator*{\argmin}{arg\,min}

\usepackage[utf8]{inputenc} 
\usepackage[T1]{fontenc}    
\usepackage[]{hyperref}       
\usepackage{url}            
\usepackage{booktabs}       
\usepackage{amsfonts}       
\usepackage{nicefrac}       
\usepackage{microtype}      
\usepackage{graphicx} 
\usepackage{tabularx}
\usepackage[nohyperlinks, printonlyused, nolist, withpage, smaller]{acronym}
\usepackage{amsmath}
\usepackage[capitalise,noabbrev]{cleveref}
\usepackage{xspace}
\usepackage{tikz}
\usepackage{pgfplots}
\pgfplotsset{compat=1.18}
\usepackage{tabularx}
\usepackage{subcaption}
\usepackage{placeins}
\usepackage{wrapfig}
\usepackage{multirow}

\usepackage{physics}
\usepackage{amsmath}
\usepackage{tikz}
\usepackage{mathdots}
\usepackage{yhmath}
\usepackage{cancel}
\usepackage{color}
\usepackage{siunitx}
\usepackage{array}
\usepackage{multirow}
\usepackage{amssymb}
\usepackage{gensymb}
\usepackage{tabularx}
\usepackage{extarrows}
\usepackage{booktabs}
\usetikzlibrary{fadings}
\usetikzlibrary{patterns}
\usetikzlibrary{shadows.blur}
\usetikzlibrary{shapes}

\usepackage{array}
\usepackage{longtable}
\usepackage{booktabs}
\usepackage{paralist} 

\definecolor{blue}{HTML}{526fae}
\definecolor{green}{HTML}{63a76b}
\newcommand{\revt}[1]{\textcolor{black}{#1}}

\usepackage[colorinlistoftodos,prependcaption,textsize=tiny]{todonotes}
\usepackage{regexpatch}
\makeatletter
\xpatchcmd{\@todo}{\setkeys{todonotes}{#1}}{\setkeys{todonotes}{inline,#1}}{}{}
\makeatother
\setuptodonotes{fancyline, bordercolor=Orchid!50, backgroundcolor=Orchid!50}
\definecolor{myLinkColor}{HTML}{660066}
\hypersetup{
    colorlinks=true,
    linkcolor={myLinkColor},
    citecolor={myLinkColor},
    urlcolor={myLinkColor}
}

\def\month{MM}  
\def\year{YYYY} 
\def\openreview{\url{https://openreview.net/forum?id=XXXX}} 

\title{\carps: A Framework for Comparing\\N Hyperparameter Optimizers on M Benchmarks}
\author{
Carolin Benjamins$^1$, 
Helena Graf$^1$, 
Sarah Segel$^1$, 
Difan Deng$^1$, 
Tim Ruhkopf$^1$ \\ 
Leona Hennig$^1$, 
Soham Basu$^2$, 
Neeratyoy Mallik$^2$, 
Edward Bergman$^2$ \\ 
Deyao Chen$^3$, 
François Clément$^4$, 
Alexander Tornede$^1$, \\
Matthias Feurer$^{5,9}$, 
Katharina Eggensperger$^6$ \\
Frank Hutter$^{7,2}$, 
Carola Doerr$^4$, 
Marius Lindauer$^{1,8}$ \\
\addr
  {\centering
  \normalfont
  \begin{tabular}{c}
  \\
    $^1$ Leibniz University Hannover, $^2$ Albert-Ludwigs University Freiburg,\\
    $^3$ University of St Andrews, 
    $^4$ Sorbonne Université Paris, $^5$ LMU Munich, \\
    $^6$ University of Tübingen, 
    $^7$ ELLIS Institute Tübingen, \\
    $^8$ L3S Research Center, $^9$ Munich Center for Machine Learning
  \end{tabular}
  }
\\
\email Corresponding author email: \texttt{c.benjamins@ai.uni-hannover.de}
}
\date{October 2024}

\begin{document}
\begin{acronym}
\acro{HP}[HP]{hyperparameter}
\acro{HPO}[HPO]{hyperparameter optimization}
\acro{BB}[BB]{black-box}
\acro{MO}[MO]{multi-objective}
\acro{MF}[MF]{multi-fidelity}
\acro{MOMF}[MOMF]{multi-fidelity-objective}
\end{acronym}

\newcommand{\code}[1]{\texttt{#1}}
\newcommand{\carps}[0]{\code{carps}\xspace}
\newcommand{\carpsfull}{\textbf{C}omprehensive \textbf{A}utomated \textbf{R}esearch \textbf{P}erformance \textbf{S}tudies\xspace}
\newcommand{\numberofoptimizers}[0]{26\xspace}
\newcommand{\subsetsizebb}[0]{30\xspace}
\newcommand{\subsetsizemf}[0]{20\xspace}
\newcommand{\subsetsizemo}[0]{10\xspace}
\newcommand{\subsetsizemomf}[0]{9\xspace}
\newcommand{\fullsizebb}[0]{1\,317\xspace}
\newcommand{\fullsizemf}[0]{1\,857\xspace}
\newcommand{\fullsizemo}[0]{135\xspace}
\newcommand{\fullsizemomf}[0]{27\xspace}
\newcommand{\numberoftasks}[0]{3\,336\xspace}
\newcommand{\numberofbenchmarks}[0]{5\xspace}
\newcommand{\numberofoptfamilies}[0]{9\xspace}
\newcommand{\cpuhours}[0]{$3035$\xspace}
\newcommand{\perfopt}{\mathbf{y}_i}

\newcommand{\carpsurl}[0]{\url{https://www.github.com/automl/CARP-S}\xspace}
\newcommand{\urldocs}{\url{https://AutoML.github.io/CARP-S/latest}}
\newcommand{\repomaintainers}{the Institute of AI (Leibniz University Hannover)\xspace}
\newcommand{\urlcontributebenchmark}{\url{https://automl.github.io/CARP-S/latest/contributing/contributing-a-benchmark/}\xspace}
\newcommand{\urlexamplerepo}{https://github.com/automl/OptBench}
\newcommand{\urlcontributeoptimizer}{\url{https://automl.github.io/CARP-S/latest/contributing/contributing-an-optimizer/}\xspace}
\newcommand{\urltemplateoptimizer}{\url{https://github.com/automl/CARP-S-template/blob/main/my-optimizer.py}\xspace}
\newcommand{\urlcontribute}{\url{https://automl.github.io/CARP-S/latest/contributing/}\xspace}


\maketitle


\begin{abstract}
Hyperparameter Optimization (HPO) is crucial to develop well-performing machine learning models.
In order to ease prototyping and benchmarking of HPO methods, we propose \carps, a benchmark framework for 
\carpsfull
allowing to evaluate $N$ optimizers on $M$ benchmark tasks.
In this first release of \carps, we focus on the four most important types of HPO task types: blackbox, multi-fidelity, multi-objective and multi-fidelity-multi-objective.
With \numberoftasks tasks from \numberofbenchmarks community benchmark collections and \numberofoptimizers variants of \numberofoptfamilies optimizer families, we offer the biggest go-to library to date to evaluate and compare HPO methods. 
The \carps framework relies on a purpose-built, lightweight interface, gluing together optimizers and benchmark tasks.
It also features an analysis pipeline, facilitating the evaluation of optimizers on benchmarks.
However, navigating a huge number of tasks while developing and comparing methods can be computationally infeasible.
To address this, we obtain a subset of representative tasks by minimizing the star discrepancy of the subset, in the space spanned by the full set.
As a result, we propose an initial subset of \subsetsizemo to \subsetsizebb diverse tasks for each task type, and include functionality to re-compute subsets as more benchmarks become available, enabling efficient evaluations.
We also establish a first set of baseline results on these tasks as a measure for future comparisons.
With \carps (\carpsurl), we make an important step in the standardization of HPO evaluation.
\end{abstract}

\section{Introduction}
\Ac{HPO} is an integral step for improving the performance of machine learning (ML) methods~\citep{feurer-automlbook19a,bischl-dmkd23a}.
In turn, to develop effective and reliable \ac{HPO}
methods, thorough benchmarking is necessary~\citep{bischl-dmkd23a}, which has led to an increasing amount of benchmarks being developed for a variety of use cases~\citep{eggensperger-aaai15a,bayesmark,pineda-neuripsdbt21a,eggensperger-neuripsdbt21a,pfisterer-automl22a,salinas-automl2022}.
However, this vast number of tasks ironically impedes the benchmarking process, on the one hand, due to computational requirements and, on the other, due to potential bias from overrepresented task types.
Furthermore, developers often go for the most easy-to-use benchmark package and not necessarily for the most informative one, particularly because the latter is not obvious. 
This poses three main challenges for benchmarking a new optimizer: (i) selecting appropriate benchmark tasks and baselines, (ii) including those different benchmark tasks and baselines into an experiment setup on a technical level, and (iii)~often having to set up a new benchmarking setup from scratch, leading to a higher chance of reproducibility issues.
To address these challenges, we propose \carps, an HPO benchmarking framework for \carpsfull, which allows evaluating $N$ optimizers on $M$ benchmarks. 

\begin{wrapfigure}{r}{0.2\linewidth}
  \centering
  \vspace{-2em}
  \includegraphics[width=0.8\linewidth,trim=3.5cm 3.5cm 3.5cm 3.5cm,clip]{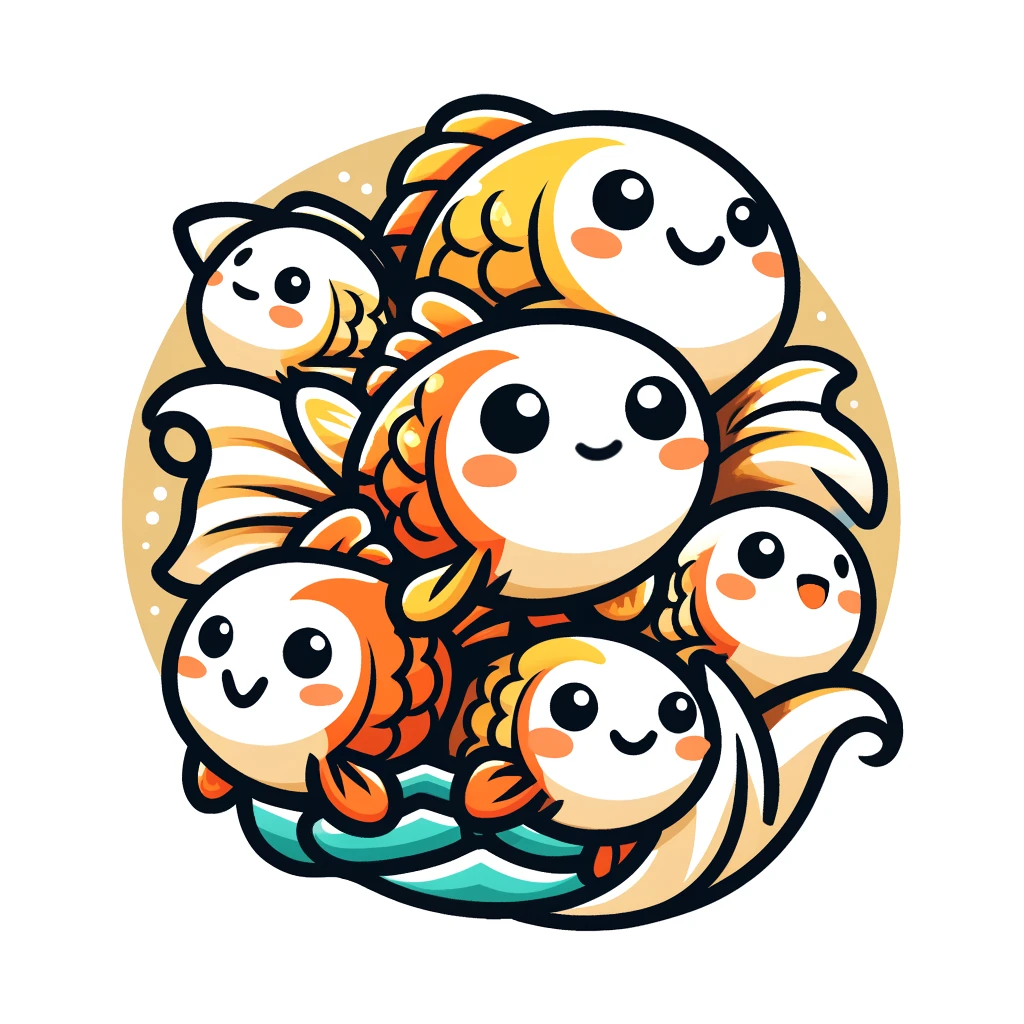}
  \caption{Our benchmarking framework \carps eases developing new optimizers by providing a multitude of diverse benchmarks with a unified interface.}
\end{wrapfigure}
\carps provides a unified access to many benchmark collections and optimizers through a lightweight interface and offers a subselection of benchmarking tasks for targeted development of new \ac{HPO} approaches.
The initial version contains a diverse set of \numberofoptfamilies optimizer families (see~\cref{tab:optimizers} in the appendix for an overview) and \numberofbenchmarks benchmark collections (see~\cref{sec:included_benchmarks}), readily available.
The lightweight interface between optimizer and benchmark eases the integration of new components, making the framework and one's own experiments easily extensible and scalable.
The included benchmarks target different HPO task types, namely blackbox (BB), multi-fidelity (MF), multi-objective (MO) and multi-fidelity-multi-objective (MOMF).
To ensure that new \ac{HPO} approaches are not overengineered to a few HPO tasks, we propose separate development and test subsets of benchmarks for these task types.
By minimizing the star discrepancy~\citep{CDP23}, we design these subsets to fully cover the entire task space.
This allows to efficiently run representative benchmarks without going for the full collection of \numberoftasks tasks.
In summary, \carps eases rapid development of optimizers and ensures quality benchmarking.

Our contributions are:
\begin{enumerate}
    \item A comprehensive experimentation pipeline from defining and running \ac{HPO} experiments to logging and analysis;
    \item A lightweight and unified interface glueing optimizer and benchmark together for ease of integration~(Section~\ref{sec:carps-interface});
    \item  Easy access to \numberoftasks HPO tasks from \numberofbenchmarks benchmark collections and \numberofoptimizers variants of \numberofoptfamilies optimizers, with seamless deployment facilitated by \href{https://hydra.cc/}{Hydra}, which supports launching their combinations on SLURM, Ray, RQ, and Joblib;
    \item Disjoint benchmark subsets for development and testing each for blackbox (BB), multi-fidelity (MF), multi-objective (MO) and multi-fidelity-multi-objective (MOMF) optimization using the star discrepancy, allowing efficient evaluation of HPO optimizers (Section~\ref{sec:carps-subset}); as well as the functionality to re-compute subsets as more benchmarks become available
    \item Experimental results available for baseline methods on these benchmark tasks (Section~\ref{sec:carps-exps}).
\end{enumerate}

\section{Background: Hyperparameter Optimization}
\label{sec:nomenclature}
Before laying out \carps as a framework with its optimizers and tasks and describing the subselection, we introduce the optimization setting and common terms.

\newcommand{\inputspace}[0]{\mathcal{X}}
\newcommand{\outputspace}[0]{\mathcal{O}}
\newcommand{\configspace}[0]{\Lambda}
\newcommand{\fidelityspace}[0]{\mathcal{F}}
\newcommand{\instancespace}[0]{\mathcal{I}}
\newcommand{\trial}[0]{\mathcal{T}}

\subsection{Optimization: Setting and Terms}
In optimization, we aim to optimize 
the \textit{objective function}, which is a mapping from an \textit{input space} $\inputspace$ to an \textit{output space} $\outputspace$:
\begin{equation}
    f: \inputspace \xrightarrow{} \outputspace \,.
\end{equation}
The output space is often half-bounded and has the number of objectives $O$ as dimensions: $\outputspace = \{(y_0, \dots, y_{O-1})\in \mathbb{R}^O | \{y_i \geq y_{i,\min}\}_{i=0,\dots, O-1}\}$.
An objective dimension can also be fully bounded for an objective like accuracy $\text{acc} \in [0, 1]$.
The output space can also contain constraints~\citep{garnett-book23a}.
The objective function can be of stochastic nature~\citep{garnett-book23a}, where we seek to optimize the expected performance, e.g., for the physical process of baking a cake and finding the optimal oven hyperparameters.
Next to that, the training of a neural network is also a stochastic process that becomes quasi-deterministic by defining a random state on the computer.
When benchmarking, instead of querying the real objective function, the objective function can also be replaced by cheap-to-evaluate alternatives like table lookups or surrogates~\citep{eggensperger-aaai15a,eggensperger-jair19a,zela-iclr22a}.

Before we elaborate on the input space and how it is composed, we define the \textit{optimization setting}.
Here, we aim to find the minimizer $x^*$ of the objective function\footnote{Note that there can be several global optima for an objective function. In general, we aim to find only one of those. We can equally formalize the optimization setting as a maximization.}:
\begin{equation}
    x^* \in \argmin_{x \in \inputspace} \, f(x) 
\end{equation}
In the case of a multi-dimensional output space, i.e. multiple objectives, $x^*$ is the Pareto front.
The optimization setting will be refined after introducing the spaces that make up the input space.

The input space itself is a cartesian product of spaces, namely the \textit{configuration space} $\configspace$, \textit{fidelity space} $\fidelityspace$ and the \textit{instance space} $\instancespace$: $\inputspace = \configspace \times \fidelityspace \times \instancespace$.
In addition, the objective function can also be of stochastic nature.

The configuration space describes the classic parameters for the objective function.
In \ac{HPO}, the objective function is an algorithm to be optimized for a certain task.
The parameters here are called \textit{hyperparameters}.
HPO can mean optimizing the hyperparameters of an ML model on a target dataset~\citep{feurer-automlbook19a, bischl-dmkd23a} for different objectives like accuracy or runtime.

Hyperparameters can be of different types, and the most common are: Continuous (\revt{float, }also on a log-scale), \revt{integers (int)}, ordinal (\revt{ord, }ordered choices or ordered discrete numbers) and categorical \revt{(cat, unordered choices or elements)}.
Next to different hyperparameter types, the configuration space can also contain constraints, conditions, and hierarchies.
An example of a mixed configuration space with hierarchies would be the choice of different optimizers for a neural network like Adam or SGD, where each choice of optimizer itself has its hyperparameters, which are different from the ones of the other optimizer choices.

In addition, for certain objective functions, we can query them using different levels of resources, called fidelity $F \in \fidelityspace$, where we assume that querying the objective function $f$ with lower resources is an approximation of querying $f$ with full resources.
An example would be the number of epochs of a training algorithm for a neural network.
Instead of evaluating the full epochs, a hyperparameter configuration $\lambda \in \configspace$ can only be evaluated with a small number of epochs to approximate the performance after a full training run.
This principle is exploited in the area of multi-fidelity optimization~\citep{jamieson-aistats16a, li-jmlr18a}, resulting in more resource-efficient algorithms.
The instance space $\instancespace$ becomes relevant in the field of algorithm configuration, where we aim to find a well-performing hyperparameter configuration for a set of objective function instances.

Therefore, with this input space $\mathcal{X}$, performing one \textit{trial} $\trial$ means evaluating the objective function with a certain hyperparameter configuration $\lambda \in \configspace$, fidelity $F \in \fidelityspace$ and instance $I \in \instancespace$, such that $x = (\lambda, F, I)$, and observing the objective function value: $T = (x, f(x))$.
In general, we \textit{only} aim to optimize in the configuration space, meaning optimizing the hyperparameters.
The fidelity space and instance space belong to the input space of the objective function, but are not a free parameter and to be optimized.
With this distinction, we define the \textit{hyperparameter optimization setting} as finding the best hyperparameter configuration $\lambda^*$ at the highest fidelity $F_{\max}$ for one fixed instance $I_f$:
\begin{equation}
    \lambda^* \in \argmin_{\lambda \in \configspace} \, f(\lambda, F=F_{\max}, I=I_f) \,.
\end{equation}
If the objective function has no notion of fidelity, it can also be viewed as evaluating always at the highest fidelity.
The best performing hyperparameter configuration $\lambda^*$ is also called the \textit{incumbent}.

One important aspect is missing when it comes to benchmarking and actual optimization.
As described, the objective function receives the input and output space.
However, it is possible to reduce the input and output space of the objective function.
For example, we can only choose to optimize two of many possible hyperparameters and only one objective of many.
Therefore, we define the \textit{task} to be an objective function with an associated (sub) input space and (sub) output space and a computational budget, e.g. a fixed number of trials or how often we can query the objective function.
Thus, we can have several tasks with the same underlying objective function.
See \cref{fig:overview_interface} for a schematic representation of a task.

Resulting of how we define the (sub) input and output spaces for an objective function as a task, we obtain several \textit{task types}.
We outline the four major task types for which we provide tasks and benchmark subselection in \carps:
\begin{description}
    \item[\ac{BB}]: Classic task type where we can only observe the output for an input, with no notion of fidelity and optimization, and one objective.
    \item[\ac{MO}]: \ac{BB} task but with multiple objectives.
    \item[\ac{MF}]: Task type, where we can query the objective function at different fidelities.
    \item[\ac{MOMF}]: Combination of \ac{MO} and \ac{MF}: We can query the objective function at different fidelities and aim to optimize over more than one objective.
\end{description}

Finally, the term \textit{task set}, also commonly referred to as \textit{benchmark}, is a collection of tasks.
In \carps we provide task sets from the literature and also a task set subselection for each of the aforementioned task types.

\subsection{Optimization Runs and Optimizers}
The key idea to optimization is to evaluate the objective function to find an optimum.
Approaches to this are diverse, from model-free to model-based, from parallel to sequential approaches.
Before diving into a quick overview of those approaches for different task types, we define the terms history, trajectory and aggregation.
\textit{History} denotes the sequence of evaluated trials together with the observed objective function value.
\textit{Trajectory} denotes the sequence of incumbents with the objective function value.
Later, to inspect and interpret results, we use \textit{aggregation} where we aggregate the optimizer histories or trajectories of a task set to a scalar.
For example, this could mean averaging the best objective function value per optimization run over tasks.
The aggregation can be performed on raw objective function values or on rankings between optimizers.

The most simple optimization algorithm is random search\footnote{Manual tuning and grid search feel also very simple and natural, but are advised against~\citep{bergstra-jmlr12a}.}, which is a model-free approach.
It can be easily parallelized and readily applied to all task types mentioned.

Then there is another important, and more sample-efficient, paradigm for optimization: Model-based and population-based approaches.
In general, model-based approaches follow a sequential optimization scheme, also being known as \textit{ask and tell} or \textit{observe and suggest}~\citep{turner-neuripscomp21a}.
First, the objective function is queried to obtain an observation.
With this observation, together with all previous observations, a model is trained to approximate the task.
This step is also called \textit{tell} or \textit{observe}.
Based on this model, the next hyperparameter configuration to be evaluated is proposed.
This step is also called \textit{ask} or \textit{suggest}.
Those steps are repeated until the optimization budget, may it be wallclock time or number of trials, is depleted.
This distinction has the advantage of moving control of the optimization run one layer up, away from the optimization method, to ensure consistent benchmarking. 
A famous representative of a model-based approach is Bayesian optimization~\citep{mockus-bo89a}.

Population-based approaches can also be formulated as sequential optimization, where after each observation a population of hyperparameter configurations is evolved.
Evolutionary algorithms like Genetic Algorithm are an example of this paradigm.

\section{Related Work}

Due to the critical role of HPO in enhancing the performance of machine learning models, several benchmarking frameworks have been developed to evaluate and compare HPO algorithms. 
In this section, we review existing benchmarking frameworks for HPO and previous approaches to select representative benchmarking subsets.

\paragraph{Benchmarking Frameworks}
We assess open-source benchmarking frameworks based on the use case, task and optimizer diversity, and extensibility.
HPOBench~\citep{eggensperger-neuripsdbt21a} and YAHPO~\citep{pfisterer-automl22a} offer collections of benchmarking tasks, with a large variety of ML algorithms and datasets but provide no integrated optimizers to compare against directly.
Bayesmark~\citep{bayesmark} is designed for Bayesian optimization methods on real machine learning tasks, supporting standard machine learning algorithms on toy datasets as well as custom data, but lacks surrogate or tabular benchmarks and support for multi-objective (MO) and multi-fidelity (MF) optimization.
In addition, it cannot be extended with new benchmarking tasks.
HPO-B~\citep{pineda-neuripsdbt21a} focuses on benchmarking black-box HPO algorithms, featuring diverse configuration spaces on numerous datasets with both tabular benchmarks and surrogates, though it has limited optimizer support and no MO or MF capabilities.
However, their tabular benchmarks do not include a search space, but only a list of available points.
Additionally, their surrogate benchmarks represent categorical hyperparameters as one-hot-encoded real-valued hyperparameters, which alters the optimization task by allowing the optimizers to query points which do not exist in the original configuration space.
Kurobako\footnote{Kurobako, \url{https://github.com/optuna/kurobako}, 2019.} is a command-line tool for evaluating black-box optimization algorithms. While Kurobako facilitates benchmarking across different task types, its primary integration is with Optuna~\citep{akiba-kdd19a} and offers only a limited selection of optimizers. Furthermore, the project has seen no active development since its last release in 2022, which may limit its suitability for ongoing and future research.
Synetune~\citep{salinas-automl2022} provides state-of-the-art algorithms with out-of-the-box tabular and surrogate benchmarks.
It supports BB and MF optimization.
\carps furthermore is the only framework that offers a sub-selection of representative benchmarking tasks based ready for developing methods and reporting results.
Please see~\cref{tab:benchmarking_frameworks} in the appendix for a compact overview.

\paragraph{Benchmark Subselection}
As a universal truth, the choice of benchmark tasks significantly influences the statistical analysis of the performance.
Evaluating the same set of algorithms on different sets of benchmarks might yield varying outcomes~\citep{cenikj-gecco22a}, motivating well-justified benchmark tasks.
\cite{pfisterer-automl22a} propose subselections for the single-objective and multi-objective cases for the surrogate-based benchmark collection YAHPO-Gym.
They have been selected based on the surrogate's approximation quality and task diversity; however, we aim to select a diverse and representative subset of tasks.
Therefore, we aim at a principled way to subselect benchmark tasks w.r.t.~diversity.
This general problem of subselecting instances to compare algorithms on has been addressed in other domains. 
For BB tasks, many methods use meta-features based on the benchmark task, excluding performance data and subselect using unsupervised learning or graph algorithms~\citep{eftimov-esa22a,cenikj-gecco22a,ispirova-esa24a,dietrich-gecco24}.
However, similar task features may result in vastly different algorithm performances~\citep{nikolikj-cec23a,long-evostar23a}.
Instead of using landscape features, \cite{cenikj-gecco22a} build a similarity graph based on trajectory features.
Based on this graph, a graph algorithm selects diverse, representative, and non-redundant tasks.
Whilst subselecting a diverse set of tasks, their method relies on hyperparameters of their selection method, which are not intuitive to set, and meta-features must be derived anew for each domain.

In the field of ML there are multiple works that create subsets of benchmarks due to an abundance of potential tasks.
The creators of OpenML proposed to filter datasets based on inclusion and exclusion criteria and created the OpenML-CC18~\citep{bischl-neuripsdbt21a}.
Due to the unsupervised nature of the process, there is no guarantee that the instances are complementary and cover the whole instance space.
Aiming to reduce the 72 datasets of OpenML-CC18 to a minimal subset, \citet{cardoso-arxiv21a} apply item-response theory~\citep{martinezplumed-ecai16a} and found that only 10 of the datasets were really hard, but also, that a subset consisting of 50\% of the datasets is sufficient.
The TabZilla benchmark suite~\citep{mcelfresh-neurips23a} is also based on OpenML data, but uses three heuristics to include only datasets that cannot be solved by baseline algorithms, are hard for most algorithms, and are hard for gradient-boosted decision trees, thereby taking information about the current landscape of ML algorithms into account.
Lastly, the Lichtenberg-MATILDA approach~\citep{pereira-dmkd24a} aims to find a maximum-coverage set of instances in a 2d projection of the instance space.
This, however, requires again instance or task features.
Instead, \carps performs subselection of tasks in the performance space.

\section{\carps: Framework Overview}

We designed the \carps framework with the following desiderata in mind:
First, adding your own optimizer, benchmarking and performing evaluations is easy by providing standardized and straightforward interfaces, lowering the entrance barrier to the field.
Second, representative benchmark collections for the development and testing of optimizers on each task type are offered to allow for fast prototyping and easy comparison between optimizers while requiring fewer computational resources.
Third, \carps contains an analysis pipeline to facilitate the interpretation and presentation of results.
Fulfilling these desiderata ensures that \carps is a framework that brings novel value to the community. 
\carps contains \numberoftasks tasks from \numberofbenchmarks benchmark collections and \numberofoptimizers optimizers from  \numberofoptfamilies optimizer suites.

For a more technical, in-depth view, we provide tutorials and a template repository on how to use the framework and how to add your own benchmark and optimizer in the documentation.\footnote{\urldocs}

\paragraph{Reproducibility and Large-Scale Experiments}
\carps is implemented and available under an OSS license.
Experiment runs can be easily parallelized and launched via \href{https://hydra.cc/}{Hydra} supporting SLURM (submitit), Ray, RQ, and Joblib.
In addition, \carps aims to be as reproducible as possible.
For this, apart from package specification, singularity containers can be used in combination with experimentation scheduling and logging to a MySQL database as an experimental feature.

\subsection{Interface} \label{sec:carps-interface}
\begin{figure}[t]
    \centering
    \includegraphics[width=0.8\linewidth, trim={0 0 2cm 0}]{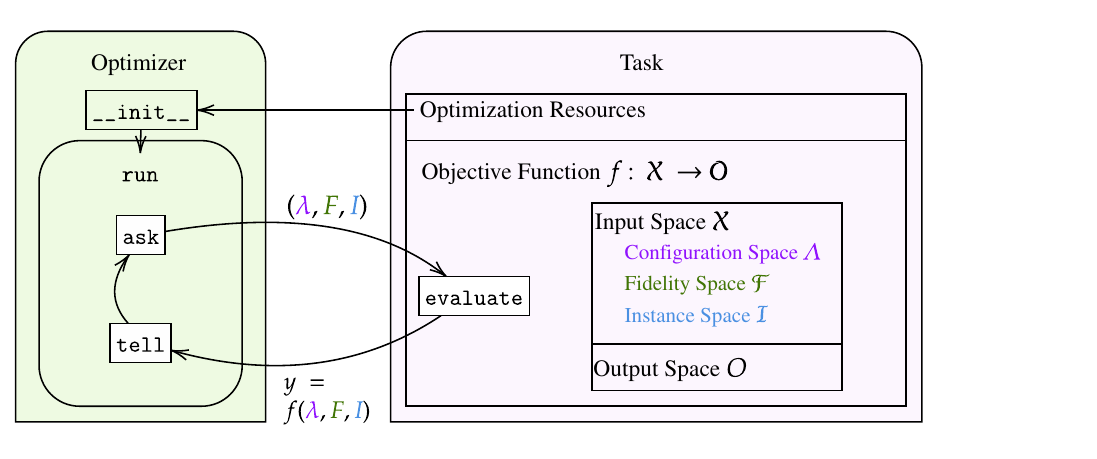}
    \caption{Overview of basic interface methods and interactions in \carps.
    The \code{Optimizer} class orchestrates optimization and runs the optimization loop via ask-and-tell. The actual optimizer variant only needs to implement the functions \code{ask} and \code{tell}.
    The \code{Task} defines the optimization resources and holds the \code{ObjectiveFunction}, which is queried in between (\code{evaluate}).
    Both are configured via files using \href{https://hydra.cc/}{Hydra}.}
    \label{fig:overview_interface}
\end{figure}

The interface between optimizers and tasks is kept as lean as possible while still allowing flexibility for different use cases.
This is achieved by providing abstract implementations for both the optimizer and the objective function that can be subclassed.
Information on evaluations is exchanged via structures holding all necessary information to perform an evaluation (\code{TrialInfo}) and information obtained after performing the evaluation (\code{TrialValue}).
The classes and structure follow the conventions introduced in~\cref{sec:nomenclature}.
An overview of the basic interface is given in~\cref{fig:overview_interface}.
In general, each component is specified and configured via configuration files. 
See \cref{sec:interface_technical_details} for a more technical interface description.

\paragraph{Optimizer}
The optimizer orchestrates the optimization.
It receives the task and, as such, must be capable of converting the configuration space (the established \code{ConfigSpace}~\citep{lindauer-arxiv19a}) to its own configuration space -- in practice, this requires only minimal efforts, e.g., matching a float hyperparameter of \code{ConfigSpace} to the float hyperparameter of the optimizer's configuration space.
The basis for the optimizer interface is then the ask-and-tell interface, wherein the \code{ask} method prompts the optimizer to return a new trial to evaluate (here, \code{TrialInfo}) and the \code{tell} method allows reporting back evaluation results (here, \code{TrialValue}).
In addition, there must be a function to obtain the current incumbent, which is especially important for MF and MO optimization, where the strategy to determine an incumbent is optimizer-dependent.
In the case of MO, the incumbent would be the Pareto front of configurations.
Although not recommended, \carps{} also allows optimizers not to implement an ask-and-tell interface; however, then comparable resource limitations cannot be guaranteed since the optimizer has to take care of it itself, which can lead to unexpected benchmarking results~\citep{eggensperger-jair19a}.

\paragraph{Objective Function and Task}
The objective function interface's mandatory methods include only two functionalities: (i) converting the objective function's configuration space into the unified configuration space for the optimizer, and (ii) evaluating a \code{TrialInfo} and returning results as a \code{TrialValue}.
Optionally, if known, the global function minimum can be returned.
Together with the optimization resources, the objective functions form a \textit{task}.
The optimization budget can be the number of trials (function evaluations) or time; we focus on the former.
The number of trials depends on the dimensionality $d$ of the objective function and is calculated with $n_{trials} = \lceil 20 + 40 \sqrt{d} \rceil$, the same as in YAHPO-Gym~\citep{pfisterer-automl22a}.

\section{Optimizer Overview}
To facilitate easy comparisons among multiple optimizers, \carps provides access to a wide range of optimization algorithms.
Aligned with the integrated task sets, \carps supports optimizers for the task types BB, MO, MF, and MOMF. 
Besides RandomSearch~\citep{bergstra-jmlr12a}, HEBO~\citep{cowenrivers-jair22a}, Skopt\footnote{Scikit Optimize, \url{https://github.com/scikit-optimize/scikit-optimize}, 2018.} and DEHB~\citep{awad-ijcai21a}, 
Ax\footnote{Ax: An accessible, general-purpose platform for understanding, managing, deploying, and automating adaptive experiments, \url{https://github.com/facebook/Ax}. Created and maintained by Meta.} based on BoTorch~\citep{balandat-neurips20a},
multiple variants of Optuna~\citep{akiba-kdd19a}, Nevergrad~\citep{rapin-2018a}, SMAC3~\citep{lindauer-jmlr22a} and Synetune~\citep{salinas-automl2022} are included. 
We use the default settings of the optimizer variants for the different task types.
\Cref{tab:optimizers} in the appendix provides an overview of the optimizers in \carps with their respective task types.

\begin{figure}[ht]
    \centering
    \includegraphics[width=0.95\textwidth]{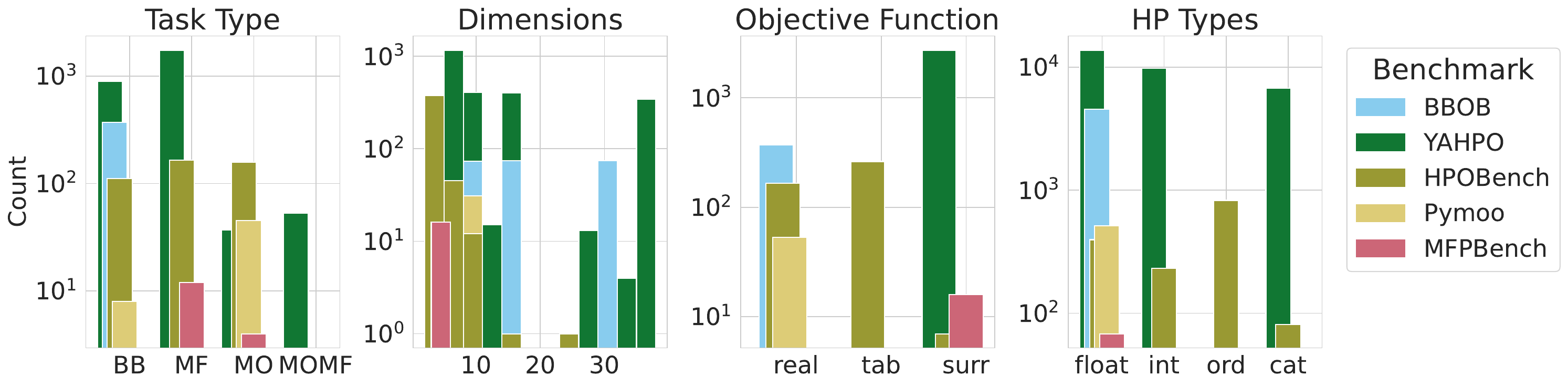}
    \caption{Statistics for all tasks included in \carps, \revt{ distinguished by benchmark collections. All benchmark collections exhibit different profiles along task types (first), dimensionality (second), objective function type (third) and hyperparameter types (fourth). As task types, we have \acf{BB}, \acf{MF}, \acf{MO}, and \acf{MOMF}. An objective function can be a real function evaluation (real), a look-up table (tab) or a surrogate (surr). As \ac{HP} types we have continuous (float), integers (int), ordinals (ord, ordered and discrete elements), and categoricals (cat, unordered elements).}}
    \label{fig:benchmark_footprints}
\end{figure}

\section{Included Task Families}
\label{sec:included_benchmarks}
With \carps we aim to ease accessibility to \ac{HPO} tasks.
We focus on four task types, namely \textit{blackbox} (BB), \textit{multi-fidelity} (MF), \textit{multi-objective} (MO) and \textit{multi-fidelity-multi-objective} (MOMF).
The integrated benchmark collections offer tasks for each task type.

There is a diverse set of benchmark tasks included in \carps. 
Upon release, the benchmark collections provided are BBOB~\citep{hansen-oms20a}\footnote{We include synthetic functions, such as BBOB and Pymoo-MO, to study performance on an established set of tasks with known characteristics.
However, since we focus here on HPO tasks, we emphasize that results on these functions do not necessarily generalize to performance on actual HPO tasks and might require different search behavior~\citep{benjamins-automl23a}.}
, HPOBench~\citep{eggensperger-neuripsdbt21a}, YAHPO~\citep{pfisterer-automl22a}, MFPBench~\citep{mallik-neurips23a} and Pymoo-MO~\citep{blank-ieeeaccess20}.
In addition to the task types they address, the benchmarks can be characterized by the number of dimensions, i.e., the number of hyperparameters and their types. 
An overview of all tasks and their respective characterization, i.e., the sum of task types, objective functions, and hyperparameter types over all tasks in the benchmark, as well as the task's dimensions, can be found in~\Cref{fig:benchmark_footprints}.

\section{Benchmark Subselection} \label{sec:carps-subset}
As \carps includes many benchmark families, with a total number of around \numberoftasks tasks, developing, evaluating and reporting the performance of an optimizer can be easily biased as the distribution of tasks per benchmark collection is not equal.
Furthermore, extensive evaluations on many tasks become computationally impractical due to optimizer and objective function evaluation overhead. 
Therefore, we propose to \textit{subselect} representative benchmarking tasks for each task type. 
In addition, we propose to use two disjoint sets of benchmarking tasks, one for the development phase and one for reporting unbiased performance, similar to the commonly applied train/test split for assessing supervised ML.
The next paragraphs describe the subselection methodology, the setup and the subselection results.


\subsection{Subselecting Representative Sets}
\label{sec:subselecting}
\begin{figure}
    \centering
    \includegraphics[width=0.95\linewidth]{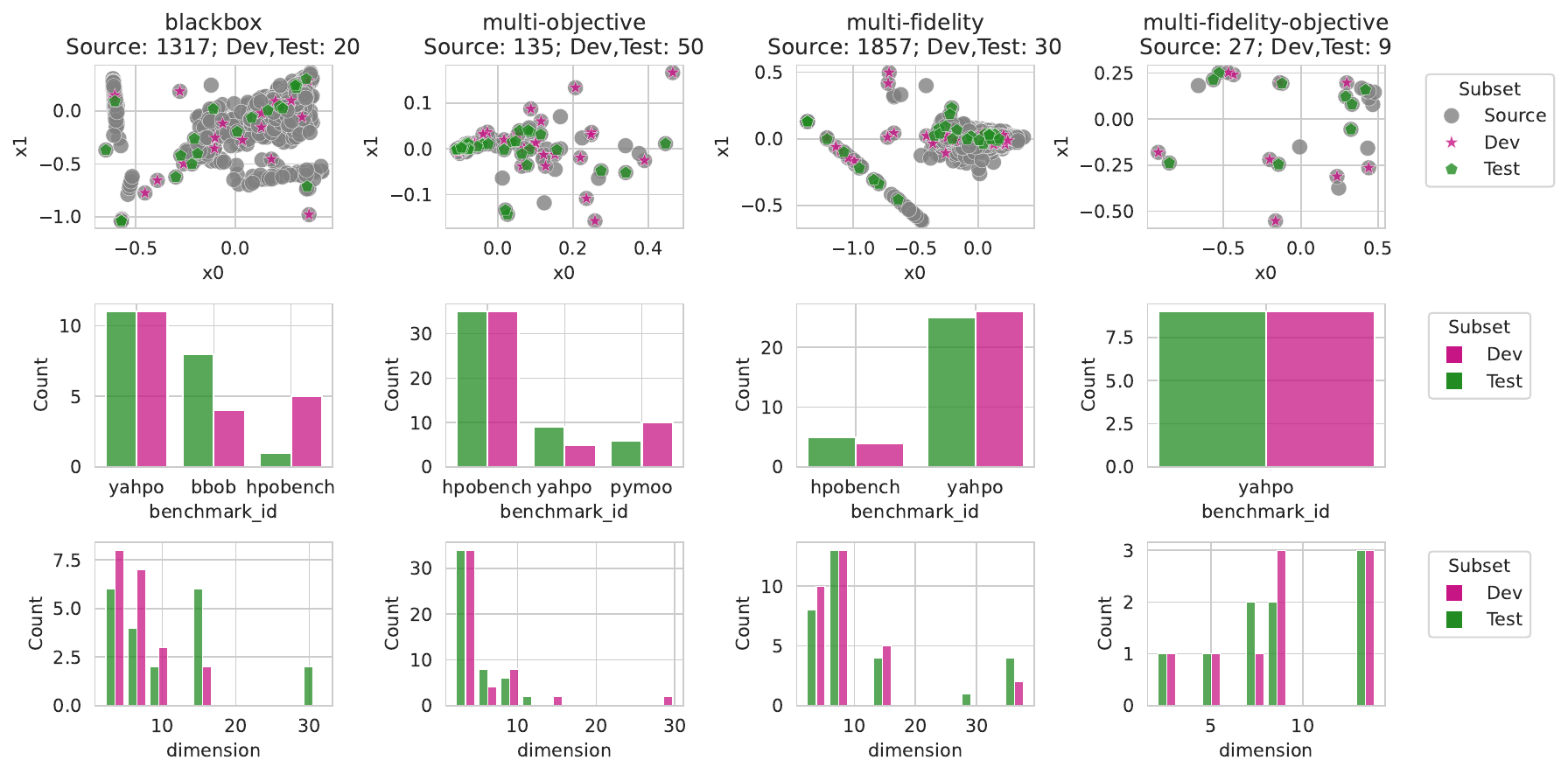}
    \caption{Subselection results summary. \textbf{Top row:} Each dot means one task and is represented by the 3-dimensional optimizer performance vector \revt{$\perfopt$ (\cref{eq:points})}, which is reduced via PCA to 2 dimensions. The subselected tasks follow the source distribution. \textbf{Middle row:} The histogram of the selected benchmark families. \textbf{Bottom row:} Histogram of the different dimensionalities of the selected tasks.}
    \label{fig:subset_stats}
\end{figure}

Because of the sheer mass of benchmarking tasks, which is also not equally distributed across benchmarks (hence with the ability to strongly bias reporting of performance~\citep{cenikj-gecco22a}), we suggest an initial subselection of representative benchmarking tasks for each task type.
In addition, we propose to select two sets: one for the development of an optimization method and one for testing its performance.
We can abstract the subselection as an optimization problem where, from a point cloud, we want to select those $k$ points that best cover the space spanned by the point cloud.
There are two avenues to create points to subselect from: (i) task features and (ii) solely from performance data.
For the first option, recent works strongly suggests that state-of-the-art features for HPO and BB tasks do not capture the objective function structure well enough for AutoML approaches~\citep{nikolikj-cec23a,long-evostar23a, vermetten-automl23a}.
Since this weakness of black-box optimization features has been observed by the community for several years now, without proper remedies in place, we resort to the second option, subselecting in the performance space, which reflects the behavior and performance of the optimizers.
Formally, we have the set of $M$ benchmarking tasks together with the performance of $N$ optimizers:
\begin{equation}
\label{eq:points}
    P = \{ \perfopt \}_{i=1,\dots,M}, \, \revt{\perfopt \in [0,1]^N} \,.
\end{equation}
$\perfopt$, therefore, holds the final performance of each optimizer, i.e. the performance of the incumbent, on the benchmarking task $i$.
\revt{The performances are scaled to the unit-interval per task to accomodate different output scales of tasks.}
We want to select two subsets of $P$, each of which is as representative as possible of the initial set of $M$ tasks.
Discrepancy measures are frequently used to quantify how uniformly distributed a given set of points is.
\revt{Let $q$ be a $d$-dimensional vector in the unit cube $q=(q_1,\ldots,q_d) \in [0,1]^d$ and $[0,q)$ the $d$-dimensional box with a corner in the origin $[0,q_1) \times [0,q_2) \times \ldots\times[0,q_d)$.}
The $L_{\infty}$ star discrepancy of a set $P$, $d_{\infty}^*(P)$, measures the worst absolute difference over all \revt{such} boxes $[0,q) \subseteq [0,1)$ between the Lebesgue measure of this box and the proportion of points $|P \cap [0,q)|/|P|$ that falls inside this box, see~\cref{fig:exdiscre} in the appendix for a visualization.

More formally, for a point set $P$ in dimension $d$, it is given by
\begin{equation}
d^*_{\infty}(P) := \sup_{q\in[0,1\revt{]}^d} \left| \frac{|P \cap [0,q)|}{|P|} - \lambda(\revt{[0,q)}) \right|,
\label{eq:disc_def}
\end{equation}

where $\lambda$ is the Lebesgue measure, \revt{$\lambda([0,q)) = \prod_{i=1}^d q_i$.
For $d=1$ it measures the length, for $d=2$ the area, and for $d=3$ the volume of the box.
In our case, the dimension equals the number of optimizers, $d=N$.}
It has been used in a very wide variety of applications from computer vision to financial mathematics and Quasi-Monte Carlo integration~\citep{MatBuilder, GalFin,santner-book03a, DickP10}.
In machine learning, low discrepancy constructions such as those suggested by Sobol'~\citep{Sobol} and Halton~\citep{Halton64} are used for hyperparameter optimization~\citep{BousquetGKTV17, CauwetCDLRRTTU20} and to initialize surrogate-based optimization algorithms~\citep{jones-jgo98a,snoek-nips12a}. 
To obtain low discrepancy subsets, we make use of an approach proposed in~\citep{CDP23}.
Interestingly, their work was originally motivated by a similar problem as ours, the search for diverse instances of the traveling salesperson problem~\citep{NeumannGDN018}.


In our setting, the $L_{\infty}$ star discrepancy can be used to characterize the quality of the selected sets.
Thus, 
one way to formulate our problem is:
Find the set of $k$ points that minimizes the star discrepancy to the complete set.
This corresponds to the approach to the Star Discrepancy Subset Selection Problem~\citep{CDP22,CDP23}.
For a description of the algorithm, please see~\cref{sec:star_disc_opt_desc}.

In order to determine the subset size $k$ and because optimization of the subset is resource-intensive,  we optimize for $k \in \{10,20,30,\dots,100\}$ (or maximum half of the full set size) and select the $k$ points, for which the sum of the star discrepancies of the development and test subset is the lowest.
To obtain two sets, the development and test set, we perform this subset selection twice: once on the $m$ initial tasks and a second time on the $m-k$ remaining ones.
With the huge number of available tasks $m$, this is possible because $m-k$ is of the same order as $m$, and the distribution of the remaining $m-k$ points will still resemble that of the full set. 
By construction, the obtained sets represent the original set in the performance space.


For each task type, we select three diverse optimizers and record their mean performance of 20 seeds per task\revt{, forming a 3-dimensional optimizer performance vector per task $\perfopt \in \mathbb{R}^3$ (\cref{eq:points}), see~\cref{fig:subset_stats} top row}.
To perform the subselection, the performance must be normalized to the unit interval to become the source space for the subselection routine.
In addition, the source space can be further log-transformed.
In total, we must set two hyperparameters for the subselection routine: The subset sizes and whether to perform the subselection in the log-transformed source space or not.
We select the subset size and transformation post-hoc via filtering and decision rule: \begin{inparaenum}
    \item Calculate ranks based on non-parametric test (\cref{sec:autorank});
    \item keep combinations, where the rank stays the same between the dev and test subset;
    \item from those, keep combinations exhibiting significant differences; and
    \item pick the combination filling the source space best, indicated by the sum of discrepancies of the dev and test subset.
\end{inparaenum}

\subsection{Subselection Setup and Results}
For the blackbox task type, we generate the source set from the performance of random search, CMA-ES and Bayesian optimization.
According to the aforementioned workflow, we obtain a subset of size $k=20$ orginating from the log-transformed source space.
For multi-fidelity, we obtain a subset size of $k=30$ from the log-transformed performance source space from Hyperband, DEHB and BOHB.
For multi-objective, the subset size is $k=50$ from the non-transformed performance source space generated from running random search, Optuna-MO and differential evolution from Nevergrad.
Last but not least, for multi-fidelity-objective, the subset size is $k=9$ from the log-transformed space from the optimizers random search, SMAC3-MOMF-GP and differential evolution from Nevergrad (the latter ignores the multi-fidelity).
The selected subsets per task type are visualized together with the histogram of the task benchmark families and dimensions in~\cref{fig:subset_stats}.
For the complete list of tasks per task type see~\cref{sec:selected_sets}.

\section{Benchmarking with \carps} \label{sec:carps-exps}
For each task type, we put \carps into action and describe the experimental results.
We select one representative optimization method from each optimizer family and run on the subselected task set for each task type.
We run everything for 20 random seeds.
Please find the code and the raw experimental results in our GitHub repository (\carpsurl).
Running the main experiments for one optimizer requires approximately \cpuhours of CPU hours, with only $21$ hours run for the black-box subsets; more details in~\cref{sec:computational_resources}.

\subsection{Analysis Pipeline}
\label{sec:autorank}
\begin{figure}[ht]
    \centering
    \begin{subfigure}{0.49\linewidth}
        \centering
            \includegraphics[width=\linewidth]{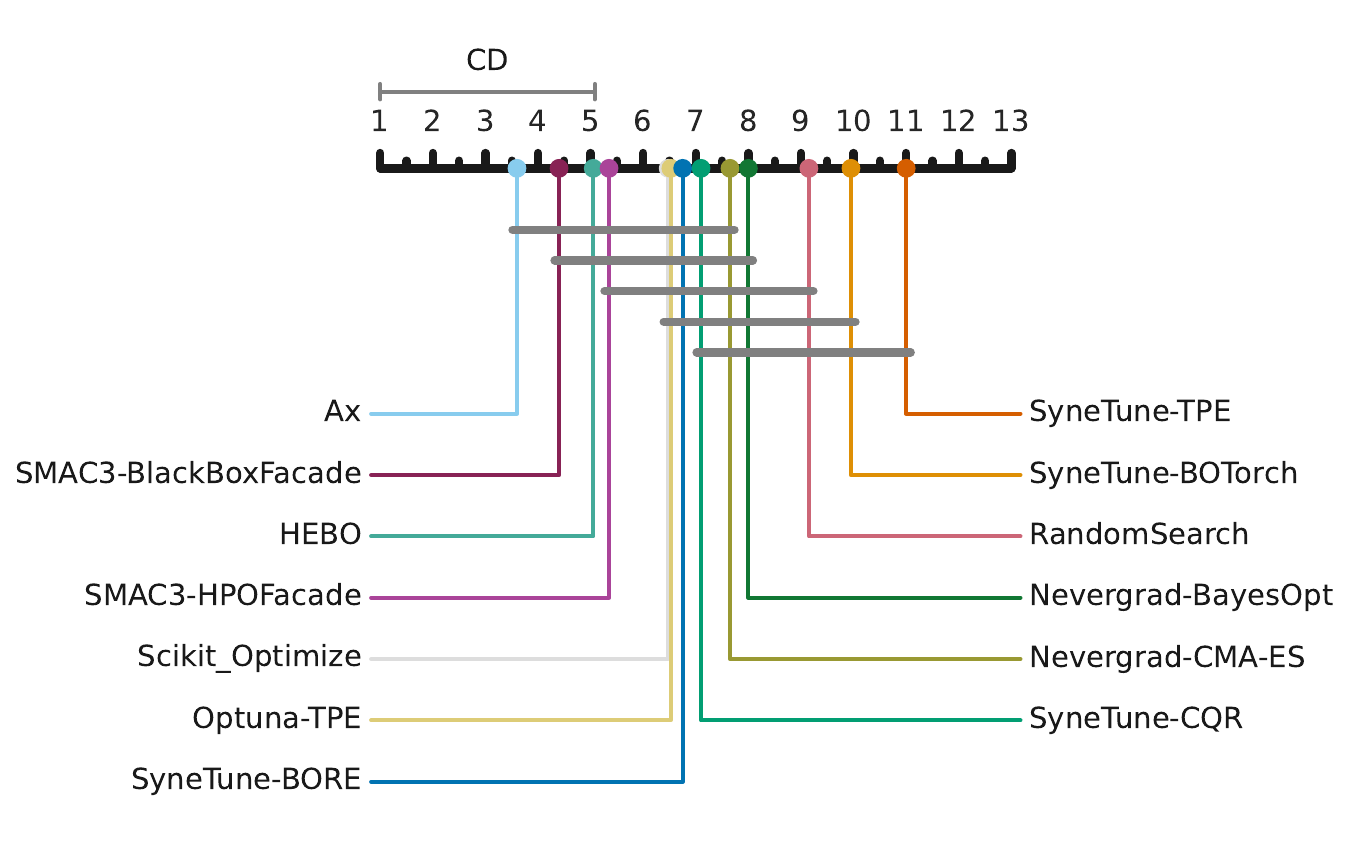}
        \caption{Critical Difference Diagram}
        \label{fig:bb_test_cd}
    \end{subfigure}%
    \begin{subfigure}{0.49\linewidth}
        \centering
        \includegraphics[width=\linewidth]{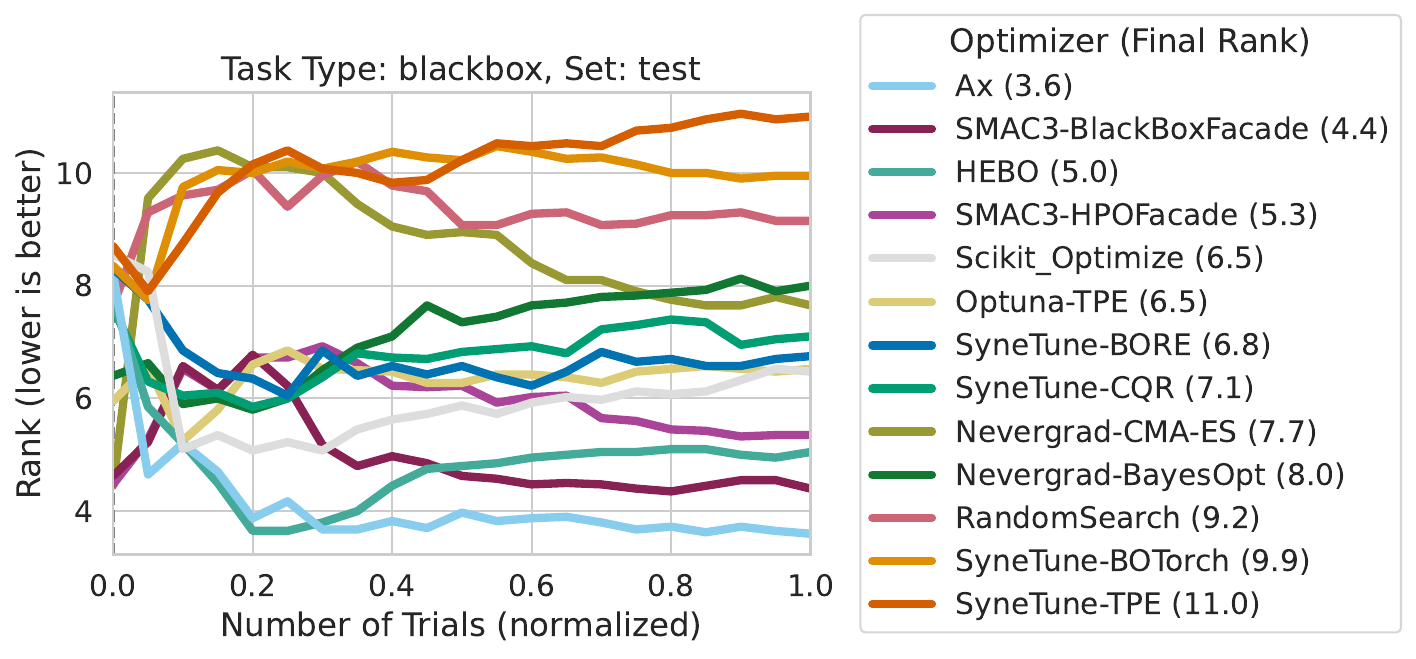}
        
        \caption{Rank over Time. The grey area indicates non-significance.}
        \label{fig:bb_test_rank}
    \end{subfigure}%
    \caption{Blackbox task type (test set)}
    \label{fig:results_bb_test}
\end{figure}

We analyze experimental results from different viewpoints.
Most importantly, we aggregate the results via rankings.
We use the library \code{autorank}~\citep{herbold-joss20} for determining the ranks and critical differences.
The ranking is performed on the raw performance values, averaged across seeds.
To be more precise, we use the frequentist approach~\citep{demsar-06a}: We use the non-parametric Friedman test as an omnibus test to determine whether there are any significant differences between the median values of the populations.
We use this test because we have more than two populations, which cannot be assumed to be normally distributed.
We use the post hoc Nemenyi test to infer which differences are significant.
The significance level is $\alpha=0.05$.
In order to be considered different, the difference between the mean ranks of two optimizers must be greater than the critical difference (CD).
We visualize the ranks for the final incumbent with indication of CD (\cref{fig:bb_test_cd}) and the ranks over time (\cref{fig:bb_test_rank}).
In addition, we also show the performance per task and optimizer as a heatmap (\cref{fig:bb_test_perfpertask}).
Please find results for all task types and subsets in \cref{sec:results_app}.


\subsection{Insights}
\begin{figure}[h]
    \centering
    \includegraphics[width=0.6\linewidth]{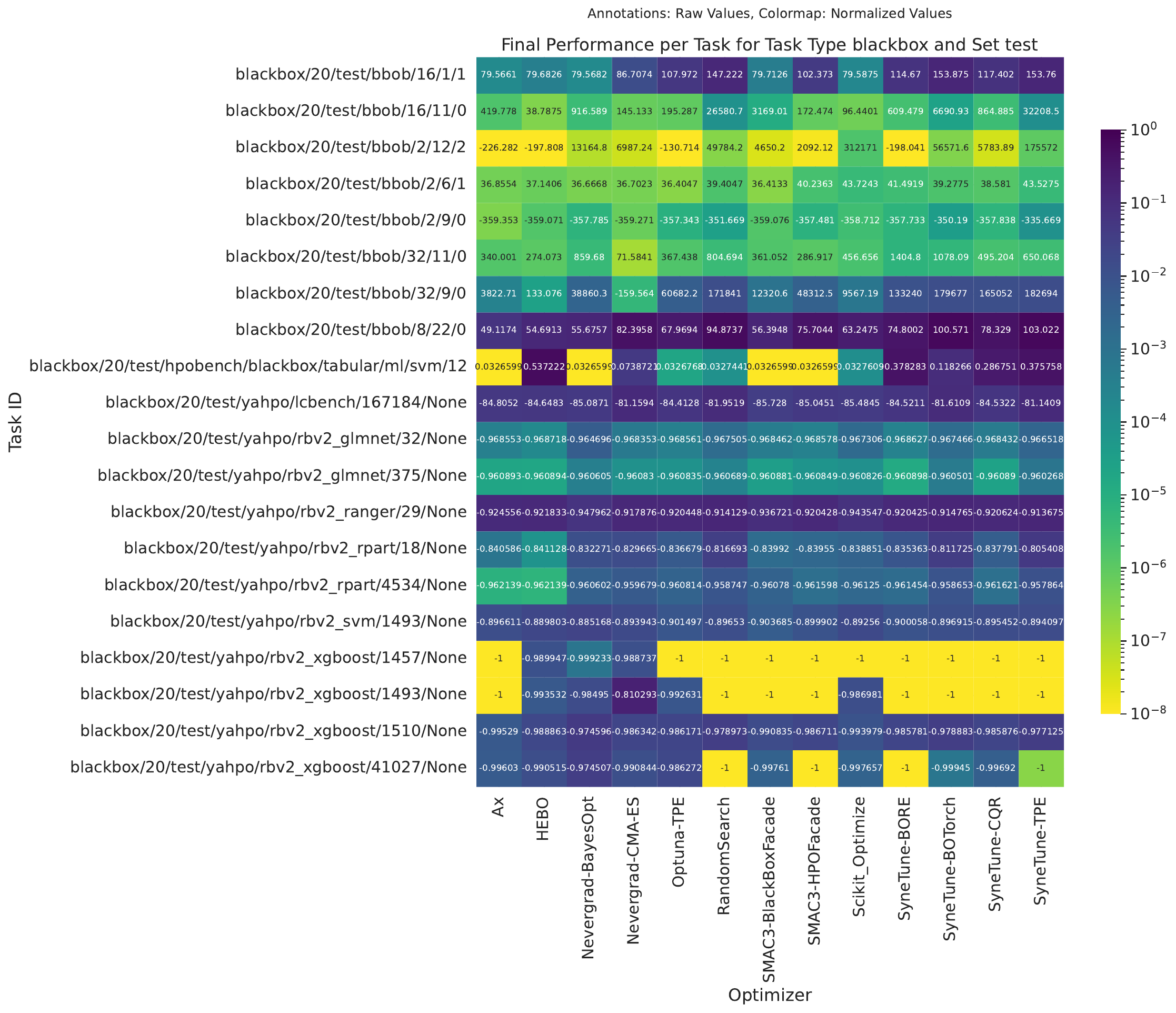}
    \caption{Blackbox task type (test set): Final incumbent cost per task, averaged over 20 seeds.}
    \label{fig:bb_test_perfpertask}
\end{figure}

\begin{figure}[ht]
    \centering
    \begin{subfigure}{0.4\linewidth}
        \centering
        \includegraphics[width=\linewidth]{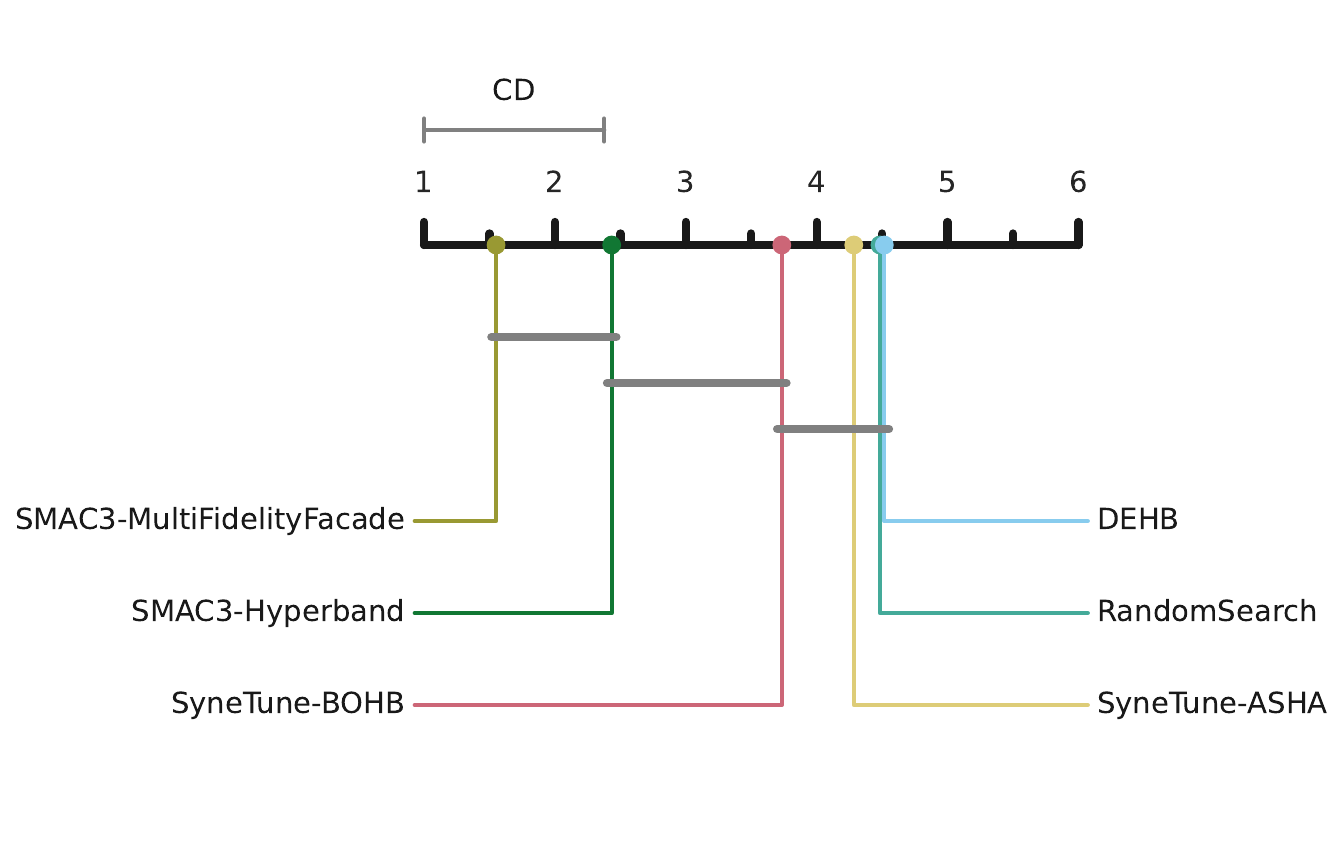}
        \caption{Multi-fidelity task type (test set)}
        \label{fig:mf_test_cd}
    \end{subfigure}
    \begin{subfigure}{0.4\linewidth}
        \centering
        \includegraphics[width=\linewidth]{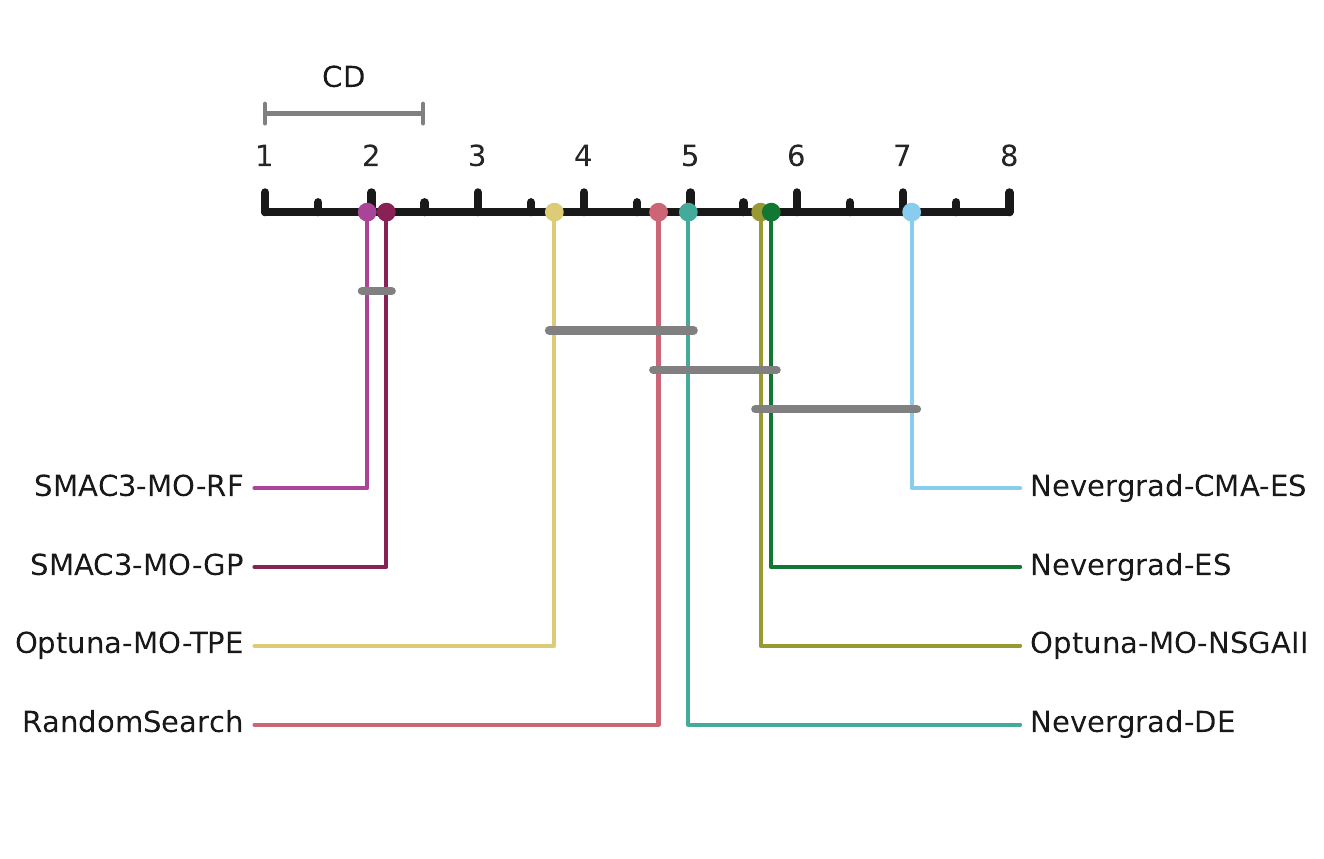}
        \caption{Multi-objective task type (test set)}
        \label{fig:mo_test_cd}
    \end{subfigure}\\
    \begin{subfigure}{0.4\linewidth}
        \centering
        \includegraphics[width=\linewidth]{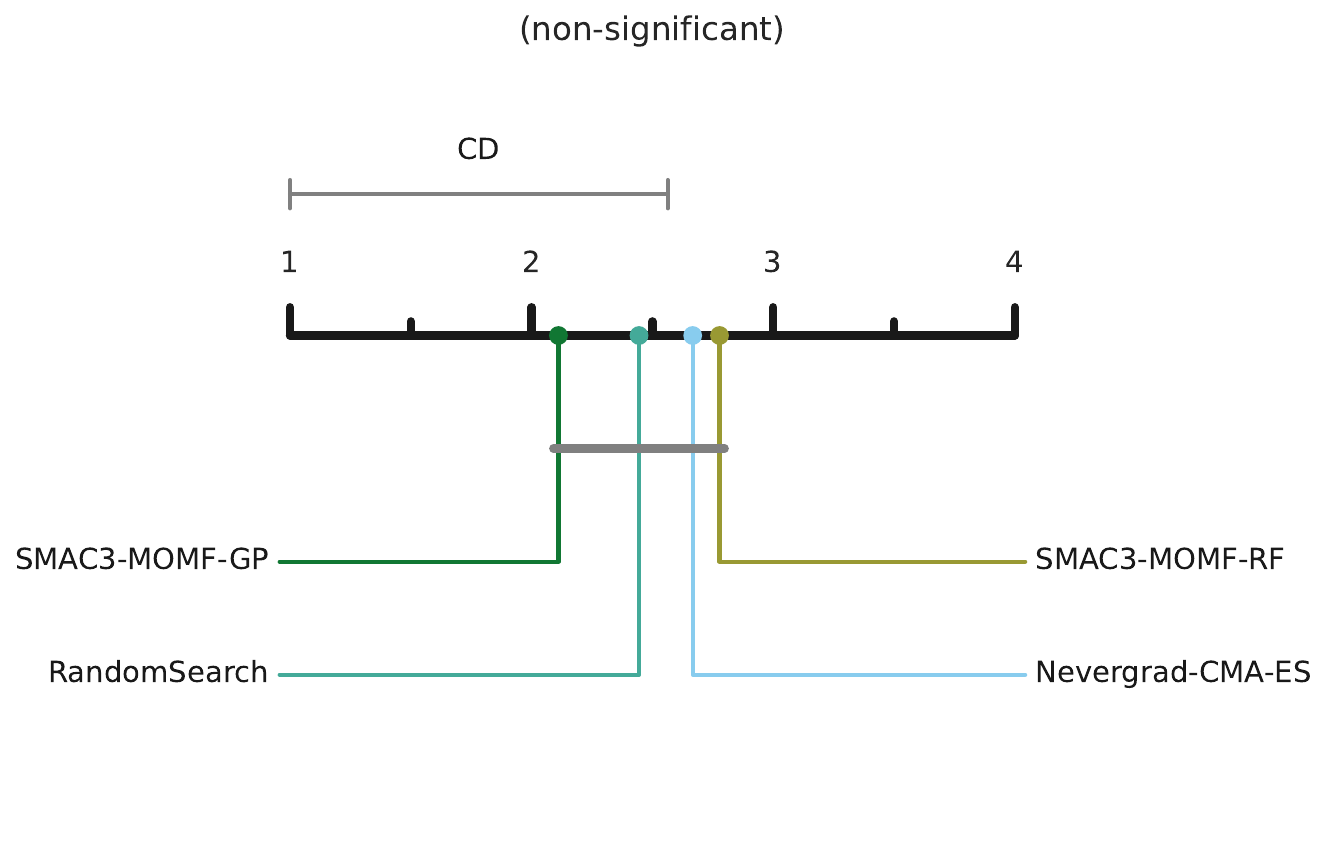}
        \caption{Multi-fidelity-objective task type (test set)}
        \label{fig:momf_test_cd}
    \end{subfigure}
    \caption{Critical Difference Diagrams}
    \label{fig:main_additional_results}
\end{figure}

    For each task type, many optimizers are not statistically significantly different.
    Their performance does differ on different tasks, indicating a possible complementarity.
    For example, in the black-box test set on BBOB task (function id 11, dimension 32), Nevergrad-CMA-ES excels whilst otherwise performing mediocre (see~\cref{fig:bb_test_perfpertask}).
    Critical difference plots for all task type (subset test) are found in \cref{fig:bb_test_cd,fig:mf_test_cd,fig:mo_test_cd,fig:momf_test_cd}.
    Sometimes, for the black-box task type, the order of ranks differs across the dev and test set.
    However, the distance of the ranks in those cases is lower than the critical difference, thus the absolute rank and the order thereof should be taken with a grain of salt.
    Upon inspection of the composition of the task type subsets, there is a mixture of easy and harder task instances, and optimizers perform in general similarly on them.
    On the topic of anytime performance, the rankings mostly stabilize after 60\%-80\% of the optimization budget, sometimes with great differences to the final rank in the first few trials.
    
    Our recommendation for research on optimization algorithms is the following:
    \begin{inparaenum}[(i)]
        \item define a task area where to advance (for example very high-dimensional black-box optimization or specific application domains),
        \item develop in this task area,
        \item verify general performance with a proposed subset,
        \item consider different ways to inspect results, and
        \item carefully report for the task area and the general case without generalizing conclusions.
    \end{inparaenum}

\section{Limitations and Future Work}
\label{sec:limitations_and_future_work}
So far, our benchmarking framework \carps focuses on benchmarking \ac{HPO}.
In principle, complete AutoML for pipelines construction could be benchmarked as well~\citep{olson-automl19a,feurer-jmlr22a}.
A limitation is that \carps depends on configuration spaces that can be expressed via \code{ConfigSpace} \citep{lindauer-arxiv19a} and thus is not compatible with unbounded configuration spaces defined by grammars or generic pipelines.
Motivated by their theoretically proven advantages in Monte Carlo methods, we have used in this work the $L_{\infty}$ star discrepancy as criterion for the selection of representative instances. Many other diversity metrics exists and could possibly be considered. For example, a property of the $L_{\infty}$ star discrepancy that we currently do not know how to fairly assess is its tendency to select more points in the upper right corner. Recent work has shown that this bias can be avoided at almost no cost by considering symmetrized versions of the $L_{\infty}$ star discrepancy notion~\citep{Clement-arxiv23}. To compare such alternative diversity measures for our benchmarking purposes, we would need to have efficient subset selection methods which are currently not available.  

In the future, we plan to extend our framework to include more optimizers and benchmarks steadily.
Furthermore, we would like to extend our task types to include constraints and plan to integrate a parallel execution system, allowing fair benchmarking, which is non-trivial, as proposed by~\cite{watanabe-automl24a}.
This is relevant as some optimizers, like ASHA~\citep{li-mlsys20a}, only then unleash their full potential.
In addition, \carps, a framework holding different benchmark collections, can be a stepping stone to active benchmarking, i.e.,
instead of evaluating an algorithm on all available tasks, the tasks are selected actively in order to build a holistic and nuanced view of the algorithm's performance, also regarding different task types.
Our work currently uses algorithm performances to span a task space, and we plan to contrast that to also using meta-features describing the tasks, possibly combining both~\citep{sim-ppsn22a}.
This would enable characterizing the strengths and weaknesses of optimizers on specific task types.

\section{Conclusion}
With \carps, we propose a lightweight benchmarking framework for HPO.
It offers numerous benchmarks and optimizers as baselines and is conceptualized to facilitate extension and scalability.
We target four task types, namely blackbox, multi-fidelity, multi-objective and multi-fidelity-multi-objective.
For those task types, we propose an initial subselection of representative benchmarking tasks, one for development and one to test an optimizer's performance, along with the inclusion of functionality to re-compute these subsets as more benchmarks become available.
Together with an analysis pipeline, \carps offers everything needed to develop new HPO optimizers.

\subsection*{Broader Impact}
\label{sec:broader_impact}
We do not foresee direct negative societal impact from our work since we do not target specific applications.
Overall, AutoML and HPO aims to democratize ML and AI further, which comes with all the general benefits and risks of making AI available to everyone.
In general, \carps eases research and computational burden of developing optimizers for HPO, and eventually contributes to better scientific practices.

    

\bibliographystyle{plainnat}
\bibliography{bib/strings,bib/lib,bib/custom,bib/proc}

\begin{thebibliography}{75}
\providecommand{\natexlab}[1]{#1}
\providecommand{\url}[1]{\texttt{#1}}
\expandafter\ifx\csname urlstyle\endcsname\relax
  \providecommand{\doi}[1]{doi: #1}\else
  \providecommand{\doi}{doi: \begingroup \urlstyle{rm}\Url}\fi

\bibitem[Akiba et~al.(2019)Akiba, Sano, Yanase, Ohta, and Koyama]{akiba-kdd19a}
T.~Akiba, S.~Sano, T.~Yanase, T.~Ohta, and M.~Koyama.
\newblock Optuna: A next-generation {H}yperparameter {O}ptimization framework.
\newblock In A.~Teredesai, V.~Kumar, Y.~Li, R.~Rosales, E.~Terzi, and G.~Karypis, editors, \emph{Proceedings of the 25th {ACM} {SIGKDD} International Conference on Knowledge Discovery {\&} Data Mining ({KDD}'19)}, pages 2623--2631. ACM Press, 2019.

\bibitem[Awad et~al.(2021)Awad, Mallik, and Hutter]{awad-ijcai21a}
N.~Awad, N.~Mallik, and F.~Hutter.
\newblock {DEHB}: Evolutionary hyperband for scalable, robust and efficient {H}yperparameter {O}ptimization.
\newblock In Z.~Zhou, editor, \emph{Proceedings of the 30th International Joint Conference on Artificial Intelligence ({IJCAI}'21)}, pages 2147--2153, 2021.

\bibitem[Balandat et~al.(2020)Balandat, Karrer, Jiang, Daulton, Letham, Wilson, and Bakshy]{balandat-neurips20a}
M.~Balandat, B.~Karrer, D.~Jiang, S.~Daulton, B.~Letham, A.~Wilson, and E.~Bakshy.
\newblock Botorch: A framework for efficient monte-carlo {Bayesian} optimization.
\newblock In H.~Larochelle, M.~Ranzato, R.~Hadsell, M.-F. Balcan, and H.~Lin, editors, \emph{Proceedings of the 34th International Conference on Advances in Neural Information Processing Systems ({N}eur{IPS}'20)}. Curran Associates, 2020.

\bibitem[Benjamins et~al.(2023)Benjamins, Raponi, Jankovic, Doerr, and Lindauer]{benjamins-automl23a}
C.~Benjamins, E.~Raponi, A.~Jankovic, C.~Doerr, and M.~Lindauer.
\newblock Self-adjusting weighted expected improvement for bayesian optimization.
\newblock In A.~Faust, C.~White, F.~Hutter, R.~Garnett, and J.~Gardner, editors, \emph{Proceedings of the Second International Conference on Automated Machine Learning}. Proceedings of Machine Learning Research, 2023.

\bibitem[Bergstra and Bengio(2012)]{bergstra-jmlr12a}
J.~Bergstra and Y.~Bengio.
\newblock Random search for hyper-parameter optimization.
\newblock \emph{Journal of Machine Learning Research}, 13:\penalty0 281--305, 2012.

\bibitem[Bergstra et~al.(2011)Bergstra, Bardenet, Bengio, and K{\'e}gl]{bergstra-nips11a}
J.~Bergstra, R.~Bardenet, Y.~Bengio, and B.~K{\'e}gl.
\newblock Algorithms for hyper-parameter optimization.
\newblock In J.~Shawe-Taylor, R.~Zemel, P.~Bartlett, F.~Pereira, and K.~Weinberger, editors, \emph{Proceedings of the 25th International Conference on Advances in Neural Information Processing Systems ({N}eur{IPS}'11)}, pages 2546--2554. Curran Associates, 2011.

\bibitem[Bergstra et~al.(2013)Bergstra, Yamins, and Cox]{bergstra-icml13a}
J.~Bergstra, D.~Yamins, and D.~Cox.
\newblock Making a science of model search: {H}yperparameter {O}ptimization in hundreds of dimensions for vision architectures.
\newblock In S.~Dasgupta and D.~McAllester, editors, \emph{Proceedings of the 30th International Conference on Machine Learning ({ICML}'13)}, pages 115--123. Omnipress, 2013.

\bibitem[Bischl et~al.(2021)Bischl, Casalicchio, Feurer, Hutter, Lang, Mantovani, van Rijn, and Vanschoren]{bischl-neuripsdbt21a}
B.~Bischl, G.~Casalicchio, M.~Feurer, F.~Hutter, M.~Lang, R.~Mantovani, J.~van Rijn, and J.~Vanschoren.
\newblock {OpenML} benchmarking suites.
\newblock In  \citet{neuripsdbt21}.

\bibitem[Bischl et~al.(2023)Bischl, Binder, Lang, Pielok, Richter, Coors, Thomas, Ullmann, Becker, Boulesteix, Deng, and Lindauer]{bischl-dmkd23a}
B.~Bischl, M.~Binder, M.~Lang, T.~Pielok, J.~Richter, S.~Coors, J.~Thomas, T.~Ullmann, M.~Becker, A.{-}L. Boulesteix, D.~Deng, and M.~Lindauer.
\newblock Hyperparameter optimization: Foundations, algorithms, best practices, and open challenges.
\newblock \emph{Wiley Interdisciplinary Reviews: Data Mining and Knowledge Discovery}, page e1484, 2023.

\bibitem[{Blank} and {Deb}(2020)]{blank-ieeeaccess20}
J.~{Blank} and K.~{Deb}.
\newblock pymoo: Multi-objective optimization in python.
\newblock \emph{IEEE Access}, 8:\penalty0 89497--89509, 2020.

\bibitem[Bousquet et~al.(2017)Bousquet, Gelly, Kurach, Teytaud, and Vincent]{BousquetGKTV17}
O.~Bousquet, S.~Gelly, K.~Kurach, O.~Teytaud, and D.~Vincent.
\newblock Critical hyper-parameters: No random, no cry.
\newblock \emph{arXiv:1706.03200 [cs.LG]}, 2017.

\bibitem[Cardoso et~al.(2021)Cardoso, Santos, Franc{\^e}s, Prud{\^e}ncio, and Alves]{cardoso-arxiv21a}
L.~F. Cardoso, V.~C.A. Santos, R.~Franc{\^e}s, R.~Prud{\^e}ncio, and R.~Alves.
\newblock Data vs classifiers, who wins?
\newblock \emph{arXiv:2107.07451 [cs.LG]}, 2021.

\bibitem[Cauwet et~al.(2020)Cauwet, Couprie, Dehos, Luc, Rapin, Rivi{\`{e}}re, Teytaud, Teytaud, and Usunier]{CauwetCDLRRTTU20}
M.{-}L. Cauwet, C.~Couprie, J.~Dehos, P.~Luc, J.~Rapin, M.~Rivi{\`{e}}re, F.~Teytaud, O.~Teytaud, and N.~Usunier.
\newblock Fully parallel hyperparameter search: Reshaped space-filling.
\newblock In H.~{Daume III} and A.~Singh, editors, \emph{Proceedings of the 37th International Conference on Machine Learning ({ICML}'20)}, volume~98, pages 1338--1348. Proceedings of Machine Learning Research, 2020.

\bibitem[Cenikj et~al.(2022)Cenikj, Lang, Engelbrecht, Doerr, Korosec, and Eftimov]{cenikj-gecco22a}
G.~Cenikj, R.~Dieter Lang, A.~Engelbrecht, C.~Doerr, P.~Korosec, and T.~Eftimov.
\newblock {SELECTOR:} selecting a representative benchmark suite for reproducible statistical comparison.
\newblock In J.~Fieldsend, editor, \emph{Proceedings of the Genetic and Evolutionary Computation Conference ({GECCO}'22)}. ACM Press, 2022.

\bibitem[Cl\'{e}ment et~al.(2022)Cl\'{e}ment, Doerr, and Paquete]{CDP22}
F.~Cl\'{e}ment, C.~Doerr, and L.~Paquete.
\newblock Star discrepancy subset selection: {P}roblem formulation and efficient approaches for low dimensions.
\newblock \emph{Journal of Complexity}, 70:\penalty0 101645, 2022.

\bibitem[Cl{\'{e}}ment et~al.(2023)Cl{\'{e}}ment, Doerr, Klamroth, and Paquete]{Clement-arxiv23}
F.~Cl{\'{e}}ment, C.~Doerr, K.~Klamroth, and L.~Paquete.
\newblock Constructing optimal l\({}_{\mbox{{\(\infty\)}}}\) star discrepancy sets.
\newblock \emph{arXiv:2311.17463 [(cs.CG)]}, 2023.

\bibitem[Cl\'{e}ment et~al.(2024)Cl\'{e}ment, Doerr, and Paquete]{CDP23}
F.~Cl\'{e}ment, C.~Doerr, and L.~Paquete.
\newblock Heuristic approaches to obtain low-discrepancy point sets via subset selection.
\newblock \emph{Journal of Complexity}, 83:\penalty0 101852, 2024.

\bibitem[Cowen-Rivers et~al.(2022)Cowen-Rivers, Lyu, Tutunov, Wang, Grosnit, Griffiths, Maraval, Jianye, Wang, Peters, and Ammar]{cowenrivers-jair22a}
A.~Cowen-Rivers, W.~Lyu, R.~Tutunov, Z.~Wang, A.~Grosnit, R.~Griffiths, A.~Maraval, H.~Jianye, J.~Wang, J.~Peters, and H.~Ammar.
\newblock {HEBO}: Pushing the limits of sample-efficient hyper-parameter optimisation.
\newblock \emph{Journal of Artificial Intelligence Research}, 74:\penalty0 1269--1349, 2022.

\bibitem[Demšar(2006)]{demsar-06a}
J.~Demšar.
\newblock Statistical comparisons of classifiers over multiple data sets.
\newblock \emph{Journal of Machine Learning Research}, 7:\penalty0 1--30, 2006.

\bibitem[Dick and Pillichshammer(2010)]{DickP10}
J.~Dick and F.~Pillichshammer.
\newblock \emph{Digital Nets and Sequences}.
\newblock Cambridge University Press, Cambridge, 2010.

\bibitem[Dietrich et~al.(2024)Dietrich, Vermetten, Doerr, and Kerschke]{dietrich-gecco24}
K.~Dietrich, D.~Vermetten, C.~Doerr, and P.~Kerschke.
\newblock Impact of training instance selection on automated algorithm selection models for numerical black-box optimization.
\newblock \emph{arXiv:2404.07539 [cs.NE]}, 2024.

\bibitem[Eftimov et~al.(2022)Eftimov, Petelin, Cenikj, Kostovska, Ispirova, Koro{\v{s}}ec, and Bogatinovski]{eftimov-esa22a}
T.~Eftimov, G.~Petelin, G.~Cenikj, A.~Kostovska, G.~Ispirova, P.~Koro{\v{s}}ec, and J.~Bogatinovski.
\newblock Less is more: Selecting the right benchmarking set of data for time series classification.
\newblock \emph{Expert Systems with Applications}, 198:\penalty0 116871, 2022.

\bibitem[Eggensperger et~al.(2015)Eggensperger, Hutter, Hoos, and Leyton-Brown]{eggensperger-aaai15a}
K.~Eggensperger, F.~Hutter, H.~Hoos, and K.~Leyton-Brown.
\newblock Efficient benchmarking of hyperparameter optimizers via surrogates.
\newblock In B.~Bonet and S.~Koenig, editors, \emph{Proceedings of the Twenty-ninth {AAAI} Conference on Artificial Intelligence ({AAAI}'15)}, pages 1114--1120. {AAAI} Press, 2015.

\bibitem[Eggensperger et~al.(2019)Eggensperger, Lindauer, and Hutter]{eggensperger-jair19a}
K.~Eggensperger, M.~Lindauer, and F.~Hutter.
\newblock Pitfalls and best practices in algorithm configuration.
\newblock \emph{Journal of Artificial Intelligence Research}, pages 861--893, 2019.

\bibitem[Eggensperger et~al.(2021)Eggensperger, M{\"u}ller, Mallik, Feurer, Sass, Klein, Awad, Lindauer, and Hutter]{eggensperger-neuripsdbt21a}
K.~Eggensperger, P.~M{\"u}ller, N.~Mallik, M.~Feurer, R.~Sass, A.~Klein, N.~Awad, M.~Lindauer, and F.~Hutter.
\newblock {HPOBench}: A collection of reproducible multi-fidelity benchmark problems for {HPO}.
\newblock In  \citet{neuripsdbt21}.

\bibitem[Falkner et~al.(2018)Falkner, Klein, and Hutter]{falkner-icml18a}
S.~Falkner, A.~Klein, and F.~Hutter.
\newblock {BOHB}: Robust and efficient {H}yperparameter {O}ptimization at scale.
\newblock In J.~Dy and A.~Krause, editors, \emph{Proceedings of the 35th International Conference on Machine Learning ({ICML}'18)}, volume~80, pages 1437--1446. Proceedings of Machine Learning Research, 2018.

\bibitem[Feurer and Hutter(2019)]{feurer-automlbook19a}
M.~Feurer and F.~Hutter.
\newblock Hyperparameter {O}ptimization.
\newblock In  \citet{hutter-book19a}, chapter~1, pages 3 -- 38.
\newblock Available for free at \url{http://automl.org/book}.

\bibitem[Feurer et~al.(2022)Feurer, Eggensperger, Falkner, Lindauer, and Hutter]{feurer-jmlr22a}
M.~Feurer, K.~Eggensperger, S.~Falkner, M.~Lindauer, and F.~Hutter.
\newblock {Auto-Sklearn} 2.0: Hands-free automl via meta-learning.
\newblock \emph{Journal of Machine Learning Research}, 23\penalty0 (261):\penalty0 1--61, 2022.

\bibitem[Galanti and Jung(1997)]{GalFin}
S.~Galanti and A.~Jung.
\newblock Low-discrepancy sequences: Monte carlo simulation of option prices.
\newblock \emph{Journal of derivatives}, 5\penalty0 (1):\penalty0 63--83, 1997.

\bibitem[Garnett(2023)]{garnett-book23a}
R.~Garnett.
\newblock \emph{{Bayesian Optimization}}.
\newblock Cambridge University Press, 2023.
\newblock Available for free at \url{https://bayesoptbook.com/}.

\bibitem[Guyon et~al.(2022)Guyon, Lindauer, van~der Schaar, Hutter, and Garnett]{automlconf22}
I.~Guyon, M.~Lindauer, M.~van~der Schaar, F.~Hutter, and R.~Garnett, editors.
\newblock \emph{Proceedings of the First International Conference on Automated Machine Learning}, 2022. Proceedings of Machine Learning Research.

\bibitem[Halton(1964)]{Halton64}
J.H. Halton.
\newblock {Algorithm 247: Radical-Inverse Quasi-random Point Sequence}.
\newblock \emph{{Communications of the ACM}}, 7\penalty0 (12):\penalty0 701~--~702, 1964.

\bibitem[Hansen et~al.(2003)Hansen, M{\"{u}}ller, and Koumoutsakos]{hansen-ec03a}
N.~Hansen, S.~D. M{\"{u}}ller, and P.~Koumoutsakos.
\newblock Reducing the time complexity of the derandomized evolution strategy with covariance matrix adaptation {(CMA-ES)}.
\newblock \emph{Evolutionary Computing}, 11\penalty0 (1):\penalty0 1--18, 2003.

\bibitem[Hansen et~al.(2019)Hansen, Akimoto, and Baudis]{hansen-2019a}
N.~Hansen, Y.~Akimoto, and P.~Baudis.
\newblock {CMA-ES/pycma} on {G}ithub, 2019.

\bibitem[Hansen et~al.(2020)Hansen, Auger, Ros, Mersman, Tu{\v s}ar, and Brockhoff]{hansen-oms20a}
N.~Hansen, A.~Auger, R.~Ros, O.~Mersman, T.~Tu{\v s}ar, and D.~Brockhoff.
\newblock {COCO}: A platform for comparing continuous optimizers in a black-box setting.
\newblock \emph{Optimization Methods and Software}, 2020.

\bibitem[Herbold(2020)]{herbold-joss20}
S.~Herbold.
\newblock Autorank: A python package for automated ranking of classifiers.
\newblock \emph{Journal of Open Source Software}, 5\penalty0 (48):\penalty0 2173, 2020.

\bibitem[Hutter et~al.(2019)Hutter, Kotthoff, and Vanschoren]{hutter-book19a}
F.~Hutter, L.~Kotthoff, and J.~Vanschoren, editors.
\newblock \emph{Automated Machine Learning: Methods, Systems, Challenges}.
\newblock Springer, 2019.
\newblock Available for free at \url{http://automl.org/book}.

\bibitem[Ispirova et~al.(2024)Ispirova, Eftimov, D{\v{z}}eroski, and Seljak]{ispirova-esa24a}
G.~Ispirova, T.~Eftimov, S.~D{\v{z}}eroski, and B.~Seljak.
\newblock Msgen: Measuring generalization of nutrient value prediction across different recipe datasets.
\newblock \emph{Expert Systems with Applications}, 237:\penalty0 121507, 2024.

\bibitem[Jamieson and Talwalkar(2016)]{jamieson-aistats16a}
K.~Jamieson and A.~Talwalkar.
\newblock Non-stochastic best arm identification and {H}yperparameter {O}ptimization.
\newblock In A.~Gretton and C.~Robert, editors, \emph{Proceedings of the Seventeenth International Conference on Artificial Intelligence and Statistics ({AISTATS}'16)}, volume~51. Proceedings of Machine Learning Research, 2016.

\bibitem[Jones et~al.(1998)Jones, Schonlau, and Welch]{jones-jgo98a}
D.~Jones, M.~Schonlau, and W.~Welch.
\newblock Efficient global optimization of expensive black box functions.
\newblock \emph{Journal of Global Optimization}, 13:\penalty0 455--492, 1998.

\bibitem[Knowles(2006)]{knowls-evoco06a}
J.~Knowles.
\newblock {ParEGO}: a hybrid algorithm with on-line landscape approximation for expensive multiobjective optimization problems.
\newblock \emph{{IEEE} Transactions on Evolutionary Computation}, 10\penalty0 (1):\penalty0 50--66, 2006.

\bibitem[Li et~al.(2018)Li, Jamieson, DeSalvo, Rostamizadeh, and Talwalkar]{li-jmlr18a}
L.~Li, K.~Jamieson, G.~DeSalvo, A.~Rostamizadeh, and A.~Talwalkar.
\newblock Hyperband: A novel bandit-based approach to {H}yperparameter {O}ptimization.
\newblock \emph{Journal of Machine Learning Research}, 18\penalty0 (185):\penalty0 1--52, 2018.

\bibitem[Li et~al.(2020)Li, Jamieson, Rostamizadeh, Gonina, Ben{-}tzur, Hardt, Recht, and Talwalkar]{li-mlsys20a}
L.~Li, K.~Jamieson, A.~Rostamizadeh, E.~Gonina, J.~Ben{-}tzur, M.~Hardt, B.~Recht, and A.~Talwalkar.
\newblock A system for massively parallel hyperparameter tuning.
\newblock In I.~Dhillon, D.~Papailiopoulos, and V.~Sze, editors, \emph{Proceedings of Machine Learning and Systems 2}, volume~2, 2020.

\bibitem[Lindauer et~al.(2019)Lindauer, Eggensperger, Feurer, Biedenkapp, Marben, M\"uller, and Hutter]{lindauer-arxiv19a}
M.~Lindauer, K.~Eggensperger, M.~Feurer, A.~Biedenkapp, J.~Marben, P.~M\"uller, and F.~Hutter.
\newblock {BOAH}: A tool suite for {M}ulti-fidelity {Bayesian} {O}ptimization \& analysis of hyperparameters.
\newblock \emph{arXiv:1908.06756 [cs.LG]}, 2019.

\bibitem[Lindauer et~al.(2022)Lindauer, Eggensperger, Feurer, Biedenkapp, Deng, Benjamins, Ruhkopf, Sass, and Hutter]{lindauer-jmlr22a}
M.~Lindauer, K.~Eggensperger, M.~Feurer, A.~Biedenkapp, D.~Deng, C.~Benjamins, T.~Ruhkopf, R.~Sass, and F.~Hutter.
\newblock {SMAC3}: A versatile bayesian optimization package for {H}yperparameter {O}ptimization.
\newblock \emph{Journal of Machine Learning Research}, 23\penalty0 (54):\penalty0 1--9, 2022.

\bibitem[Long et~al.(2023)Long, Vermetten, van Stein, and Kononova]{long-evostar23a}
F.~Long, D.~Vermetten, B.~van Stein, and A.~Kononova.
\newblock {BBOB} instance analysis: Landscape properties and algorithm performance across problem instances.
\newblock In \emph{International Conference on the Applications of Evolutionary Computation (Part of EvoStar)}, pages 380--395. Springer, 2023.

\bibitem[Mallik et~al.(2023)Mallik, Hvarfner, Bergman, Stoll, Janowski, Lindauer, Nardi, and Hutter]{mallik-neurips23a}
N.~Mallik, C.~Hvarfner, E.~Bergman, D.~Stoll, M.~Janowski, M.~Lindauer, L.~Nardi, and F.~Hutter.
\newblock {PriorBand}: Practical hyperparameter optimization in the age of deep learning.
\newblock In  \citet{neurips23}.

\bibitem[Mart\'{\i}nez-Plumed et~al.(2016)Mart\'{\i}nez-Plumed, Prud\^{e}ncio, Mart\'{\i}nez-Us\'{o}, and Hern\'{a}ndez-Orallo]{martinezplumed-ecai16a}
F.~Mart\'{\i}nez-Plumed, R.~Prud\^{e}ncio, A.~Mart\'{\i}nez-Us\'{o}, and J.~Hern\'{a}ndez-Orallo.
\newblock Making sense of item response theory in machine learning.
\newblock In \emph{Proceedings of the Twenty-Second European Conference on Artificial Intelligence}, page 1140–1148, 2016.

\bibitem[McElfresh et~al.(2023)McElfresh, Khandagale, Valverde, {Prasad C}, Ramakrishnan, Goldblum, and White]{mcelfresh-neurips23a}
D.~McElfresh, S.~Khandagale, J.~Valverde, V.~{Prasad C}, G.~Ramakrishnan, M.~Goldblum, and C.~White.
\newblock When do neural nets outperform boosted trees on tabular data?
\newblock In  \citet{neurips23}, pages 76336--76369.

\bibitem[Mockus(1989)]{mockus-bo89a}
J.~Mockus.
\newblock \emph{{Bayesian} Approach to Global Optimization. Theory and Applications}.
\newblock Kluwer Academic Publishers, 1989.

\bibitem[Neumann et~al.(2018)Neumann, Gao, Doerr, Neumann, and Wagner]{NeumannGDN018}
A.~Neumann, W.~Gao, C.~Doerr, F.~Neumann, and M.~Wagner.
\newblock Discrepancy-based evolutionary diversity optimization.
\newblock In H.~Aguirre, editor, \emph{Proceedings of the Genetic and Evolutionary Computation Conference ({GECCO}'18)}, pages 991--998. ACM Press, 2018.

\bibitem[Niederreiter(1972)]{NieBox}
H.~Niederreiter.
\newblock Discrepancy and convex programming.
\newblock \emph{Annali di Matematica Pura ed Applicata}, 93:\penalty0 89--97, 1972.

\bibitem[Nikolikj et~al.(2023)Nikolikj, Pluháček, Doerr, Korošec, and Eftimov]{nikolikj-cec23a}
A.~Nikolikj, M.~Pluháček, C.~Doerr, P.~Korošec, and T.~Eftimov.
\newblock Sensitivity analysis of rf+clust for leave-one-problem-out performance prediction.
\newblock In \emph{2023 IEEE Congress on Evolutionary Computation (CEC)}, pages 1--8, 2023.

\bibitem[Nogueira(2014)]{nogueira-2014a}
F.~Nogueira.
\newblock {Bayesian Optimization}: Open source constrained global optimization tool for {Python}, 2014.

\bibitem[Oh et~al.(2023)Oh, Naumann, Globerson, Saenko, Hardt, and Levine]{neurips23}
A.~Oh, T.~Naumann, A.~Globerson, K.~Saenko, M.~Hardt, and S.~Levine, editors.
\newblock \emph{Proceedings of the 37th International Conference on Advances in Neural Information Processing Systems ({N}eur{IPS}'23)}, 2023. Curran Associates.

\bibitem[Olson and Moore(2019)]{olson-automl19a}
R.~Olson and J.~Moore.
\newblock {TPOT}: A tree-based pipeline optimization tool for automating machine learning.
\newblock In  \citet{hutter-book19a}, pages 151--160.
\newblock Available for free at \url{http://automl.org/book}.

\bibitem[Paulin et~al.(2022)Paulin, Bonneel, Coeurjolly, Iehl, Keller, and Ostromoukhov]{MatBuilder}
L.~Paulin, N.~Bonneel, D.~Coeurjolly, J.-C. Iehl, A.~Keller, and V.~Ostromoukhov.
\newblock Matbuilder: Mastering sampling uniformity over projections.
\newblock \emph{ACM Transactions on Graphics (TOG)}, 41\penalty0 (4):\penalty0 1--13, 2022.

\bibitem[Pereira et~al.(2024)Pereira, Smith-Miles, Muñoz, and Lorena]{pereira-dmkd24a}
J.~Pereira, K.~Smith-Miles, M.~Muñoz, and A.~Lorena.
\newblock Optimal selection of benchmarking datasets for unbiased machine learning algorithm evaluation.
\newblock \emph{Data Mining and Knowledge Discovery}, 38:\penalty0 461--500, 2024.

\bibitem[Pfisterer et~al.(2022)Pfisterer, Schneider, Moosbauer, Binder, and Bischl]{pfisterer-automl22a}
F.~Pfisterer, L.~Schneider, J.~Moosbauer, M.~Binder, and B.~Bischl.
\newblock {YAHPO Gym} -- an efficient multi-objective multi-fidelity benchmark for hyperparameter optimization.
\newblock In  \citet{automlconf22}.

\bibitem[Pineda~Arango et~al.(2021)Pineda~Arango, Jomaa, Wistuba, and Grabocka]{pineda-neuripsdbt21a}
S.~Pineda~Arango, H.~Jomaa, M.~Wistuba, and J.~Grabocka.
\newblock {HPO-B}: A large-scale reproducible benchmark for black-box {HPO} based on {OpenML}.
\newblock In  \citet{neuripsdbt21}.

\bibitem[Rapin and Teytaud(2018)]{rapin-2018a}
J.~Rapin and O.~Teytaud.
\newblock {Nevergrad - A Gradient-free Optimization platform}, 2018.

\bibitem[Salinas et~al.(2022)Salinas, Seeger, Klein, Perrone, Wistuba, and Archambeau]{salinas-automl2022}
D.~Salinas, M.~Seeger, A.~Klein, V.~Perrone, M.~Wistuba, and C.~Archambeau.
\newblock Syne {T}une: A library for large scale hyperparameter tuning and reproducible research.
\newblock In  \citet{automlconf22}.

\bibitem[Salinas et~al.(2023)Salinas, Golebiowski, Klein, Seeger, and Archambeau]{salinas-icml23a}
D.~Salinas, J.~Golebiowski, A.~Klein, M.~Seeger, and C.~Archambeau.
\newblock Optimizing hyperparameters with conformal quantile regression.
\newblock In A.~Krause, E.~Brunskill, K.~Cho, B.~Engelhardt, S.~Sabato, and J.~Scarlett, editors, \emph{Proceedings of the 40th International Conference on Machine Learning ({ICML}'23)}, volume 202 of \emph{Proceedings of Machine Learning Research}. PMLR, 2023.

\bibitem[Santner et~al.(2003)Santner, Williams, and Notz]{santner-book03a}
T.~Santner, B.~Williams, and W.~Notz.
\newblock \emph{The design and analysis of computer experiments}.
\newblock Springer, 2003.

\bibitem[Sim and Hart(2022)]{sim-ppsn22a}
K.~Sim and E.~Hart.
\newblock Evolutionary approaches to improving the layouts of instance-spaces.
\newblock In G{\"u}nter Rudolph, Anna~V. Kononova, Hern{\'a}n Aguirre, Pascal Kerschke, Gabriela Ochoa, and Tea Tu{\v{s}}ar, editors, \emph{Parallel Problem Solving from Nature -- PPSN XVII}. Springer International Publishing, 2022.

\bibitem[Snoek et~al.(2012)Snoek, Larochelle, and Adams]{snoek-nips12a}
J.~Snoek, H.~Larochelle, and R.~Adams.
\newblock Practical {B}ayesian optimization of machine learning algorithms.
\newblock In P.~Bartlett, F.~Pereira, C.~Burges, L.~Bottou, and K.~Weinberger, editors, \emph{Proceedings of the 26th International Conference on Advances in Neural Information Processing Systems ({N}eur{IPS}'12)}, pages 2960--2968. Curran Associates, 2012.

\bibitem[Sobol(1967)]{Sobol}
I.M. Sobol.
\newblock {On the Distribution of Points in a Cube and the Approximate Evaluation of Integrals}.
\newblock \emph{{USSR Computational Mathematics and Mathematical Physics}}, 7\penalty0 (4):\penalty0 86~--~112, 1967.

\bibitem[Storn and Price(1997)]{storn-jgo97a}
R.~Storn and K.~Price.
\newblock Differential evolution – a simple and efficient heuristic for global optimization over continuous spaces.
\newblock \emph{Journal of Global Optimization}, 11:\penalty0 341--359, 1997.

\bibitem[Tiao et~al.(2021)Tiao, Klein, Seeger, Bonilla, Archambeau, and Ramos]{tiao-icml21a}
L.~Tiao, A.~Klein, M.~Seeger, E.~Bonilla, C.~Archambeau, and F.~Ramos.
\newblock {BORE: Bayesian} optimization by density-ratio estimation.
\newblock In M.~Meila and T.~Zhang, editors, \emph{Proceedings of the 38th International Conference on Machine Learning ({ICML}'21)}, volume 139 of \emph{Proceedings of Machine Learning Research}, pages 10289--10300. PMLR, 2021.

\bibitem[Turner and Eriksson(2019)]{bayesmark}
R.~Turner and D.~Eriksson.
\newblock {Bayesmark}: Benchmark framework to easily compare {Bayesian Optimization} methods on real machine learning tasks.
\newblock \url{github.com/uber/bayesmark}, 2019.

\bibitem[Turner et~al.(2021)Turner, Eriksson, McCourt, Kiili, Laaksonen, Xu, and Guyon]{turner-neuripscomp21a}
R.~Turner, D.~Eriksson, M.~McCourt, J.~Kiili, E.~Laaksonen, Z.~Xu, and I.~Guyon.
\newblock Bayesian optimization is superior to random search for machine learning hyperparameter tuning: Analysis of the {Black-Box Optimization Challenge 2020}.
\newblock In H.~Escalante and K.~Hofmann, editors, \emph{Proceedings of the Neural Information Processing Systems Track Competition and Demonstration}, pages 3--26. Curran Associates, 2021.

\bibitem[Vanschoren and Yeung(2021)]{neuripsdbt21}
J.~Vanschoren and S.~Yeung, editors.
\newblock \emph{Proceedings of the Neural Information Processing Systems Track on Datasets and Benchmarks}, 2021. Curran Associates.

\bibitem[Vermetten et~al.(2023)Vermetten, Ye, B{\"{a}}ck, and Doerr]{vermetten-automl23a}
D.~Vermetten, F.~Ye, T.~B{\"{a}}ck, and C.~Doerr.
\newblock {MA-BBOB:} many-affine combinations of {BBOB} functions for evaluating automl approaches in noiseless numerical black-box optimization contexts.
\newblock In \emph{International Conference on Automated Machine Learning}, pages 7--1. PMLR, 2023.

\bibitem[Watanabe et~al.(2024)Watanabe, Mallik, Bergman, and Hutter]{watanabe-automl24a}
S.~Watanabe, N.~Mallik, E.~Bergman, and F.~Hutter.
\newblock Fast benchmarking of asynchronous multi-fidelity optimization on zero-cost benchmarks.
\newblock In \emph{AutoML}, volume 256 of \emph{PMLR}, pages 14/1--18. {PMLR}, 2024.

\bibitem[Zela et~al.(2022)Zela, Siems, Zimmer, Lukasik, Keuper, and Hutter]{zela-iclr22a}
A.~Zela, J.~Siems, L.~Zimmer, J.~Lukasik, M.~Keuper, and F.~Hutter.
\newblock Surrogate {NAS} benchmarks: Going beyond the limited search spaces of tabular {NAS} benchmarks.
\newblock In \emph{The Tenth International Conference on Learning Representations ({ICLR}'22)}. ICLR, 2022.
\newblock Published online: \url{iclr.cc}.

\end{thebibliography}

\FloatBarrier
\newpage
\newpage

\section{Glossary}

\begin{longtable}{@{}p{0.2\linewidth} p{0.75\linewidth}@{}}
    \toprule
    \textbf{Term} & \textbf{Definition} \\
    \midrule
    \endfirsthead
    
    \toprule
    \textbf{Term} & \textbf{Definition} \\
    \midrule
    \endhead

    \bottomrule
    \endfoot
    
    \bottomrule
    \endlastfoot
    
    HPO &  Paradigm: Hyperparameter Optimization ~\citep{feurer-automlbook19a,bischl-dmkd23a}\\
    ML & Machine Learning \\
    BB &  Paradigm: Blackbox; an optimization task type in which solely inputs and outputs of the optimization object are available. \\
    MF & Paradigm: Multi-Fidelity; an optimization task type with cheaper approximations of the objective function available.\\
    MO &  Paradigm: Multi-Objective; an optimization task type, where more then one objective need to be optimized. \\
    MOMF &  Paradigm: Multi-Fidelity Multi-Objective optimization. \\
   
    HPOBench & Benchmark: Hyperparameter Optimization benchmark ~\citep{eggensperger-neuripsdbt21a}\\
    YAHPO-Gym & Benchmark: Yet Another Hyperparameter Optimization Gym ~\citep{pfisterer-automl22a}\\
    HPO-B & Benchmark: A Large-Scale Reproducible Benchmark for Black-Box HPO based on OpenML.~\citep{pineda-neuripsdbt21a} \\
    MFPBench  & Benchmark: A multi-fidleity prior benchmark. ~\citep{mallik-neurips23a} \\
    BBOB  & Benchmark: Black-box optimization benchmarking. ~\citep{hansen-oms20a}\\
    OpenML-CC18 & Benchmark: OpenML curated classification tasks. ~\citep{bischl-neuripsdbt21a} \\
    Pymoo-MO & Benchmark: Multi-objective Optimization in Python ~\citep{blank-ieeeaccess20}\\
    CARPS & Benchmark:  Comprehensive Automated Research Performance Studies (ours)\\
    
    ASHA & Optimizer:  Asynchronous Successive Halving ~\citep{li-mlsys20a}\\
    DEHB & Optimizer: Differential Evolution Hyperband ~\citep{awad-ijcai21a}\\
    BOHB & Optimizer: Bayesian Optimization Hyperband ~\citep{falkner-icml18a}\\
    HEBO & Optimizer: Heteroscedastic and Evolutionary Bayesian Optimisation solver~\citep{cowenrivers-jair22a} \\
    Skopt & Optimizer: Scikit Optimize \footnote{Scikit Optimize, \url{https://github.com/scikit-optimize/scikit-optimize}, 2018.}\\
    SMAC3 & Optimizer: Sequential Model-based Algorithm Configuration ~\citep{lindauer-jmlr22a} \\
    SMAC3-MOMF-GP & Optimizer: SMAC3 Multi-Objective Multi-fidelity Gaussian Process  \\
    Nevergrad-DE & Optimizer: Differential Evolution~\citep{storn-jgo97a} Strategy implemented in Nevergrad~\citep{rapin-2018a} \\
    Nevergrad-CMA-ES & Optimizer: CMA-ES~\citep{hansen-ec03a} Strategy implemented in Nevergrad~\citep{rapin-2018a} \\
    Optuna-MO  & Optimizer: Multi-objective optimization implemented in Optuna~\citep{akiba-kdd19a}\\

    CMA-ES & Covariance Matrix Adaptation Evolution Strategy\\
    CD & Critical difference. \\
    Lichtenberg-MATILDA & \\
    RQ & \href{https://python-rq.org/}{Redis Queue (RQ)} \\

\end{longtable}

\appendix

\section{Appendix}
The appendix is structured as follows.
We detail benchmarking frameworks in~\cref{sec:benchmark_frameworks}, optimizers variants in~\cref{sec:optimizer_overview_app} and the maintenance plan in~\cref{sec:maintenance_plan}.
Further, we discuss computational resources used (\cref{sec:computational_resources}) and technical details of the interface (\cref{sec:interface_technical_details}) and of the optimization of the star discrepancy (\cref{sec:star_disc_opt_desc}).
Details on the subselections per task type and results can be found in~\cref{sec:subselection_app} and~\cref{sec:results_app}.


\section{Benchmark Frameworks}
\label{sec:benchmark_frameworks}
We provide a compact overview of benchmarking frameworks in~\cref{tab:benchmarking_frameworks}.

\begin{table}[ht]
\caption{Benchmarking Frameworks}
\label{tab:benchmarking_frameworks}
\begin{tabularx}{\textwidth}{
    >{\raggedright\arraybackslash}p{1.8cm} 
    >{\raggedright\arraybackslash}p{2.3cm} 
    >{\raggedright\arraybackslash}p{2.5cm} 
    >{\raggedright\arraybackslash}p{1.5cm} 
    >{\raggedright\arraybackslash}p{1.2cm} 
    >{\raggedright\arraybackslash}X
}
\toprule
\textbf{Framework} & \textbf{Use Case} & \textbf{Tasks} & \textbf{Optimizers} & \textbf{Task Types} & \textbf{Extensibility} \\ \midrule
Bayesmark \citep{bayesmark} & Bayesian optimization on real ML tasks & Cross-product of 9 ML algorithms on 6 Sklearn toy datasets & 10 optimizers & BB & Add new optimizers via wrappers, datasets via CSV files, no new ML algorithms \\ \hline
HPO-B \citep{pineda-neuripsdbt21a} & Benchmarking black-box HPO algorithms with tabular and surrogate benchmarks & 16 configuration spaces on 101 OpenML-based datasets (HPO-B-v2) & 4 optimizers & BB & Add optimizers via wrappers \\ \hline
HPOBench \citep{eggensperger-neuripsdbt21a} & Collection of multi-fidelity benchmarking tasks for HPO & 12 benchmark families with 110 benchmarking tasks & No included optimizers & BB, MO, MF & Add new benchmarks via adding a new benchmark class, tutorial on this \\ \hline
Kurobako\footnote{Kurobako, \url{https://github.com/optuna/kurobako}, 2019.} & Benchmarking black-box optimization & NAS-Bench-101, HPOBench, Sigopt Evalset, Two-objective ZDT functions & 4 optimizers & BB, MO, MF & Add new benchmarks and optimizers via implementation as a command line program \\ \hline
Synetune \citep{salinas-automl2022} & SOTA HPO with tabulated and surrogate benchmarks & NASBench201, FCNet, LCBench, HPO-B, TabRepo & 10 optimizers & BB, MF & Add new benchmarks via code contribution and optimizers via wrappers, tutorials for both  \\ \hline
YAHPO \citep{pfisterer-automl22a} & Collection of benchmarking tasks for HPO and black-box optimization methods & 14 ML algorithms with 852 benchmarking tasks & No included optimizers & BB, MO, MF, MOMF & Add new benchmarks via adding benchmark configuration and meta-data, tutorial on this  \\ \hline
\textbf{\carps (ours)} & HPO for BB, MO, MF, and MOMF with representative benchmarking tasks & HPOBench, YAHPO, MFPBench, BBOB, Pymoo-MO. Subselections based on performance data & \numberofoptimizers optimizers from different frameworks & BB, MO, MF, MOMF & Add new benchmarks and optimizers via wrappers and tutorials for both \\ \bottomrule
\end{tabularx}
\end{table}

\section{Optimizer Overview}
\label{sec:optimizer_overview_app}
See~\cref{tab:optimizers} for an overview of optimizers.
\begin{table}[htb]
\caption{Optimizers included in \carps}
\label{tab:optimizers}
\centering
\begin{tabularx}{\textwidth}{
    >{\raggedright\arraybackslash}p{2.2cm} 
    >{\raggedright\arraybackslash}X
    >{\centering\arraybackslash}p{0.5cm} 
    >{\centering\arraybackslash}p{0.5cm} 
    >{\centering\arraybackslash}p{0.5cm} 
    >{\centering\arraybackslash}p{1.5cm}
}
\toprule
\textbf{Optimizer} & \textbf{Variant} & \textbf{BB} & \textbf{MO} & \textbf{MF} & \textbf{MO-MF} \\ \midrule
RandomSearch & & \checkmark & &  &  \\
HEBO         & & \checkmark & &  &  \\
Skopt        & & \checkmark & &  &  \\
Ax        & & \checkmark & &  &  \\
DEHB         & &  & & \checkmark &  \\ \hline
Optuna       & Optuna-TPE & \checkmark & &  &  \\
             & Optuna-MO-TPE &  & \checkmark &  &  \\
             & Optuna-MO-NSGAII &  & \checkmark &  &  \\ \hline
SMAC3-2.0 & BlackBoxFacade       & \checkmark & &  &  \\ 
         & MO (ParEGO)~\citep{knowls-evoco06a} & & \checkmark &  &  \\
         & MultiFidelityFacade  & & & \checkmark &  \\
         & Hyperband~\citep{li-jmlr18a}  & & & \checkmark &  \\
         & MultiFidelityFacade (GP or RF) with MO (ParEGO)~\citep{knowls-evoco06a}  & & & & \checkmark  \\\hline
Nevergrad & NGOpt                               & \checkmark & &  &  \\
          & NoisyBandit                         & \checkmark & &  &  \\
          & BayesOpt~\citep{nogueira-2014a}      & \checkmark & &  &  \\ 
          & Hyperopt~\citep{bergstra-icml13a}   & \checkmark & &  &  \\ 
          & CMA-ES~\citep{hansen-2019a}         & \checkmark & &  &  \\ 
          & EvolutionStrategy                   & & \checkmark &  &  \\ 
          & DifferentialEvolution              & & \checkmark &  &  \\ \hline
SyneTune & Conformal Quantile Regression (CQR)~\citep{salinas-icml23a} &  \checkmark & & & \\
         & BORE~\citep{tiao-icml21a}                   & \checkmark &  & &  \\
         & BOTorch~\citep{balandat-neurips20a} & \checkmark & & \\
         & TPE~\citep{bergstra-nips11a} & \checkmark & & \\
         & ASHA~\citep{li-mlsys20a}                   & &  & \checkmark &  \\
         & BOHB~\citep{falkner-icml18a}                   & &  & \checkmark &  \\
         \bottomrule
\end{tabularx}
\end{table}

\section{Maintenance Plan}
\label{sec:maintenance_plan}
Following~\citep{eggensperger-neuripsdbt21a} and~\citep{pfisterer-automl22a} we describe our maintenance plan of \carps.

\textbf{Who Maintains}
\carps is developed and maintained by \repomaintainers.

\textbf{Contact}
Questions and issues regarding the repository and code can be posted in the GitHub repository (\carpsurl).
Other questions can be asked via the provided e-mail.

\textbf{Erratum}
There is no erratum.

\textbf{Library Updates}
We plan on updating the library with new features for experiment running, analysis improvement and more optimizers and benchmarks, potentially via external pull requests.
Changes will be communicated via Github releases as well as a CHANGELOG.

\textbf{Support for Older Versions}
Older versions of \carps will be available on GitHub but with limited support.

\textbf{Contributions}
Contributions to our benchmarking framework \carps are very welcome.
These can either be general features or more optimizers and benchmarks.
For the latter, we provide a tutorial for contributing a benchmark (\urlcontributebenchmark) with an example repository (\urlexamplerepo), and one for contributing an optimizer (\urlcontributeoptimizer) with a template repository (\urltemplateoptimizer).
See \urlcontribute for a guide how to contribute.
Contributions are managed via pull requests.

\textbf{Dependencies}
Python package dependencies are accessible in \code{pyproject.toml} in the repository. 
The \carps version used for the experiments is 1.0.0.

\section{Computational Resources}
\label{sec:computational_resources}
All experiments were run on single CPUs of type AMD Milan 7763, 2.45 GHz, up to 3.5 GHz, each 2x 64 cores, 128GB main memory.
The estimated runtime for the experiments for on optimizer on the subselection is \cpuhours CPU hours, see~\cref{tab:runtimes}.

\begin{table}[h]
\centering
\caption{Runtimes in CPU Hours per Task Type}
\label{tab:runtimes}
\begin{tabular}{lr}
\toprule
 & time \\
\midrule
blackbox & 21 \\
multi-fidelity & 2589 \\
multi-fidelity-objective & 95 \\
multi-objective & 330 \\
total & 3035 \\
\bottomrule
\end{tabular}
\end{table}

\section{Interface: Technical Details}
\label{sec:interface_technical_details}
In order to communicate between optimizer and objective functions we propose a standardized interface.
This interface relies on \code{ConfigSpace.ConfigurationSpace} as the representation of the configuration space and \code{TrialInfo} and \code{TrialValue} to hold information about a trial.
A trial info is associated with the \ac{HP} configuration, the budget for multi-fidelity, the instance in the case of algorithm configuration and an optional seed.
In addition, a name can be associated with a trial info as well as a checkpoint path.
The benchmark task requires a definition of the configuration space and must provide the function \code{evaluate}.
The optimizer class holds one objective function and requires ask-and-tell to be implemented.
In the case the underlying optimizer does not support ask-and-tell its normal run method can be used.

\section{Optimization of the Star Discrepancy: Technical Details}
\label{sec:star_disc_opt_desc}
\begin{figure}[ht]
   \centering
   \includegraphics[width=0.4\textwidth]{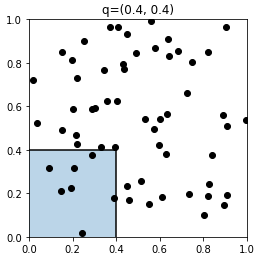}
   \caption{The discrepancy for the box defined by $q$ is given by the volume, 0.16, minus the proportion of points inside $[0,q)$, 7/60. It gives a value of 0.0433. Shifting $q$ over the entire space and keeping the worst discrepancy value gives the discrepancy value of the point set $P$.}
   \label{fig:exdiscre}
\end{figure}

Subset Selection was formally introduced in~\citep{CDP22} with exact methods in dimension 2, and extended by heuristics in~\citep{CDP23} for higher dimensions and any number of points.
Since our test cases are in dimension 3 (because we describe every task with the performance of \revt{three diverse optimization methods -- for the black box task types it would be }random search, Bayesian optimization and CMA-ES) with thousands of data points, we use the heuristics from~\citep{CDP23}.
While multiple versions of the heuristic are introduced, the general principle behind each of them is the same.
For the general principle see~\cref{fig:exdiscre}.
It is an iterative process where, at each step, one point in the current best subset is replaced by a point which is not yet selected.
If the discrepancy of the new set is lower than that of the previous, the new set is kept for the next step.
Since there are a very large number of possible point exchanges at each step, a careful selection of both the outgoing and the incoming points has to be made to avoid numerous expensive discrepancy calculations.
The outgoing point is chosen as one of the points defining the box with worst local discrepancy (see~\citet{NieBox} for a description of the discrete structure of the $L_{\infty}$ star discrepancy calculation).
This point is associated to the discrepancy value for a specific dimension, the incoming point is then selected from neighbouring points in this dimension.
In the ``nobrute'' implementations, the heuristic stops once, for a given set, all such swaps have been performed and no improvement on the discrepancy could be made.
For the other implementations, a further brute force verification of all possible remaining swaps is made to guarantee that the final set is a local optimum for our heuristic.
The heuristic then returns the subset and its associated discrepancy value.
In all cases, the result of a single run of the heuristic is highly dependent on the initial, random, set used to start the heuristic.
A large number of runs are therefore performed with the best (discrepancy value, associated set) combination kept.
This is particularly important when $n$ is much larger than the number of selected points $k$, as is the case in our setting.

\section{Subselection}
\label{sec:subselection_app}
We cover additional information about the subselections per task type, i.e. show the different subset sizes, validate the ranking across subsets, visualize the subsets and show statistics as well as listing the tasks for the dev and test subset.

\subsection{Different Subset Sizes}
\label{sec:subset_diff_k}
\begin{figure}[h]
    \centering
    \begin{subfigure}[t]{0.45\textwidth}    \includegraphics[width=\linewidth]{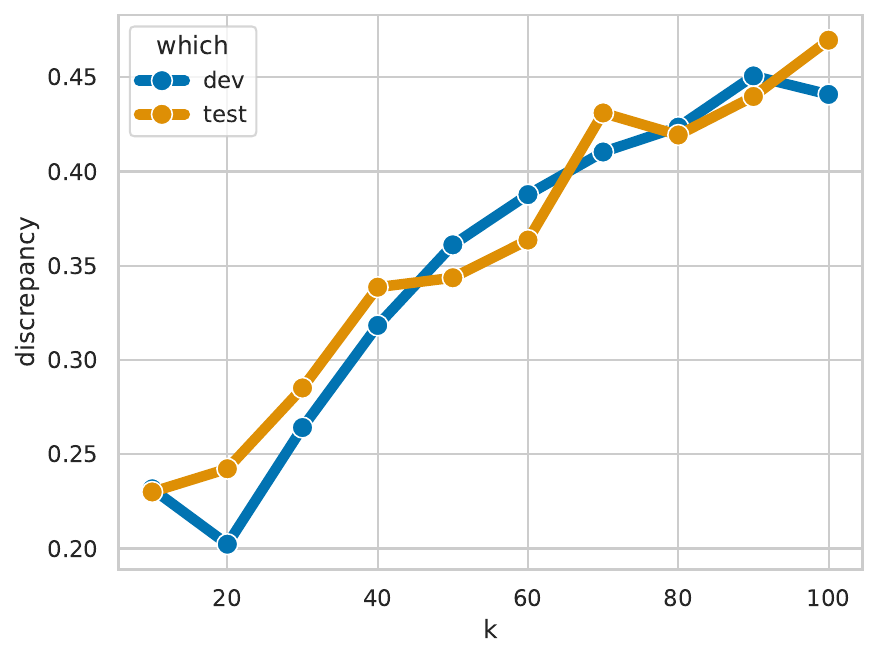}
    \caption{Blackbox task type}
    \label{fig:subset_bb_diff_k}
    \end{subfigure}\hfill    
    \begin{subfigure}[t]{0.45\textwidth}    \includegraphics[width=\linewidth]{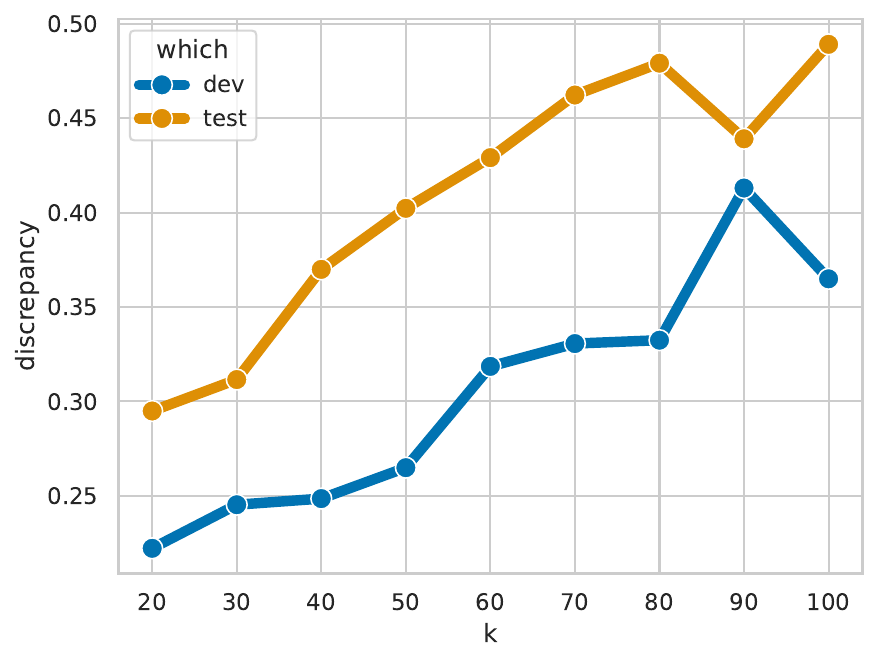}
    \caption{Multi-fidelity task type}
    \label{fig:subset_mf_diff_k}
    \end{subfigure}\\
    \begin{subfigure}[t]{0.45\textwidth}    \includegraphics[width=\linewidth]{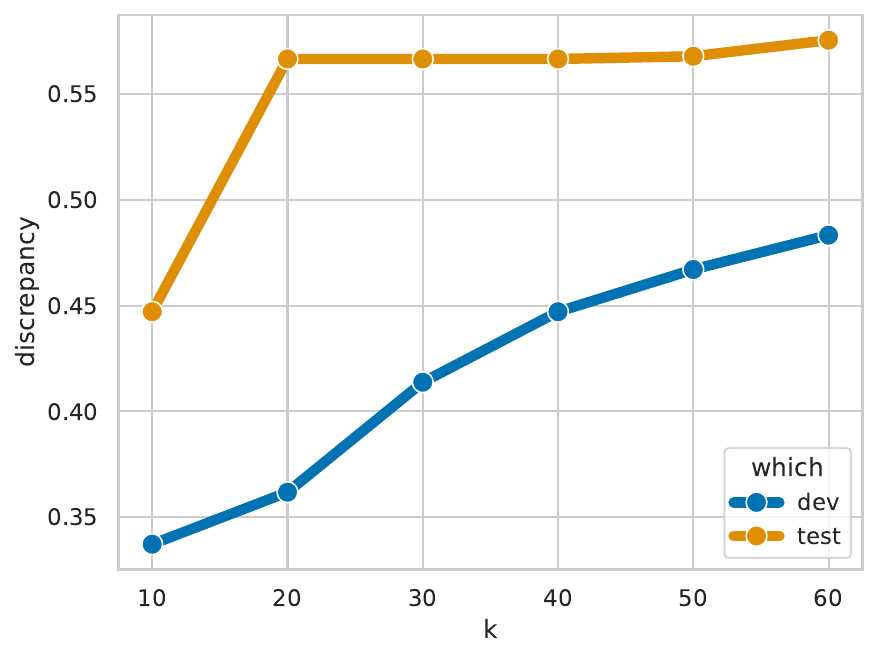}
    \caption{Multi-objective task type}
    \label{fig:subset_mo_diff_k}
    \end{subfigure}\hfill    
    \begin{subfigure}[t]{0.45\textwidth}    \includegraphics[width=\linewidth]{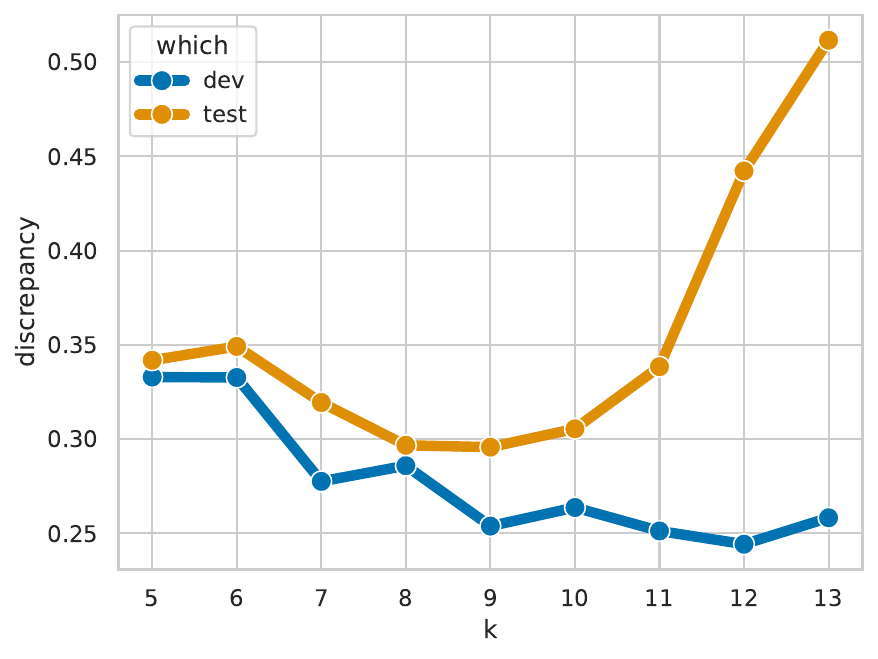}
    \caption{Multi-objective-multi-fidelity task type (Here the fulls set size is \fullsizemomf so we select among $k \in \{5,6,...,13\}$.)}
    \label{fig:subset_momf_diff_k}
    \end{subfigure}
\caption{Different subset sizes $k$ with the star discrepancy}

\end{figure}

For the blackbox task type we optimize the star discrepancy for different $k \in \{20, 30, \dots, 100\}$.
We choose $k=30$ because the discrepancy sum of both sets, dev and test, is the lowest, see~\cref{fig:subset_bb_diff_k}.
For the multi-fidelity task type we choose $k = 20$, see~\cref{fig:subset_mf_diff_k}, and for multi-objective $k=10$ (\cref{fig:subset_mo_diff_k}).

\subsection{Ranking Validation}
\label{sec:ranking_validation}
In order to validate that the ranking remains consistent across the subsets, we calculate the rank the same way as described in \cref{sec:autorank} for each task type and then determine the order.
The following tables (\cref{tab:ranking_validation_blackbox}, \cref{tab:ranking_validation_multi-fidelity}, \cref{tab:ranking_validation_multi-objective}, \cref{tab:ranking_validation_multi-fidelity-objective}) show that the ranking is consistent across the subsets for each task type. 

\begin{table}[h]
    \caption{Mean Ranking for Scenario blackbox}
    \label{tab:ranking_validation_blackbox}
    \centering
    \begin{tabular}{lllll}
\toprule
optimizer\_id & Nevergrad-CMA-ES & SMAC3-BlackBoxFacade & RandomSearch & significant \\
set\_id &  &  &  &  \\
\midrule
dev & 1.65 (2) & 1.65 (2) & 2.70 (3) & yes \\
test & 1.82 (2) & 1.50 (1) & 2.67 (3) & yes \\
\bottomrule
\end{tabular}
\end{table}

\begin{table}[h]
    \caption{Mean Ranking for Scenario multi-fidelity}
    \label{tab:ranking_validation_multi-fidelity}
    \centering
    \begin{tabular}{lllll}
\toprule
optimizer\_id & SMAC3-MultiFidelityFacade & DEHB & SMAC3-Hyperband & significant \\
set\_id &  &  &  &  \\
\midrule
dev & 1.35 (1) & 2.23 (2) & 2.42 (3) & yes \\
test & 1.57 (1) & 2.15 (2) & 2.27 (3) & yes \\
\bottomrule
\end{tabular}
\end{table}

\begin{table}[h]
    \caption{Mean Ranking for Scenario multi-fidelity-objective}
    \label{tab:ranking_validation_multi-fidelity-objective}
    \centering
    \begin{tabular}{lllll}
\toprule
optimizer\_id & SMAC3-MOMF-GP & RandomSearch & Nevergrad-DE & significant \\
set\_id &  &  &  &  \\
\midrule
dev & 1.56 (1) & 1.78 (2) & 2.67 (3) & yes \\
test & 1.44 (1) & 2.00 (2) & 2.56 (3) & no \\
\bottomrule
\end{tabular}
\end{table}

\begin{table}[h]
    \caption{Mean Ranking for Scenario multi-objective}
    \label{tab:ranking_validation_multi-objective}
    \centering
    \begin{tabular}{lllll}
\toprule
optimizer\_id & Optuna-MO-TPE & Nevergrad-DE & RandomSearch & significant \\
set\_id &  &  &  &  \\
\midrule
dev & 1.30 (1) & 2.20 (2) & 2.50 (3) & yes \\
test & 1.80 (2) & 1.70 (1) & 2.50 (3) & no \\
\bottomrule
\end{tabular}
\end{table}

\FloatBarrier

\subsection{Selected Sets}
\label{sec:selected_sets}
\begin{table}[h]
    \centering
    \caption{Details of Subselection}
    \label{tab:selected_sets_details}
    \begin{tabular}{lrrrr}
    \toprule
       Task Type  & Statistics & Task List (dev) & Task List (test) \\
    \midrule
       Blackbox  & \cref{fig:subset_stats} & \cref{tab:selectedtasks-blackbox-dev} & \cref{tab:selectedtasks-blackbox-test} \\
       Multi-Fidelity & \cref{fig:subset_stats} & \cref{tab:selectedtasks-multi-fidelity-dev} & \cref{tab:selectedtasks-multi-fidelity-test} \\
       Multi-Objective & \cref{fig:subset_stats} & \cref{tab:selectedtasks-multi-objective-dev} & \cref{tab:selectedtasks-multi-fidelity-objective-test} \\
       Multi-Objective-Multi-Fidelity & \cref{fig:subset_stats} & \cref{tab:selectedtasks-multi-fidelity-objective-dev} & \cref{tab:selectedtasks-multi-fidelity-objective-test} \\
    \bottomrule
    \end{tabular}
\end{table}

\begin{table}
\caption{Selected tasks ('blackbox', 'dev')}
\label{tab:selectedtasks-blackbox-dev}
\centering
\resizebox{\textwidth}{!}{\begin{tabular}{llrrrrrr}
\toprule
benchmark\_id & task & dimensions & n\_trials & n\_floats & n\_integers & n\_categoricals & n\_ordinals \\
\midrule
YAHPO & blackbox/20/dev/yahpo/rbv2\_xgboost/23512/None & 14 & 170 & 10 & 2 & 2 & 0 \\
HPOBench & blackbox/20/dev/hpobench/bb/tab/ml/lr/146818 & 2 & 77 & 0 & 0 & 0 & 2 \\
YAHPO & blackbox/20/dev/yahpo/rbv2\_aknn/458/None & 6 & 118 & 0 & 4 & 2 & 0 \\
HPOBench & blackbox/20/dev/hpobench/bb/tab/nas/SliceLocalizationBenchmark & 9 & 140 & 0 & 0 & 3 & 6 \\
BBOB & blackbox/20/dev/bbob/noiseless/2/12/0 & 2 & 77 & 2 & 0 & 0 & 0 \\
BBOB & blackbox/20/dev/bbob/noiseless/2/20/0 & 2 & 77 & 2 & 0 & 0 & 0 \\
YAHPO & blackbox/20/dev/yahpo/rbv2\_ranger/40927/None & 8 & 134 & 2 & 3 & 3 & 0 \\
YAHPO & blackbox/20/dev/yahpo/rbv2\_svm/24/None & 6 & 118 & 3 & 1 & 2 & 0 \\
HPOBench & blackbox/20/dev/hpobench/bb/tab/nas/NavalPropulsionBenchmark & 9 & 140 & 0 & 0 & 3 & 6 \\
HPOBench & blackbox/20/dev/hpobench/bb/tab/ml/xgboost/146212 & 4 & 100 & 0 & 0 & 0 & 4 \\
YAHPO & blackbox/20/dev/yahpo/rbv2\_svm/182/None & 6 & 118 & 3 & 1 & 2 & 0 \\
YAHPO & blackbox/20/dev/yahpo/rbv2\_xgboost/42/None & 14 & 170 & 10 & 2 & 2 & 0 \\
YAHPO & blackbox/20/dev/yahpo/rbv2\_aknn/312/None & 6 & 118 & 0 & 4 & 2 & 0 \\
YAHPO & blackbox/20/dev/yahpo/rbv2\_aknn/40498/None & 6 & 118 & 0 & 4 & 2 & 0 \\
YAHPO & blackbox/20/dev/yahpo/rbv2\_glmnet/41157/None & 3 & 90 & 2 & 0 & 1 & 0 \\
HPOBench & blackbox/20/dev/hpobench/bb/tab/ml/rf/146212 & 4 & 100 & 0 & 0 & 0 & 4 \\
BBOB & blackbox/20/dev/bbob/noiseless/4/6/1 & 4 & 100 & 4 & 0 & 0 & 0 \\
YAHPO & blackbox/20/dev/yahpo/rbv2\_aknn/1462/None & 6 & 118 & 0 & 4 & 2 & 0 \\
YAHPO & blackbox/20/dev/yahpo/lcbench/168335/None & 7 & 126 & 4 & 3 & 0 & 0 \\
BBOB & blackbox/20/dev/bbob/noiseless/2/12/1 & 2 & 77 & 2 & 0 & 0 & 0 \\
\bottomrule
\end{tabular}}
\end{table}

\begin{table}
\caption{Selected tasks ('blackbox', 'test')}
\label{tab:selectedtasks-blackbox-test}
\centering
\resizebox{\textwidth}{!}{\begin{tabular}{llrrrrrr}
\toprule
benchmark\_id & task & dimensions & n\_trials & n\_floats & n\_integers & n\_categoricals & n\_ordinals \\
\midrule
YAHPO & blackbox/20/test/yahpo/lcbench/167184/None & 7 & 126 & 4 & 3 & 0 & 0 \\
BBOB & blackbox/20/test/bbob/noiseless/16/11/0 & 16 & 180 & 16 & 0 & 0 & 0 \\
BBOB & blackbox/20/test/bbob/noiseless/2/6/1 & 2 & 77 & 2 & 0 & 0 & 0 \\
BBOB & blackbox/20/test/bbob/noiseless/16/1/1 & 16 & 180 & 16 & 0 & 0 & 0 \\
YAHPO & blackbox/20/test/yahpo/rbv2\_ranger/29/None & 8 & 134 & 2 & 3 & 3 & 0 \\
BBOB & blackbox/20/test/bbob/noiseless/32/9/0 & 32 & 247 & 32 & 0 & 0 & 0 \\
BBOB & blackbox/20/test/bbob/noiseless/32/11/0 & 32 & 247 & 32 & 0 & 0 & 0 \\
YAHPO & blackbox/20/test/yahpo/rbv2\_svm/1493/None & 6 & 118 & 3 & 1 & 2 & 0 \\
BBOB & blackbox/20/test/bbob/noiseless/8/22/0 & 8 & 134 & 8 & 0 & 0 & 0 \\
HPOBench & blackbox/20/test/hpobench/bb/tab/ml/svm/12 & 2 & 77 & 0 & 0 & 0 & 2 \\
YAHPO & blackbox/20/test/yahpo/rbv2\_xgboost/1457/None & 14 & 170 & 10 & 2 & 2 & 0 \\
YAHPO & blackbox/20/test/yahpo/rbv2\_xgboost/1510/None & 14 & 170 & 10 & 2 & 2 & 0 \\
YAHPO & blackbox/20/test/yahpo/rbv2\_xgboost/41027/None & 14 & 170 & 10 & 2 & 2 & 0 \\
YAHPO & blackbox/20/test/yahpo/rbv2\_glmnet/32/None & 3 & 90 & 2 & 0 & 1 & 0 \\
YAHPO & blackbox/20/test/yahpo/rbv2\_rpart/18/None & 5 & 110 & 1 & 3 & 1 & 0 \\
YAHPO & blackbox/20/test/yahpo/rbv2\_rpart/4534/None & 5 & 110 & 1 & 3 & 1 & 0 \\
BBOB & blackbox/20/test/bbob/noiseless/2/9/0 & 2 & 77 & 2 & 0 & 0 & 0 \\
YAHPO & blackbox/20/test/yahpo/rbv2\_glmnet/375/None & 3 & 90 & 2 & 0 & 1 & 0 \\
YAHPO & blackbox/20/test/yahpo/rbv2\_xgboost/1493/None & 14 & 170 & 10 & 2 & 2 & 0 \\
BBOB & blackbox/20/test/bbob/noiseless/2/12/2 & 2 & 77 & 2 & 0 & 0 & 0 \\
\bottomrule
\end{tabular}}
\end{table}

\begin{table}
\caption{Selected tasks ('multi-fidelity', 'dev')}
\label{tab:selectedtasks-multi-fidelity-dev}
\centering
\resizebox{\textwidth}{!}{\begin{tabular}{llrrrrrrlrr}
\toprule
benchmark\_id & task & dimensions & n\_trials & n\_floats & n\_integers & n\_categoricals & n\_ordinals & fidelity\_type & min\_budget & max\_budget \\
\midrule
YAHPO & multifidelity/20/dev/yahpo/rbv2\_ranger/40983/trainsize & 8 & 134 & 2 & 3 & 3 & 0 & trainsize & 0.03 & 1.00 \\
YAHPO & multifidelity/20/dev/yahpo/rbv2\_ranger/41161/trainsize & 8 & 134 & 2 & 3 & 3 & 0 & trainsize & 0.03 & 1.00 \\
YAHPO & multifidelity/20/dev/yahpo/rbv2\_xgboost/40499/trainsize & 14 & 170 & 10 & 2 & 2 & 0 & trainsize & 0.03 & 1.00 \\
YAHPO & multifidelity/20/dev/yahpo/rbv2\_svm/24/trainsize & 6 & 118 & 3 & 1 & 2 & 0 & trainsize & 0.03 & 1.00 \\
YAHPO & multifidelity/20/dev/yahpo/rbv2\_rpart/1220/repl & 5 & 110 & 1 & 3 & 1 & 0 & repl & 1.00 & 10.00 \\
YAHPO & multifidelity/20/dev/yahpo/rbv2\_xgboost/375/trainsize & 14 & 170 & 10 & 2 & 2 & 0 & trainsize & 0.03 & 1.00 \\
YAHPO & multifidelity/20/dev/yahpo/rbv2\_ranger/41161/repl & 8 & 134 & 2 & 3 & 3 & 0 & repl & 1.00 & 10.00 \\
HPOBench & multifidelity/20/dev/hpobench/mf/real/ml/rf/31/subsample & 4 & 100 & 1 & 3 & 0 & 0 & subsample & 0.10 & 1.00 \\
YAHPO & multifidelity/20/dev/yahpo/rbv2\_svm/24/repl & 6 & 118 & 3 & 1 & 2 & 0 & repl & 1.00 & 10.00 \\
HPOBench & multifidelity/20/dev/hpobench/mf/real/ml/nn/146821/iter & 5 & 110 & 2 & 3 & 0 & 0 & iter & 3.00 & 243.00 \\
YAHPO & multifidelity/20/dev/yahpo/rbv2\_rpart/41165/repl & 5 & 110 & 1 & 3 & 1 & 0 & repl & 1.00 & 10.00 \\
YAHPO & multifidelity/20/dev/yahpo/rbv2\_aknn/1497/trainsize & 6 & 118 & 0 & 4 & 2 & 0 & trainsize & 0.03 & 1.00 \\
YAHPO & multifidelity/20/dev/yahpo/rbv2\_svm/40975/repl & 6 & 118 & 3 & 1 & 2 & 0 & repl & 1.00 & 10.00 \\
YAHPO & multifidelity/20/dev/yahpo/rbv2\_super/40984/repl & 38 & 267 & 18 & 13 & 7 & 0 & repl & 1.00 & 10.00 \\
MFPBench & multifidelity/20/dev/mfpbench/SO/mfh/mfh6\_moderate & 6 & 118 & 6 & 0 & 0 & 0 & z & 1.00 & 100.00 \\
YAHPO & multifidelity/20/dev/yahpo/rbv2\_glmnet/24/repl & 3 & 90 & 2 & 0 & 1 & 0 & repl & 1.00 & 10.00 \\
YAHPO & multifidelity/20/dev/yahpo/rbv2\_ranger/41159/repl & 8 & 134 & 2 & 3 & 3 & 0 & repl & 1.00 & 10.00 \\
YAHPO & multifidelity/20/dev/yahpo/rbv2\_xgboost/1476/repl & 14 & 170 & 10 & 2 & 2 & 0 & repl & 1.00 & 10.00 \\
YAHPO & multifidelity/20/dev/yahpo/iaml\_glmnet/41146/trainsize & 2 & 77 & 2 & 0 & 0 & 0 & trainsize & 0.03 & 1.00 \\
YAHPO & multifidelity/20/dev/yahpo/rbv2\_glmnet/334/repl & 3 & 90 & 2 & 0 & 1 & 0 & repl & 1.00 & 10.00 \\
\bottomrule
\end{tabular}}
\end{table}

\begin{table}
\caption{Selected tasks ('multi-fidelity', 'test')}
\label{tab:selectedtasks-multi-fidelity-test}
\centering
\resizebox{\textwidth}{!}{\begin{tabular}{llrrrrrrlrr}
\toprule
benchmark\_id & task & dimensions & n\_trials & n\_floats & n\_integers & n\_categoricals & n\_ordinals & fidelity\_type & min\_budget & max\_budget \\
\midrule
YAHPO & multifidelity/20/test/yahpo/rbv2\_rpart/38/repl & 5 & 110 & 1 & 3 & 1 & 0 & repl & 1.00 & 10.00 \\
YAHPO & multifidelity/20/test/yahpo/rbv2\_xgboost/1480/repl & 14 & 170 & 10 & 2 & 2 & 0 & repl & 1.00 & 10.00 \\
YAHPO & multifidelity/20/test/yahpo/rbv2\_xgboost/1476/trainsize & 14 & 170 & 10 & 2 & 2 & 0 & trainsize & 0.03 & 1.00 \\
YAHPO & multifidelity/20/test/yahpo/rbv2\_super/40900/repl & 38 & 267 & 18 & 13 & 7 & 0 & repl & 1.00 & 10.00 \\
YAHPO & multifidelity/20/test/yahpo/rbv2\_aknn/1476/trainsize & 6 & 118 & 0 & 4 & 2 & 0 & trainsize & 0.03 & 1.00 \\
YAHPO & multifidelity/20/test/yahpo/rbv2\_super/458/trainsize & 38 & 267 & 18 & 13 & 7 & 0 & trainsize & 0.03 & 1.00 \\
YAHPO & multifidelity/20/test/yahpo/rbv2\_aknn/50/trainsize & 6 & 118 & 0 & 4 & 2 & 0 & trainsize & 0.03 & 1.00 \\
YAHPO & multifidelity/20/test/yahpo/rbv2\_super/41156/trainsize & 38 & 267 & 18 & 13 & 7 & 0 & trainsize & 0.03 & 1.00 \\
HPOBench & multifidelity/20/test/hpobench/mf/real/ml/xgboost/3/subsample & 4 & 100 & 3 & 1 & 0 & 0 & subsample & 0.10 & 1.00 \\
YAHPO & multifidelity/20/test/yahpo/rbv2\_ranger/40923/trainsize & 8 & 134 & 2 & 3 & 3 & 0 & trainsize & 0.03 & 1.00 \\
HPOBench & multifidelity/20/test/hpobench/mf/real/ml/nn/146821/iter & 5 & 110 & 2 & 3 & 0 & 0 & iter & 3.00 & 243.00 \\
YAHPO & multifidelity/20/test/yahpo/rbv2\_xgboost/41163/repl & 14 & 170 & 10 & 2 & 2 & 0 & repl & 1.00 & 10.00 \\
YAHPO & multifidelity/20/test/yahpo/rbv2\_super/4154/repl & 38 & 267 & 18 & 13 & 7 & 0 & repl & 1.00 & 10.00 \\
YAHPO & multifidelity/20/test/yahpo/rbv2\_super/458/repl & 38 & 267 & 18 & 13 & 7 & 0 & repl & 1.00 & 10.00 \\
YAHPO & multifidelity/20/test/yahpo/rbv2\_svm/40981/trainsize & 6 & 118 & 3 & 1 & 2 & 0 & trainsize & 0.03 & 1.00 \\
YAHPO & multifidelity/20/test/yahpo/rbv2\_aknn/40670/trainsize & 6 & 118 & 0 & 4 & 2 & 0 & trainsize & 0.03 & 1.00 \\
YAHPO & multifidelity/20/test/yahpo/rbv2\_glmnet/41162/trainsize & 3 & 90 & 2 & 0 & 1 & 0 & trainsize & 0.03 & 1.00 \\
YAHPO & multifidelity/20/test/yahpo/rbv2\_xgboost/40685/trainsize & 14 & 170 & 10 & 2 & 2 & 0 & trainsize & 0.03 & 1.00 \\
YAHPO & multifidelity/20/test/yahpo/rbv2\_ranger/40923/repl & 8 & 134 & 2 & 3 & 3 & 0 & repl & 1.00 & 10.00 \\
YAHPO & multifidelity/20/test/yahpo/rbv2\_aknn/1461/repl & 6 & 118 & 0 & 4 & 2 & 0 & repl & 1.00 & 10.00 \\
\bottomrule
\end{tabular}}
\end{table}

\begin{table}
\caption{Selected tasks ('multi-objective', 'dev')}
\label{tab:selectedtasks-multi-objective-dev}
\centering
\resizebox{\textwidth}{!}{\begin{tabular}{llrrrrrrr}
\toprule
benchmark\_id & task & dimensions & n\_trials & n\_floats & n\_integers & n\_categoricals & n\_ordinals & n\_objectives \\
\midrule
HPOBench & multiobjective/10/dev/hpobench/MO/tab/ml/nn/146821 & 5 & 110 & 0 & 0 & 0 & 5 & 2 \\
HPOBench & multiobjective/10/dev/hpobench/MO/tab/ml/xgboost/31 & 4 & 100 & 0 & 0 & 0 & 4 & 2 \\
Pymoo & multiobjective/10/dev/Pymoo/ManyO/unconstraint/wfg7\_10\_5 & 10 & 147 & 10 & 0 & 0 & 0 & 5 \\
YAHPO & multiobjective/10/dev/yahpo/mo/rbv2\_xgboost/12/None & 14 & 170 & 10 & 2 & 2 & 0 & 2 \\
Pymoo & multiobjective/10/dev/Pymoo/ManyO/unconstraint/dtlz5 & 10 & 147 & 10 & 0 & 0 & 0 & 3 \\
HPOBench & multiobjective/10/dev/hpobench/MO/tab/ml/lr/14965 & 2 & 77 & 0 & 0 & 0 & 2 & 2 \\
HPOBench & multiobjective/10/dev/hpobench/MO/tab/ml/nn/9952 & 5 & 110 & 0 & 0 & 0 & 5 & 2 \\
YAHPO & multiobjective/10/dev/yahpo/mo/rbv2\_xgboost/28/None & 14 & 170 & 10 & 2 & 2 & 0 & 2 \\
YAHPO & multiobjective/10/dev/yahpo/mo/rbv2\_rpart/1476/None & 5 & 110 & 1 & 3 & 1 & 0 & 2 \\
HPOBench & multiobjective/10/dev/hpobench/MO/tab/ml/rf/12 & 4 & 100 & 0 & 0 & 0 & 4 & 2 \\
\bottomrule
\end{tabular}}
\end{table}

\begin{table}
\caption{Selected tasks ('multi-objective', 'test')}
\label{tab:selectedtasks-multi-objective-test}
\centering
\resizebox{\textwidth}{!}{\begin{tabular}{llrrrrrrr}
\toprule
benchmark\_id & task & dimensions & n\_trials & n\_floats & n\_integers & n\_categoricals & n\_ordinals & n\_objectives \\
\midrule
Pymoo & multiobjective/10/test/Pymoo/MO/unconstraint/zdt1 & 30 & 240 & 30 & 0 & 0 & 0 & 2 \\
YAHPO & multiobjective/10/test/yahpo/mo/rbv2\_xgboost/182/None & 14 & 170 & 10 & 2 & 2 & 0 & 2 \\
HPOBench & multiobjective/10/test/hpobench/MO/tab/ml/rf/168911 & 4 & 100 & 0 & 0 & 0 & 4 & 2 \\
HPOBench & multiobjective/10/test/hpobench/MO/tab/ml/xgboost/146212 & 4 & 100 & 0 & 0 & 0 & 4 & 2 \\
HPOBench & multiobjective/10/test/hpobench/MO/tab/ml/nn/3917 & 5 & 110 & 0 & 0 & 0 & 5 & 2 \\
YAHPO & multiobjective/10/test/yahpo/mo/lcbench/189873/None & 7 & 126 & 4 & 3 & 0 & 0 & 2 \\
HPOBench & multiobjective/10/test/hpobench/MO/tab/ml/lr/12 & 2 & 77 & 0 & 0 & 0 & 2 & 2 \\
HPOBench & multiobjective/10/test/hpobench/MO/tab/ml/lr/3 & 2 & 77 & 0 & 0 & 0 & 2 & 2 \\
HPOBench & multiobjective/10/test/hpobench/MO/tab/ml/rf/167119 & 4 & 100 & 0 & 0 & 0 & 4 & 2 \\
HPOBench & multiobjective/10/test/hpobench/MO/tab/ml/rf/167120 & 4 & 100 & 0 & 0 & 0 & 4 & 2 \\
\bottomrule
\end{tabular}}
\end{table}

\begin{table}
\caption{Selected tasks ('multi-fidelity-objective', 'dev')}
\label{tab:selectedtasks-multi-fidelity-objective-dev}
\centering
\resizebox{\textwidth}{!}{\begin{tabular}{llrrrrrrlrrr}
\toprule
benchmark\_id & task & dimensions & n\_trials & n\_floats & n\_integers & n\_categoricals & n\_ordinals & fidelity\_type & min\_budget & max\_budget & n\_objectives \\
\midrule
YAHPO & momf/9/dev/yahpo/MOMF/repl/rbv2\_xgboost/12/repl & 14 & 170 & 10 & 2 & 2 & 0 & repl & 1.00 & 10.00 & 2 \\
YAHPO & momf/9/dev/yahpo/MOMF/trainsize/rbv2\_ranger/375/trainsize & 8 & 134 & 2 & 3 & 3 & 0 & trainsize & 0.03 & 1.00 & 2 \\
YAHPO & momf/9/dev/yahpo/MOMF/trainsize/iaml\_ranger/1489/trainsize & 8 & 134 & 2 & 3 & 3 & 0 & trainsize & 0.03 & 1.00 & 3 \\
YAHPO & momf/9/dev/yahpo/MOMF/trainsize/rbv2\_rpart/1476/trainsize & 5 & 110 & 1 & 3 & 1 & 0 & trainsize & 0.03 & 1.00 & 2 \\
YAHPO & momf/9/dev/yahpo/MOMF/trainsize/rbv2\_ranger/6/trainsize & 8 & 134 & 2 & 3 & 3 & 0 & trainsize & 0.03 & 1.00 & 2 \\
YAHPO & momf/9/dev/yahpo/MOMF/trainsize/rbv2\_xgboost/12/trainsize & 14 & 170 & 10 & 2 & 2 & 0 & trainsize & 0.03 & 1.00 & 2 \\
YAHPO & momf/9/dev/yahpo/MOMF/trainsize/rbv2\_xgboost/28/trainsize & 14 & 170 & 10 & 2 & 2 & 0 & trainsize & 0.03 & 1.00 & 2 \\
YAHPO & momf/9/dev/yahpo/MOMF/trainsize/iaml\_glmnet/1489/trainsize & 2 & 77 & 2 & 0 & 0 & 0 & trainsize & 0.03 & 1.00 & 2 \\
YAHPO & momf/9/dev/yahpo/MOMF/epoch/lcbench/167185/epoch & 7 & 126 & 4 & 3 & 0 & 0 & epoch & 1.00 & 52.00 & 2 \\
\bottomrule
\end{tabular}}
\end{table}

\begin{table}
\caption{Selected tasks ('multi-fidelity-objective', 'test')}
\label{tab:selectedtasks-multi-fidelity-objective-test}
\centering
\resizebox{\textwidth}{!}{\begin{tabular}{llrrrrrrlrrr}
\toprule
benchmark\_id & task & dimensions & n\_trials & n\_floats & n\_integers & n\_categoricals & n\_ordinals & fidelity\_type & min\_budget & max\_budget & n\_objectives \\
\midrule
YAHPO & momf/9/test/yahpo/MOMF/epoch/lcbench/189873/epoch & 7 & 126 & 4 & 3 & 0 & 0 & epoch & 1.00 & 52.00 & 2 \\
YAHPO & momf/9/test/yahpo/MOMF/epoch/lcbench/167152/epoch & 7 & 126 & 4 & 3 & 0 & 0 & epoch & 1.00 & 52.00 & 2 \\
YAHPO & momf/9/test/yahpo/MOMF/trainsize/iaml\_xgboost/1489/trainsize & 13 & 165 & 10 & 2 & 1 & 0 & trainsize & 0.03 & 1.00 & 4 \\
YAHPO & momf/9/test/yahpo/MOMF/repl/rbv2\_rpart/40499/repl & 5 & 110 & 1 & 3 & 1 & 0 & repl & 1.00 & 10.00 & 2 \\
YAHPO & momf/9/test/yahpo/MOMF/repl/rbv2\_xgboost/182/repl & 14 & 170 & 10 & 2 & 2 & 0 & repl & 1.00 & 10.00 & 2 \\
YAHPO & momf/9/test/yahpo/MOMF/repl/rbv2\_ranger/6/repl & 8 & 134 & 2 & 3 & 3 & 0 & repl & 1.00 & 10.00 & 2 \\
YAHPO & momf/9/test/yahpo/MOMF/trainsize/iaml\_glmnet/1067/trainsize & 2 & 77 & 2 & 0 & 0 & 0 & trainsize & 0.03 & 1.00 & 2 \\
YAHPO & momf/9/test/yahpo/MOMF/trainsize/rbv2\_xgboost/182/trainsize & 14 & 170 & 10 & 2 & 2 & 0 & trainsize & 0.03 & 1.00 & 2 \\
YAHPO & momf/9/test/yahpo/MOMF/trainsize/rbv2\_ranger/40979/trainsize & 8 & 134 & 2 & 3 & 3 & 0 & trainsize & 0.03 & 1.00 & 2 \\
\bottomrule
\end{tabular}}
\end{table}

\FloatBarrier
\section{Experimental Results}
\label{sec:results_app}

\begin{figure}[h]
    \centering
    \includegraphics[width=0.49\linewidth]{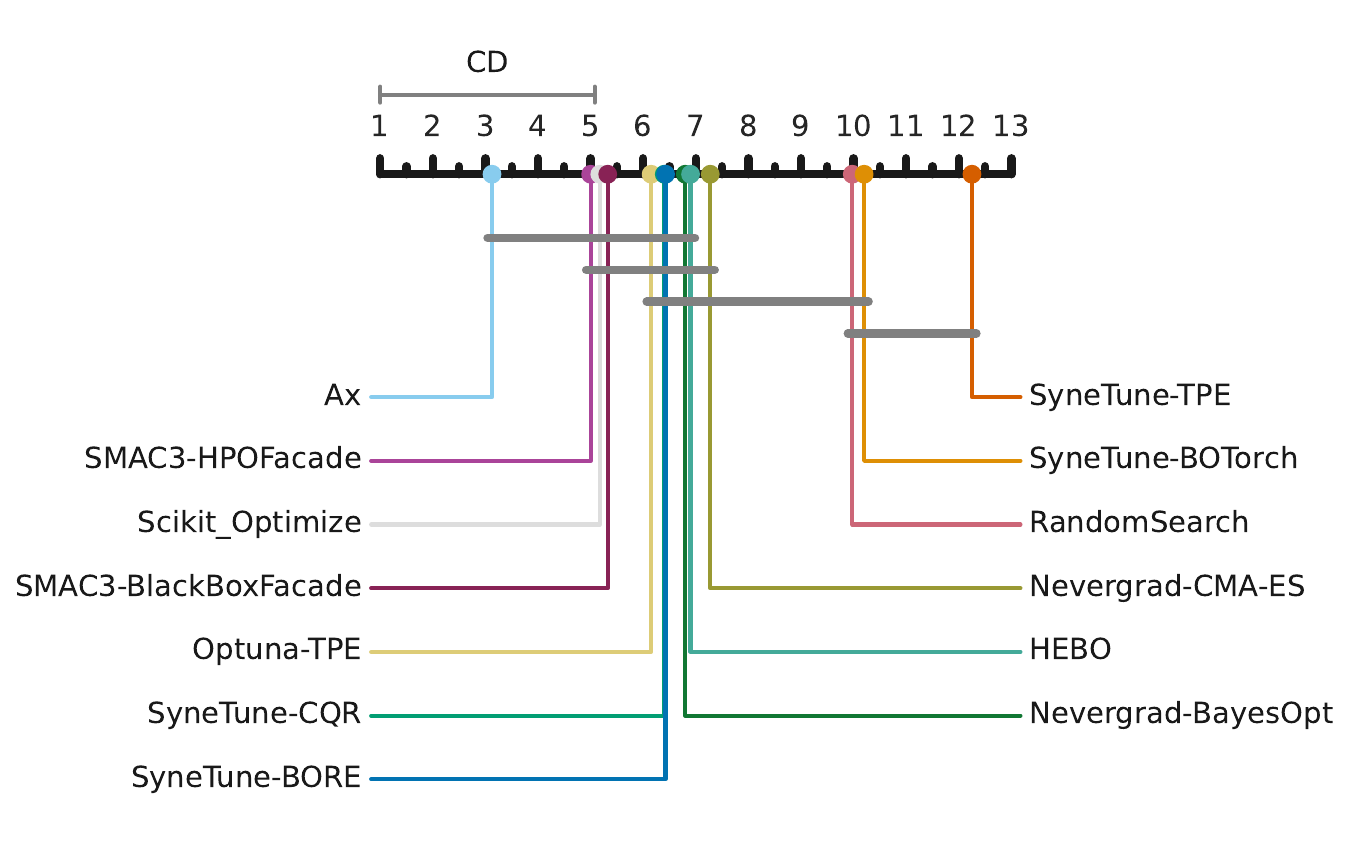}
    \hfill
    \includegraphics[width=0.49\linewidth]{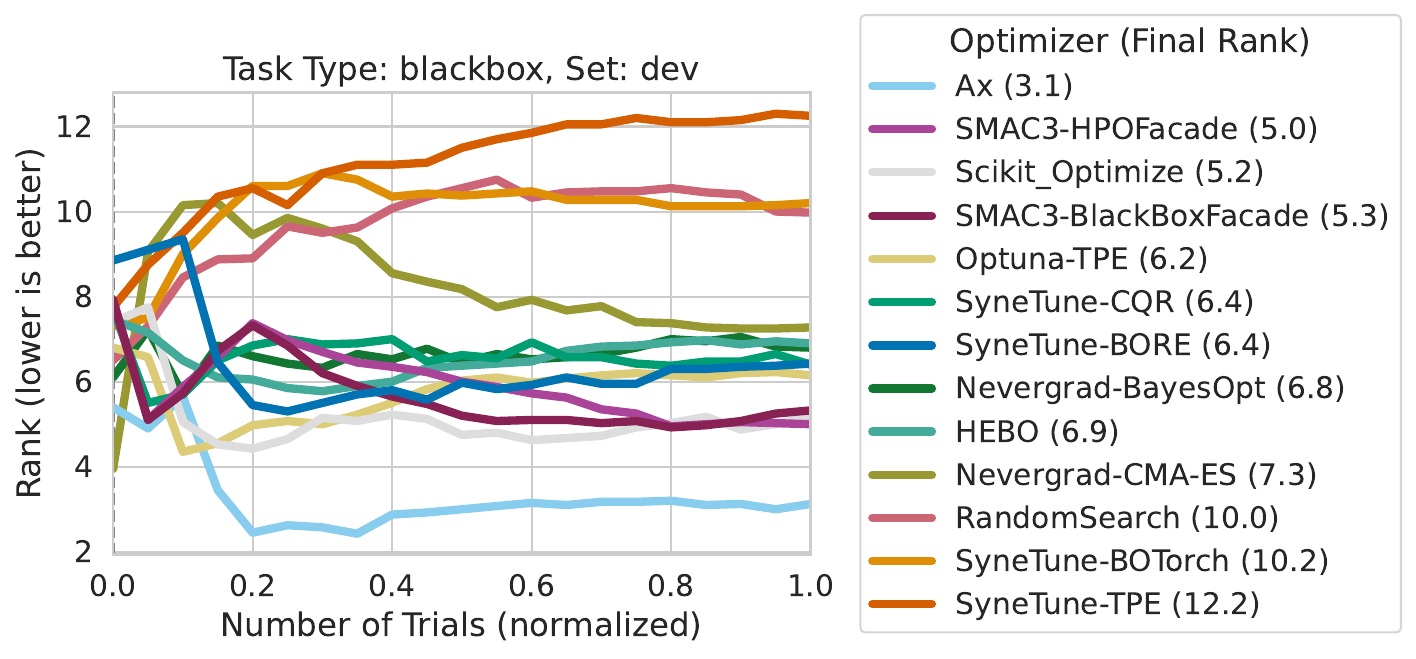}\\
    
    \includegraphics[width=0.8\linewidth]{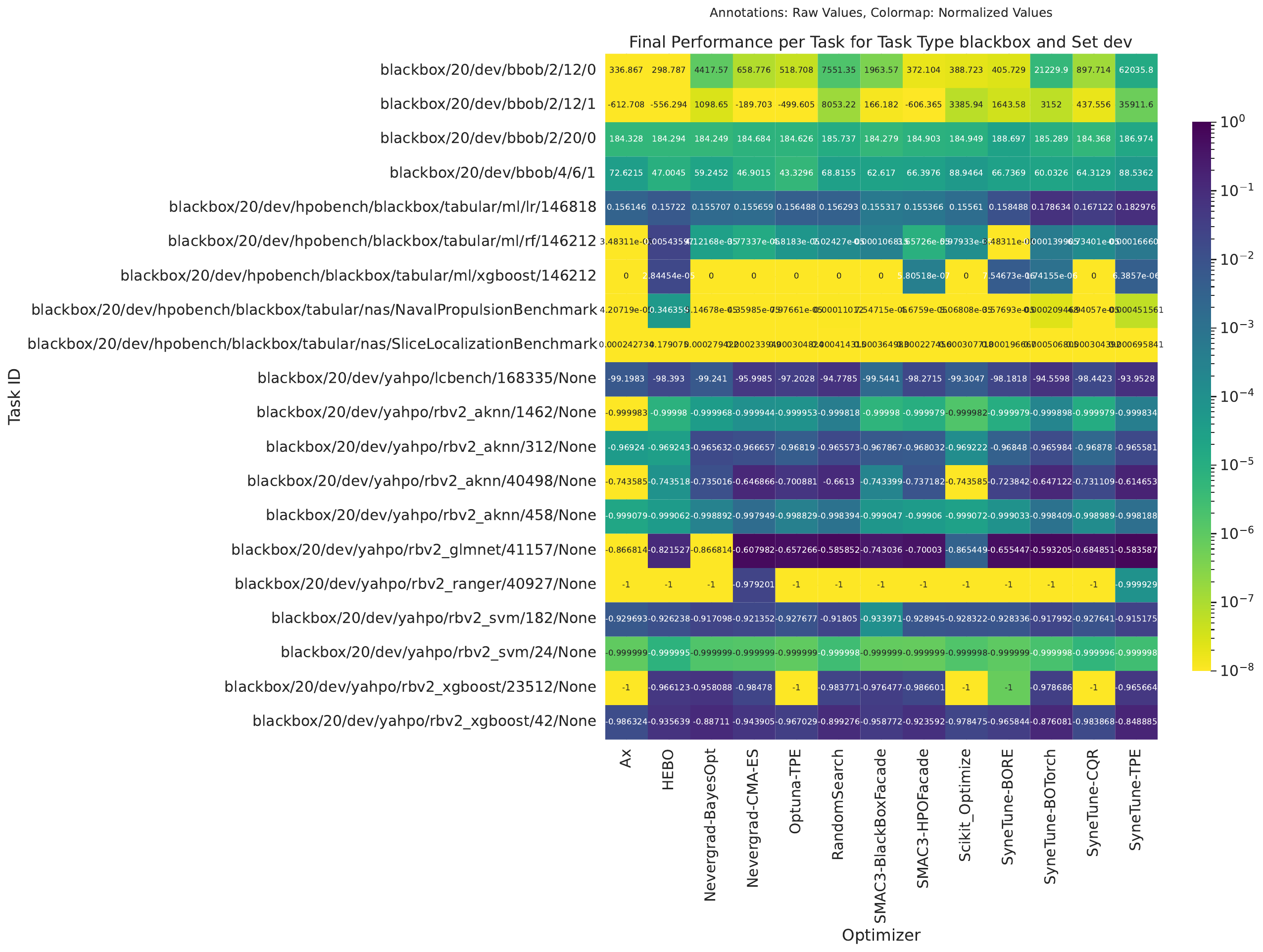}
    
    \caption{Result summary for Scenario blackbox and Set dev. First row: Critical difference diagram of final performance; rank over time based on statistical test. The grey area indicates non-significance based on statistical testing. Second row:  Performance per problem. The annotations of the heatmap cells indicate the raw final performance (mean over seeds) and the colormap indicates the normalized values.}
    \label{fig:bb_dev_summary}
\end{figure}
\begin{figure}[h]
    \centering
    \includegraphics[width=0.49\linewidth]{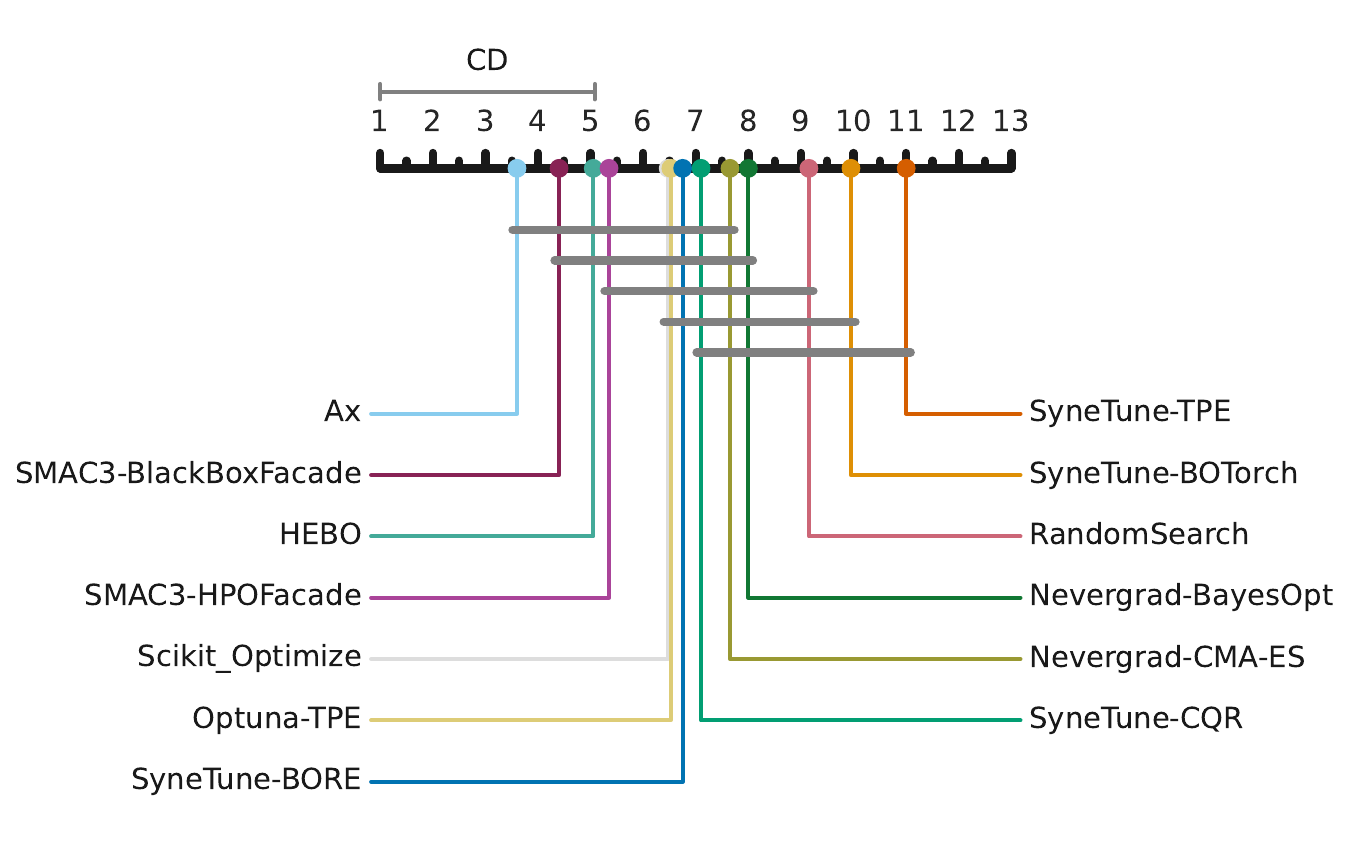}
    \hfill
    \includegraphics[width=0.49\linewidth]{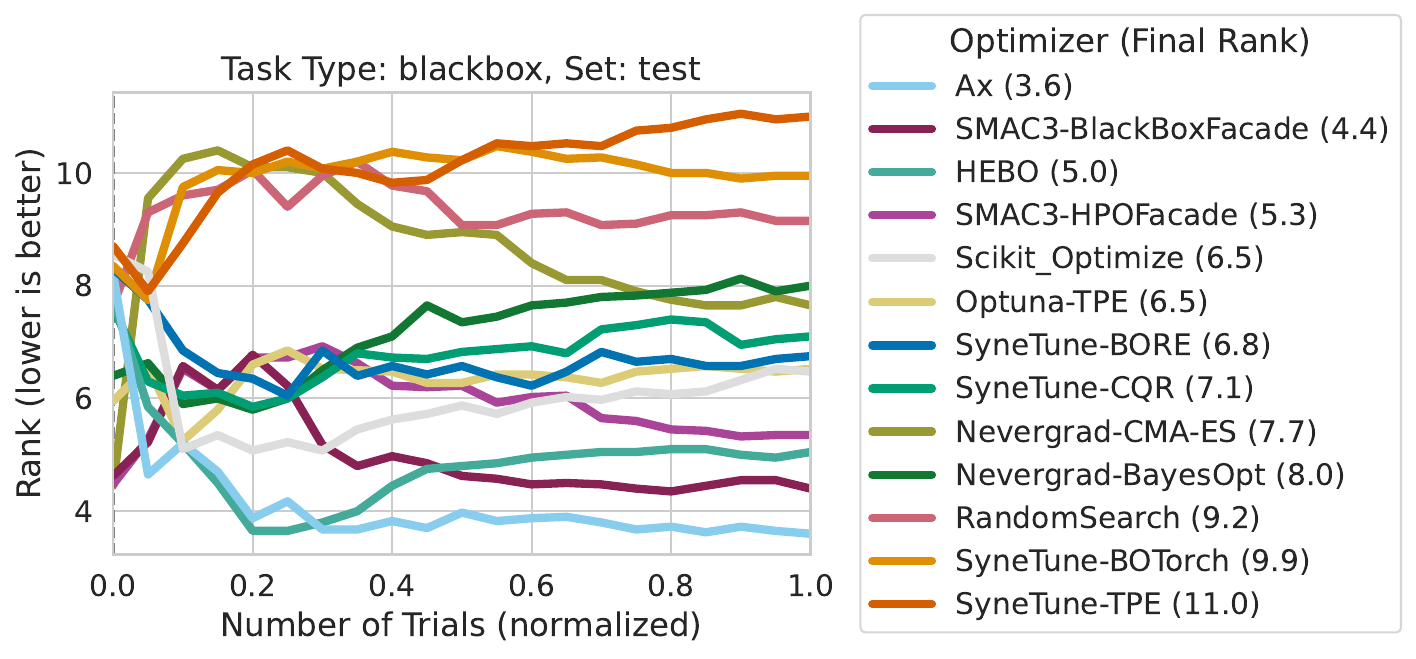}\\
    
    \includegraphics[width=0.8\linewidth]{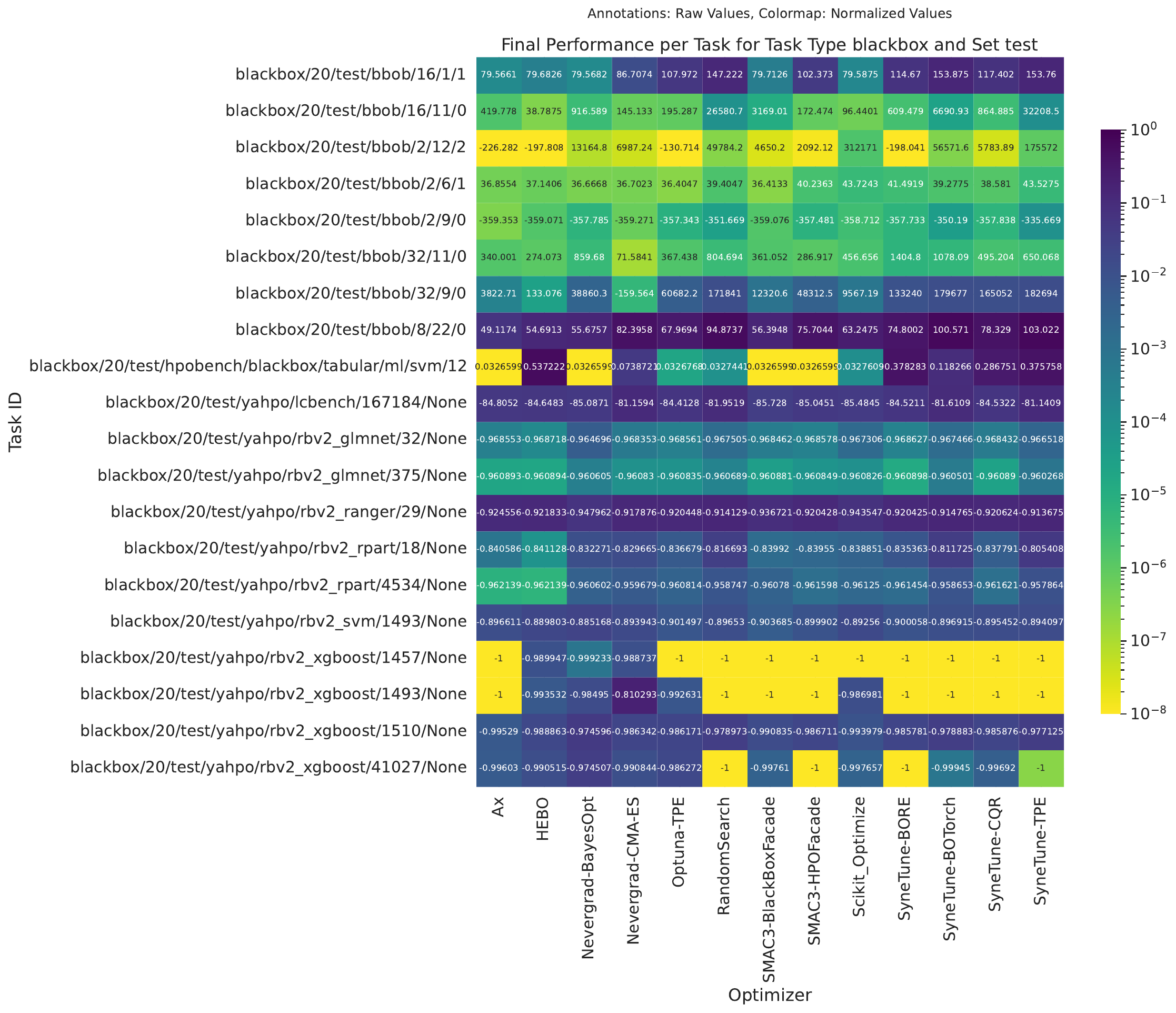}
    
    \caption{Result summary for Scenario blackbox and Set test. First row: Critical difference diagram of final performance; rank over time based on statistical test. The grey area indicates non-significance based on statistical testing. Second row:  Performance per problem. The annotations of the heatmap cells indicate the raw final performance (mean over seeds) and the colormap indicates the normalized values.}
    \label{fig:bb_test_summary}
\end{figure}
\begin{figure}[h]
    \centering
    \includegraphics[width=0.49\linewidth]{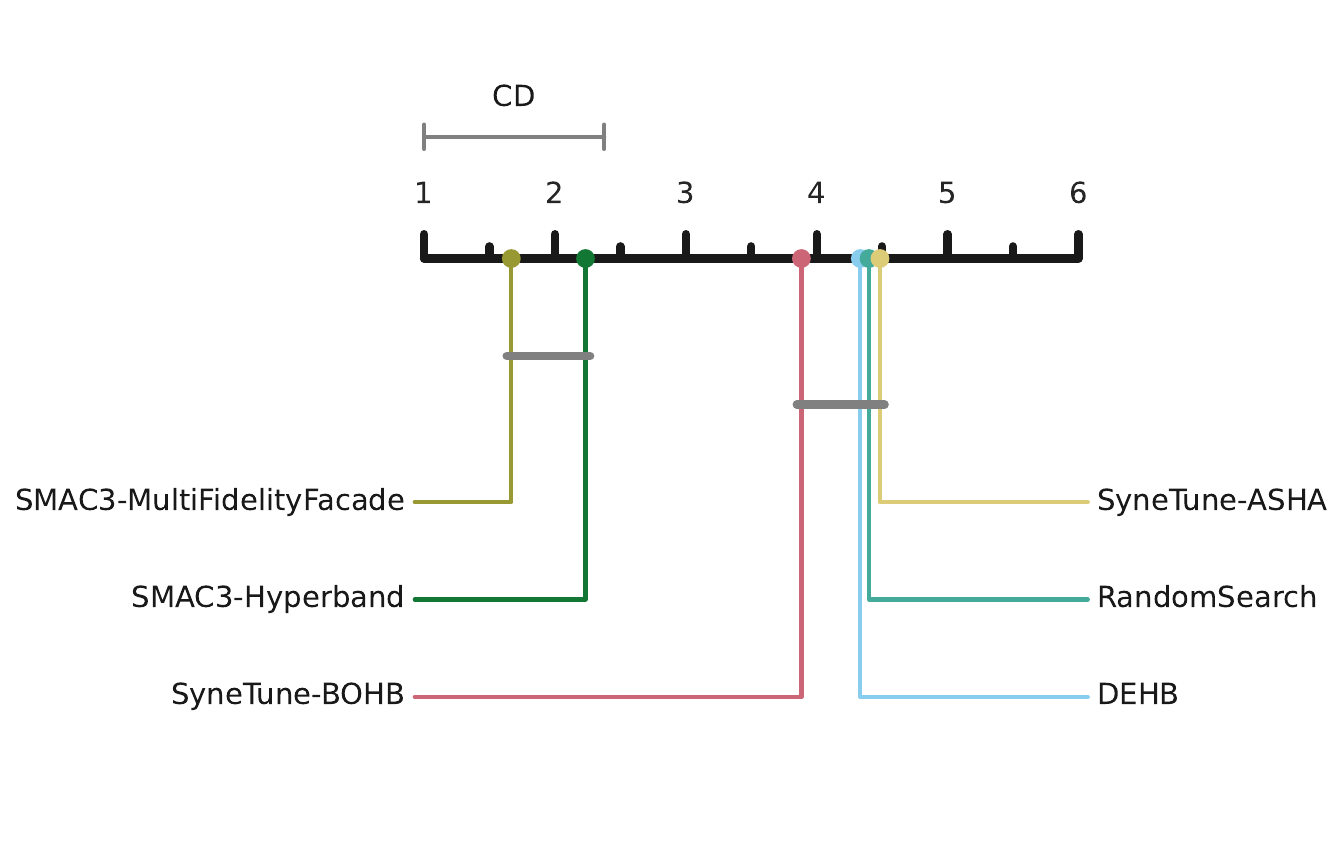}
    \hfill
    \includegraphics[width=0.49\linewidth]{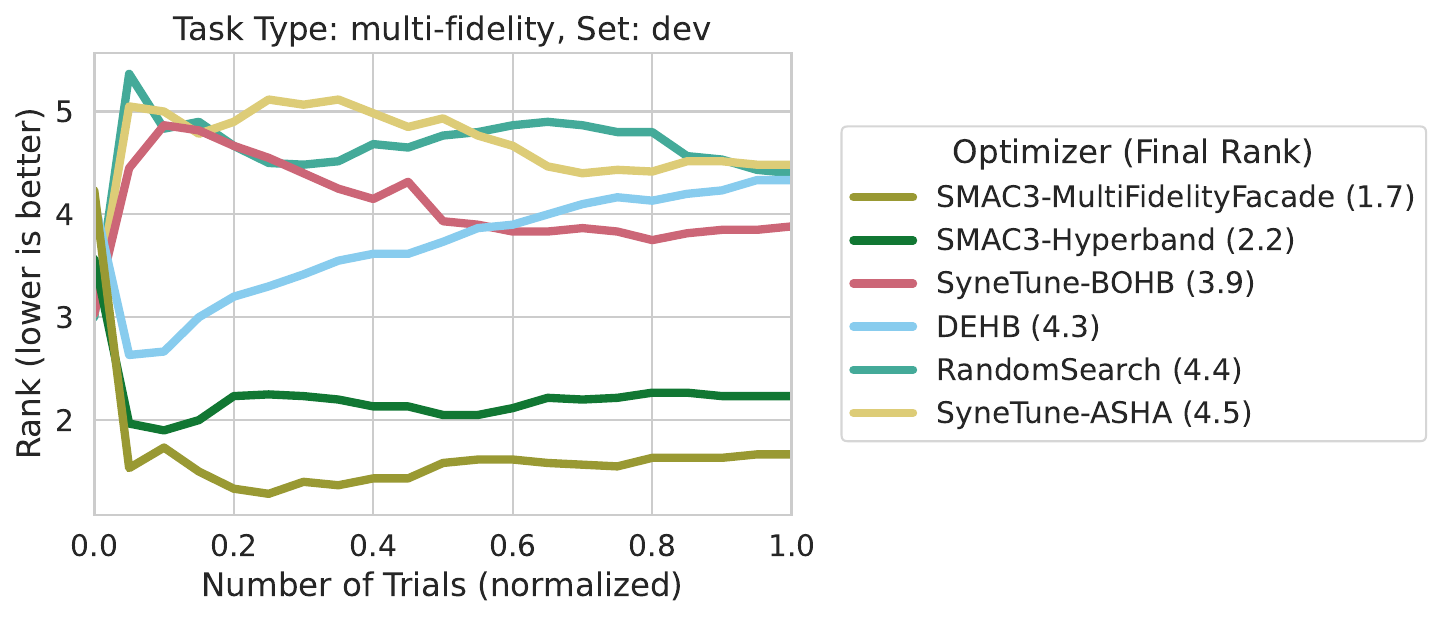}\\
    
    \includegraphics[width=0.8\linewidth]{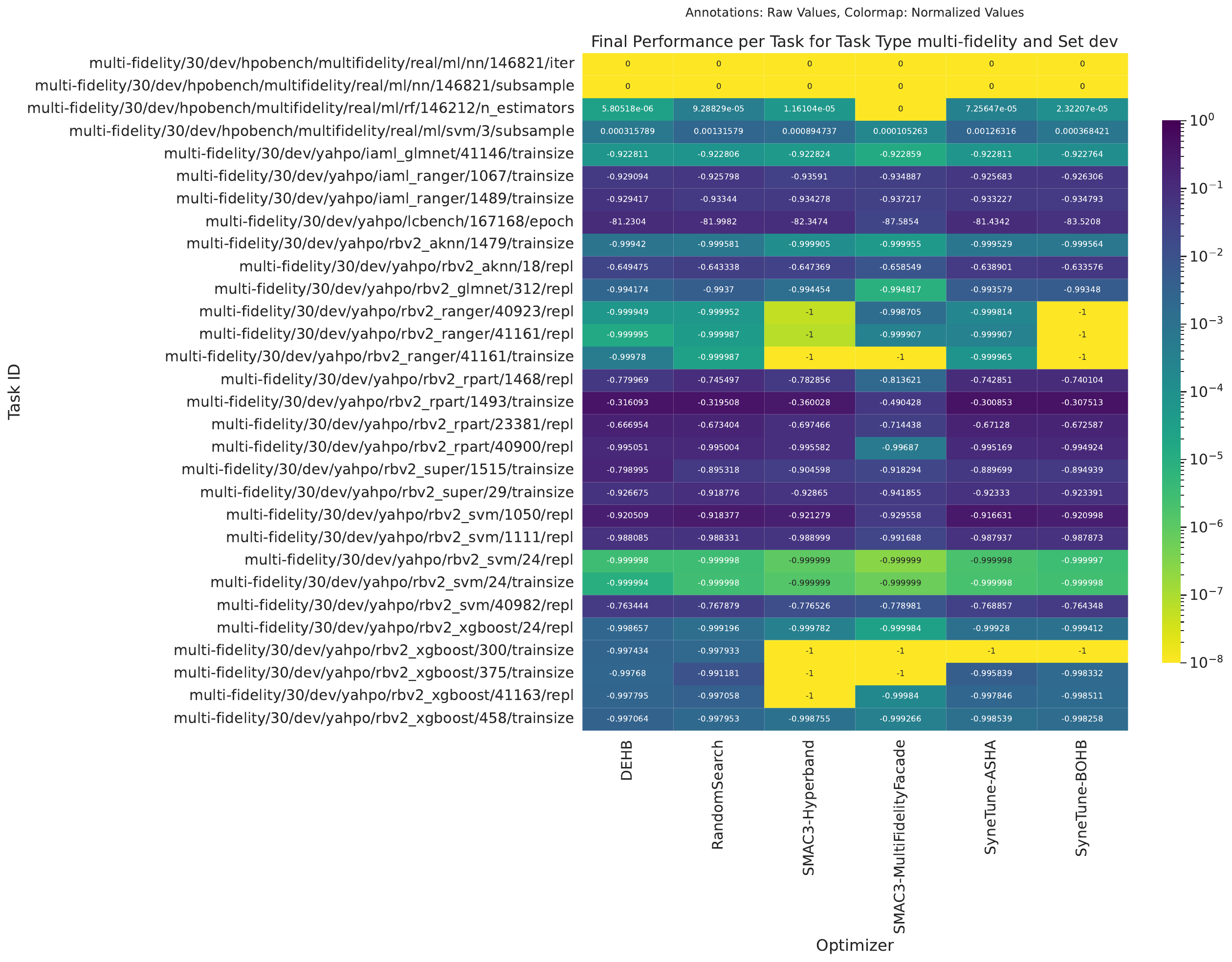}
    
    \caption{Result summary for Scenario multi-fidelity and Set dev. First row: Critical difference diagram of final performance; rank over time based on statistical test. The grey area indicates non-significance based on statistical testing. Second row:  Performance per problem. The annotations of the heatmap cells indicate the raw final performance (mean over seeds) and the colormap indicates the normalized values.}
    \label{fig:mf_dev_summary}
\end{figure}
\begin{figure}[h]
    \centering
    \includegraphics[width=0.49\linewidth]{figures/results/multi-fidelity_test_criticaldifference.pdf}
    \hfill
    \includegraphics[width=0.49\linewidth]{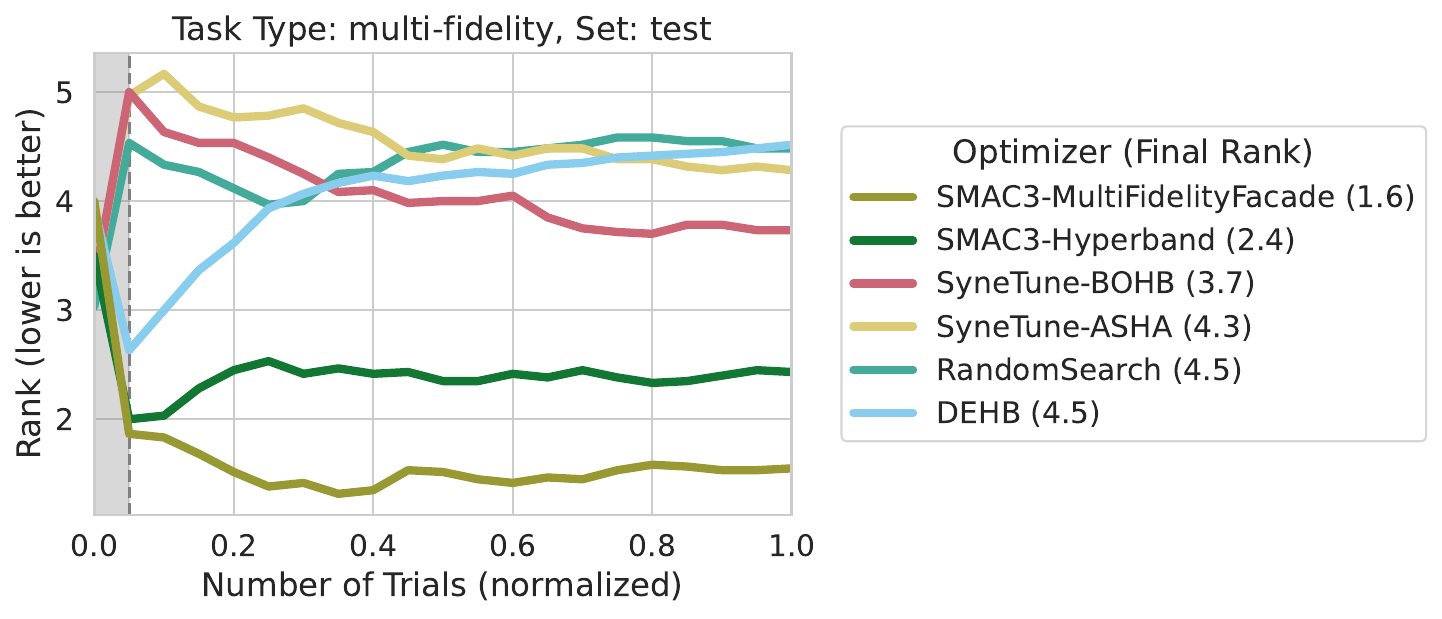}\\
    
    \includegraphics[width=0.8\linewidth]{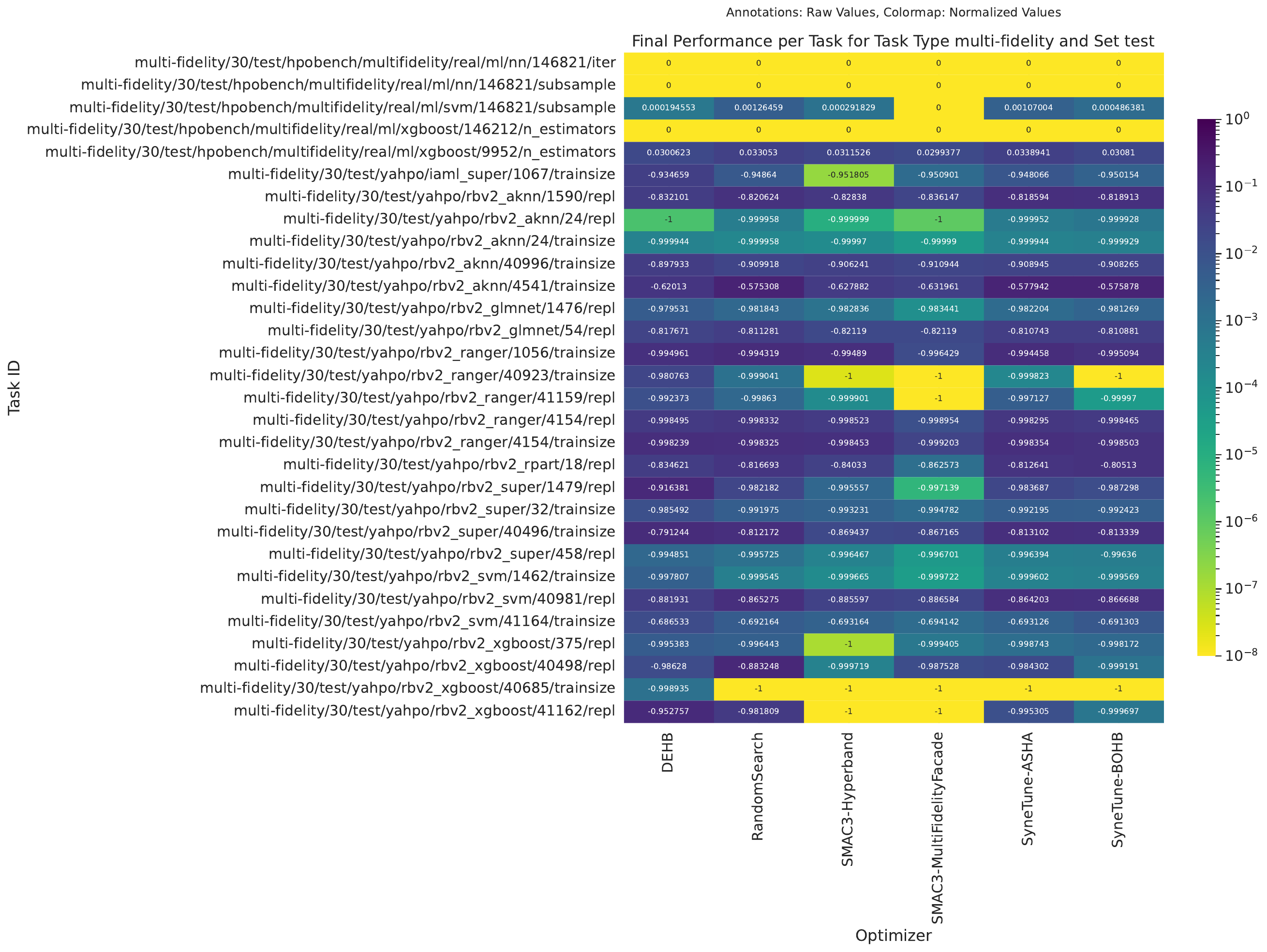}
    
    \caption{Result summary for Scenario multi-fidelity and Set test. First row: Critical difference diagram of final performance; rank over time based on statistical test. The grey area indicates non-significance based on statistical testing. Second row:  Performance per problem. The annotations of the heatmap cells indicate the raw final performance (mean over seeds) and the colormap indicates the normalized values.}
    \label{fig:mf_test_summary}
\end{figure}
\begin{figure}[h]
    \centering
    \includegraphics[width=0.49\linewidth]{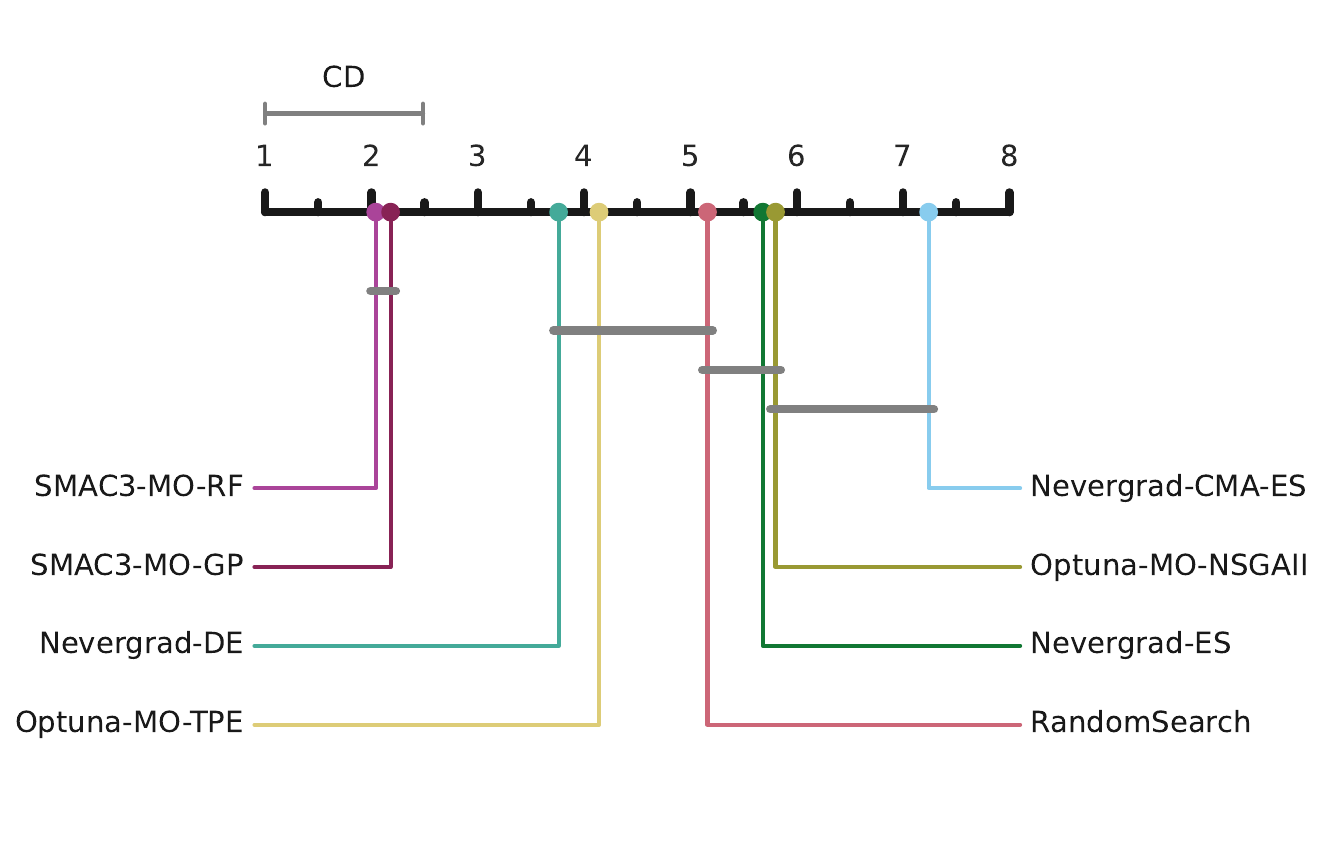}
    \hfill
    \includegraphics[width=0.49\linewidth]{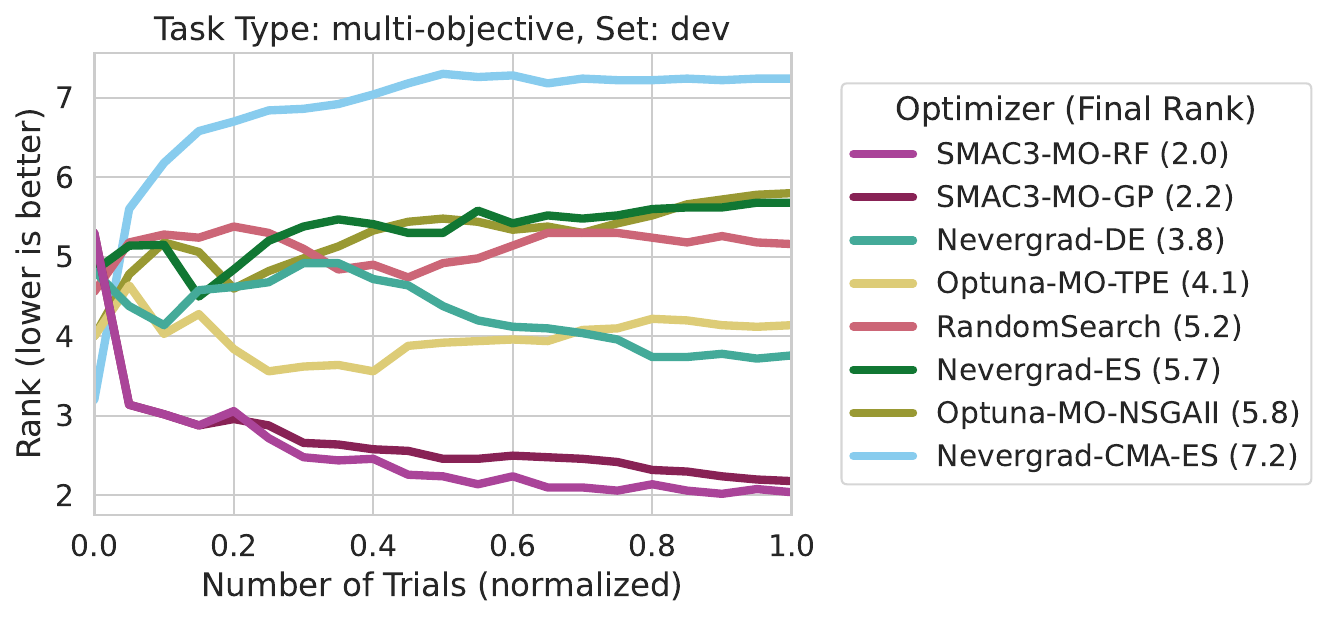}\\
    
    \includegraphics[width=0.8\linewidth]{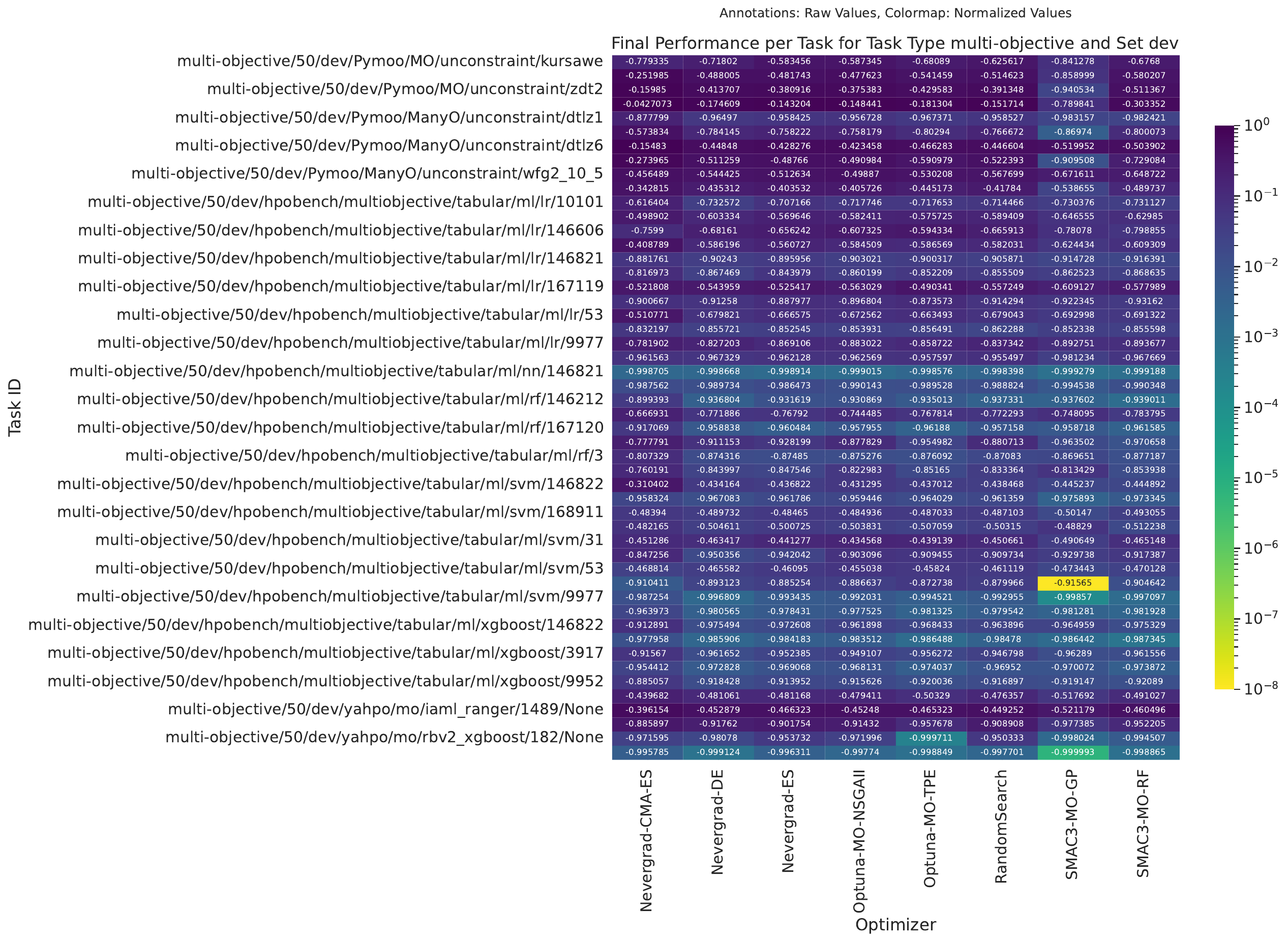}
    
    \caption{Result summary for Scenario multi-objective and Set dev. First row: Critical difference diagram of final performance; rank over time based on statistical test. The grey area indicates non-significance based on statistical testing. Second row:  Performance per problem. The annotations of the heatmap cells indicate the raw final performance (mean over seeds) and the colormap indicates the normalized values.}
    \label{fig:mo_dev_summary}
\end{figure}
\begin{figure}[h]
    \centering
    \includegraphics[width=0.49\linewidth]{figures/results/multi-objective_test_criticaldifference.pdf}
    \hfill
    \includegraphics[width=0.49\linewidth]{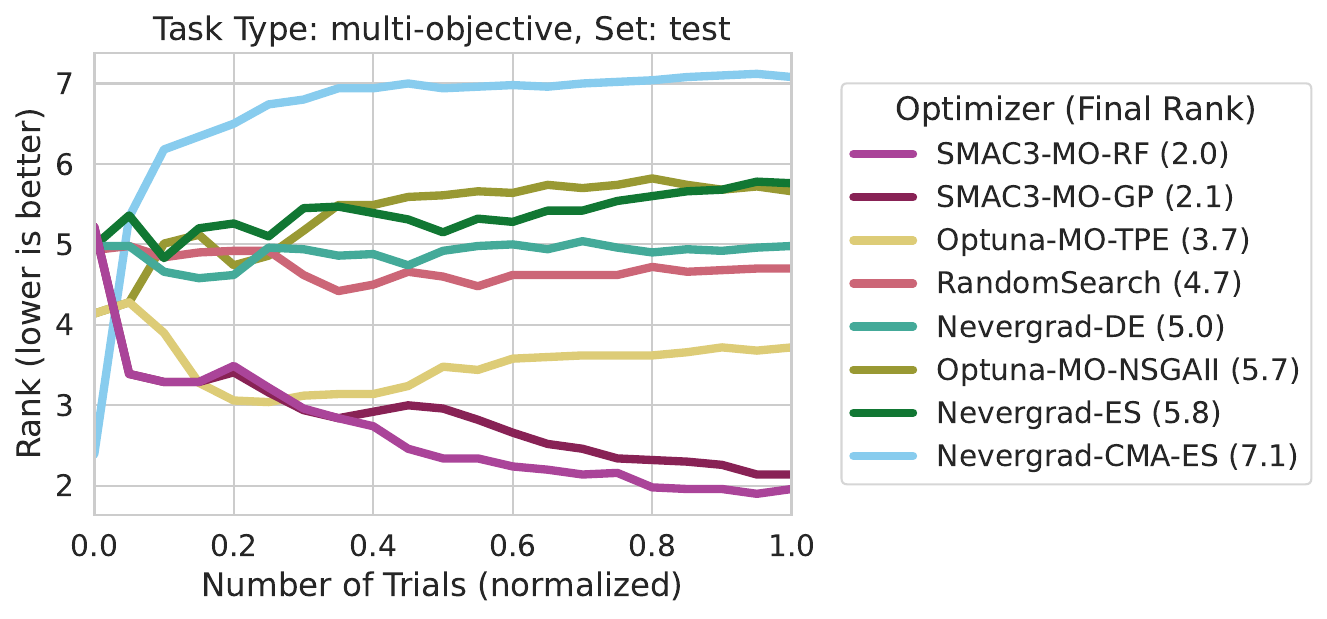}\\
    
    \includegraphics[width=0.8\linewidth]{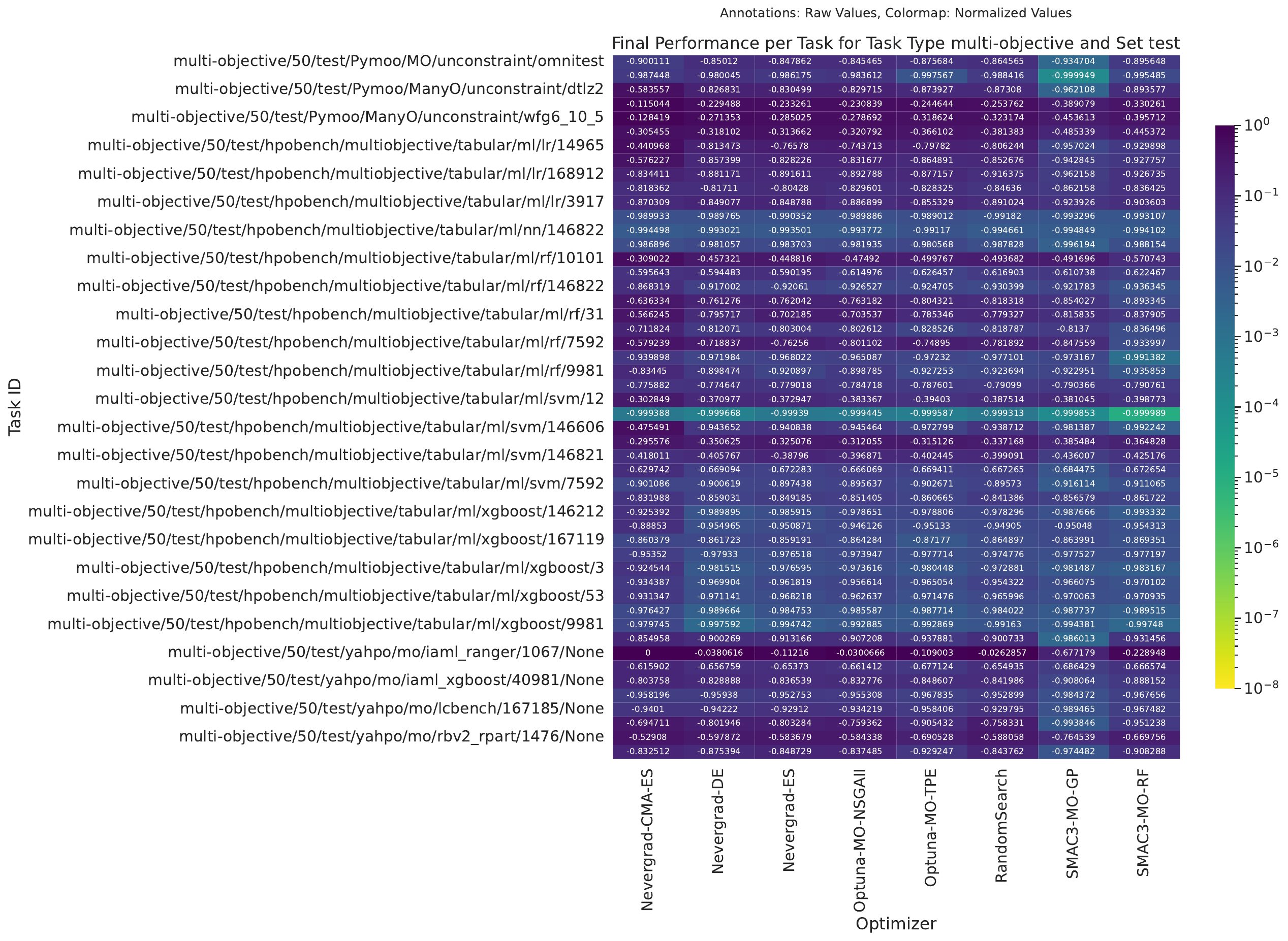}
    
    \caption{Result summary for Scenario multi-objective and Set test. First row: Critical difference diagram of final performance; rank over time based on statistical test. The grey area indicates non-significance based on statistical testing. Second row:  Performance per problem. The annotations of the heatmap cells indicate the raw final performance (mean over seeds) and the colormap indicates the normalized values.}
    \label{fig:mo_test_summary}
\end{figure}
\begin{figure}[h]
    \centering
    \includegraphics[width=0.49\linewidth]{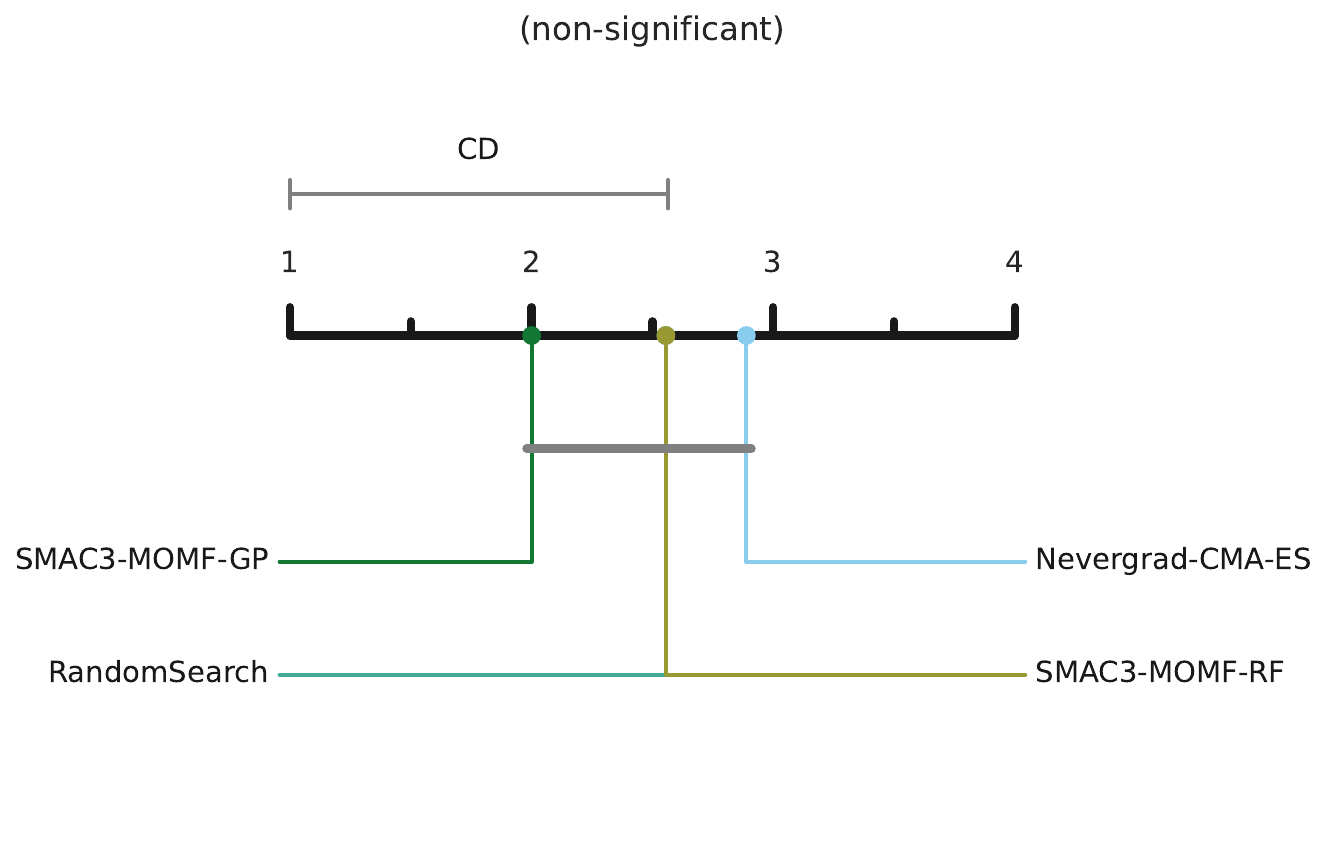}
    \hfill
    \includegraphics[width=0.49\linewidth]{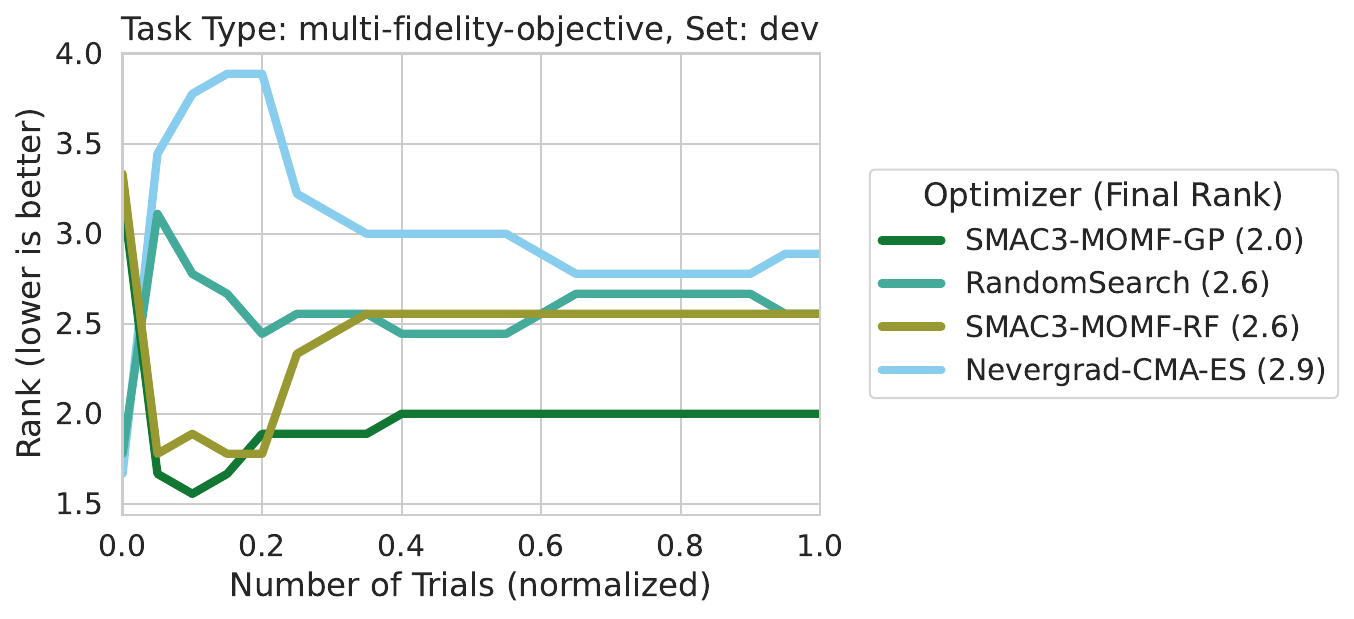}\\
    
    \includegraphics[width=0.8\linewidth]{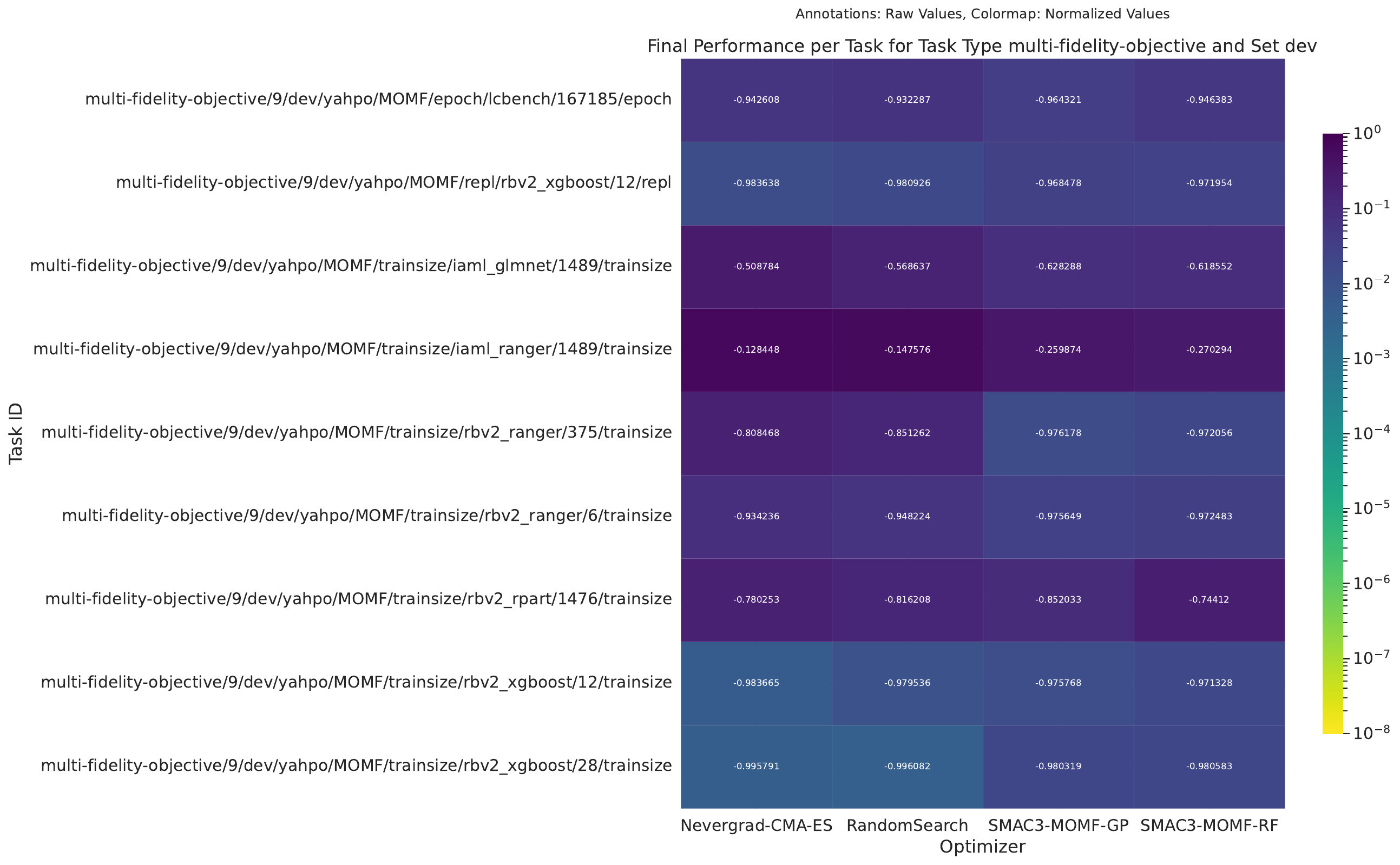}
    
    \caption{Result summary for Scenario multi-fidelity-objective and Set dev. First row: Critical difference diagram of final performance; rank over time based on statistical test. The grey area indicates non-significance based on statistical testing. Second row:  Performance per problem. The annotations of the heatmap cells indicate the raw final performance (mean over seeds) and the colormap indicates the normalized values.}
    \label{fig:momf_dev_summary}
\end{figure}
\begin{figure}[h]
    \centering
    \includegraphics[width=0.49\linewidth]{figures/results/multi-fidelity-objective_test_criticaldifference.pdf}
    \hfill
    \includegraphics[width=0.49\linewidth]{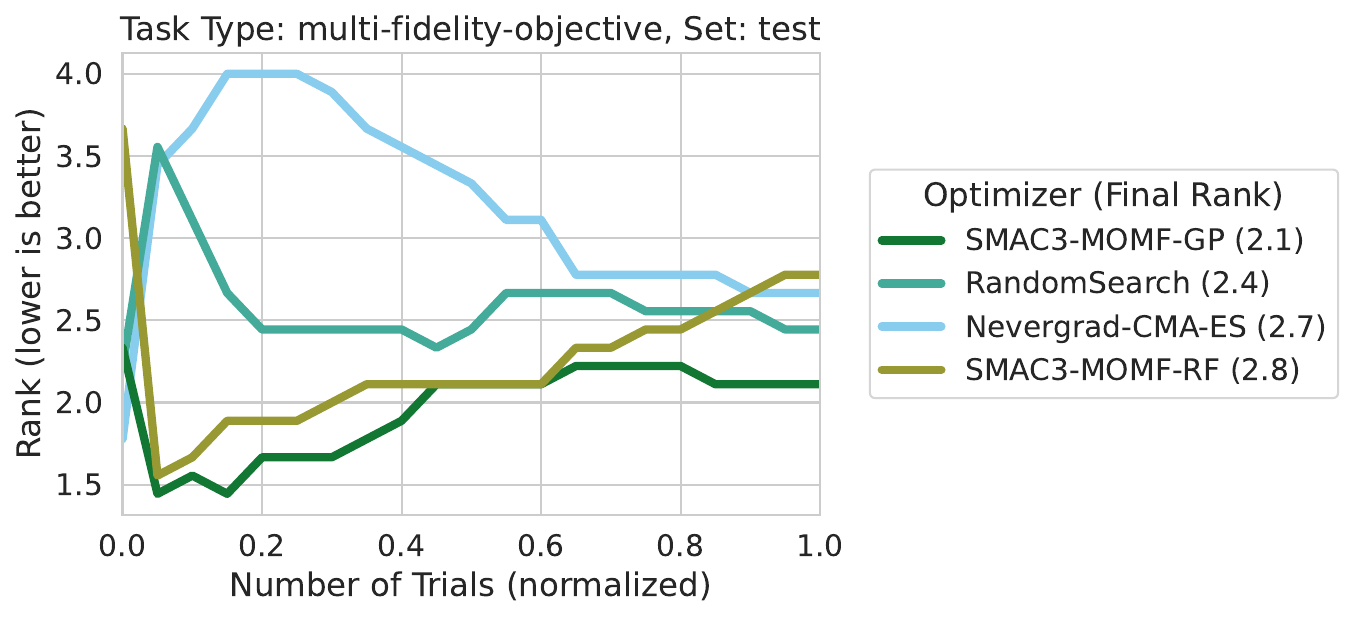}\\
    
    \includegraphics[width=0.8\linewidth]{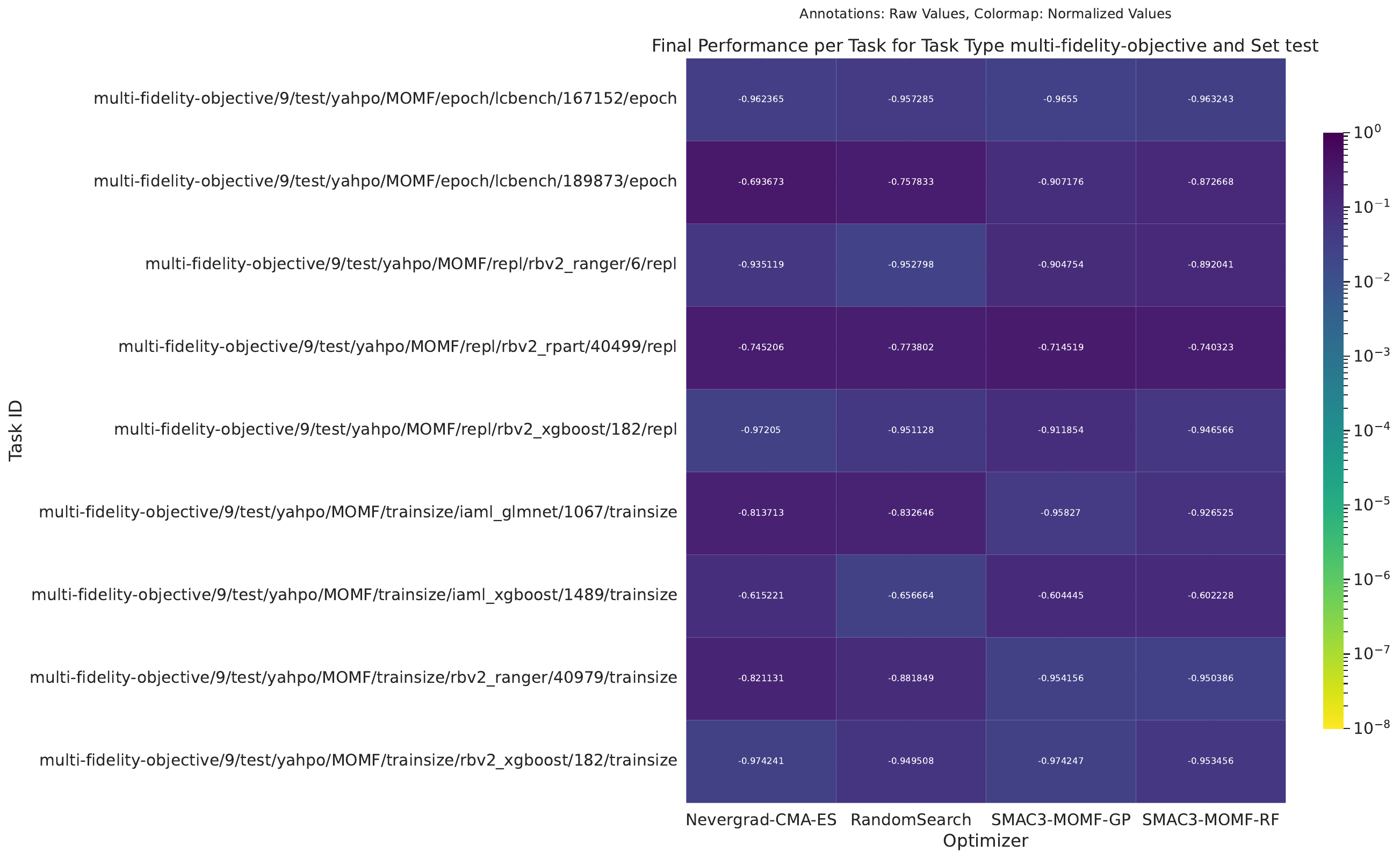}
    
    \caption{Result summary for Scenario multi-fidelity-objective and Set test. First row: Critical difference diagram of final performance; rank over time based on statistical test. The grey area indicates non-significance based on statistical testing. Second row:  Performance per problem. The annotations of the heatmap cells indicate the raw final performance (mean over seeds) and the colormap indicates the normalized values.}
    \label{fig:momf_test_summary}
\end{figure}


\end{document}